\documentclass[12pt]{article}
\usepackage[textwidth=390pt,textheight=548.5pt]{geometry}
\usepackage{xcolor}

\usepackage[T1]{fontenc}
\usepackage{amsmath,amssymb,amsthm}
\usepackage{graphicx}
\usepackage{booktabs}
\usepackage{longtable}
\usepackage{array}
\usepackage{enumitem}
\usepackage[natbibapa]{apacite}
\usepackage{microtype}
\usepackage{flafter}
\theoremstyle{plain}
\newtheorem{theorem}{Theorem}
\newtheorem{proposition}{Proposition}
\newtheorem{lemma}{Lemma}
\newtheorem{corollary}{Corollary}
\theoremstyle{definition}
\newtheorem{assumption}{Assumption}

\theoremstyle{remark}

\newtheorem{example}{Example}

\usepackage{bm}
\usepackage{algorithm}
\usepackage{algpseudocode}
\usepackage[colorlinks=true,linkcolor=blue!60!black,citecolor=blue!60!black,urlcolor=blue!60!black]{hyperref}
\newcommand{\R}{\mathbb{R}}
\newcommand{\T}{^{\top}}
\newcommand{\one}{\mathbf{1}}
\newcommand{\zero}{\mathbf{0}}
\newcommand{\vv}{\bm{v}}
\newcommand{\ww}{\bm{w}}
\newcommand{\cc}{\bm{c}}
\newcommand{\rr}{\bm{r}}
\newcommand{\cw}{\bm{\pi}}
\newcommand{\tht}{\bm{\theta}}
\newcommand{\ph}{\bm{\phi}}
\newcommand{\Ph}{\Phi}
\newcommand{\pos}[1]{\bigl(#1\bigr)_{+}}
\DeclareMathOperator*{\argmin}{arg\,min}

\newcommand{\minus}{\ensuremath{-}}  
\newcommand{\magn}[1]{{\renewcommand{\minus}{}#1}}  

\newcommand{\addGainChebUtaSmall}{3.2}
\newcommand{\addGainChebUtaSmallSE}{0.7}

\newcommand{\addGainLsuAcenMax}{0.3}
\newcommand{\addGainLsuAcenMin}{\minus 0.4}

\newcommand{\addGainLsuChebLosses}{1}
\newcommand{\addGainLsuChebMax}{0.1}
\newcommand{\addGainLsuChebMin}{\minus 0.9}

\newcommand{\addGainLsuUtaMMSmall}{2.1}

\newcommand{\atConsistentAbsMax}{0.9}

\newcommand{\auditConicSolved}{26}
\newcommand{\auditDevMax}{1.1e-08}

\newcommand{\cenAcenInnerFail}{2306}
\newcommand{\cenAcenQsInnerFail}{115}
\newcommand{\cenCand}{3}
\newcommand{\cenChebInnerFail}{2186}
\newcommand{\cenChebQsInnerFail}{0}

\newcommand{\cenFallback}{6}
\newcommand{\cenFits}{1800}
\newcommand{\cevBound}{0.0035}
\newcommand{\cevShortfall}{0.40}
\newcommand{\cevShortfallMilli}{0.040}

\newcommand{\citDzClasses}{2}
\newcommand{\citDzMargin}{1}
\newcommand{\citDzRef}{1500}
\newcommand{\citLiuClasses}{5}
\newcommand{\citLiuSimMax}{1800}
\newcommand{\citLiuSimMin}{600}
\newcommand{\citLiuUniv}{600}
\newcommand{\citMkAltMax}{1728}
\newcommand{\citMkAltMin}{120}
\newcommand{\citMkMrLearners}{2}
\newcommand{\citMkReps}{100}
\newcommand{\citMkSegments}{3}
\newcommand{\citMkSets}{9}
\newcommand{\citMkShareMax}{80}
\newcommand{\citMkShareMid}{50}
\newcommand{\citMkShareMin}{20}
\newcommand{\citPedBetter}{7}
\newcommand{\citPedData}{9}
\newcommand{\citPedSig}{3}
\newcommand{\citPedWays}{2}
\newcommand{\citRuAltMax}{167}
\newcommand{\citRuAltMin}{48}
\newcommand{\citRuMethods}{2}
\newcommand{\citRuShareMax}{20}
\newcommand{\citRuShareMin}{5}
\newcommand{\citRuTables}{10}
\newcommand{\citWojMax}{10}
\newcommand{\citWojMin}{3}
\newcommand{\citWojProcedures}{16}
\newcommand{\computeHours}{15}
\newcommand{\conicDevMax}{3e-08}
\newcommand{\conicGapMax}{1e-14}
\newcommand{\conicRidgeFits}{4100}

\newcommand{\conicSolvedShare}{84}

\newcommand{\envClarabel}{0.11.1}
\newcommand{\envCpu}{AMD Ryzen AI 7 PRO 350}
\newcommand{\envCvxpy}{1.8.1}
\newcommand{\envLogical}{16}
\newcommand{\envMemory}{87.6}
\newcommand{\envNumpy}{2.4.2}
\newcommand{\envPython}{3.14.3}
\newcommand{\envScipy}{1.17.1}
\newcommand{\envSklearn}{1.8.0}
\newcommand{\fitsCount}{4100}
\newcommand{\fixCells}{6}
\newcommand{\fixChebBehindM}{80}

\newcommand{\fixFits}{14400}

\newcommand{\fixGainLsuAcenQSSmall}{2.6}
\newcommand{\fixGainLsuAcenQSSmallSE}{0.6}

\newcommand{\fixGainLsuChebMax}{0.8}

\newcommand{\fixGainLsuChebMin}{\minus 0.5}

\newcommand{\fixGainLsuOlogAAddSmall}{\minus 1.5}
\newcommand{\fixGainLsuOlogAAddSmallSE}{0.5}
\newcommand{\fixGainLsuOlogALosses}{3}
\newcommand{\fixGainLsuOlogAMax}{0.9}

\newcommand{\fixGainLsuOlogAMin}{\minus 1.5}
\newcommand{\fixGainLsuOlogAQSLarge}{\minus 0.2}
\newcommand{\fixGainLsuOlogAQSLargeSE}{0.1}

\newcommand{\fixGainLsuOlogAQSSmall}{0.9}
\newcommand{\fixGainLsuOlogAQSSmallSE}{0.5}

\newcommand{\fixGainLsuUtaMMQSSmall}{4.0}
\newcommand{\fixGainLsuUtaMMQSSmallSE}{0.5}

\newcommand{\fixGainLsuUtaRidgeAddSmall}{0.0}
\newcommand{\fixGainLsuUtaRidgeAddSmallSE}{0.1}

\newcommand{\fixGainLsuUtaRidgeMax}{1.6}
\newcommand{\fixGainLsuUtaRidgeMaxM}{80}

\newcommand{\fixGainLsuUtaRidgeQSSmall}{0.7}
\newcommand{\fixGainLsuUtaRidgeQSSmallSE}{0.2}
\newcommand{\fixGainLsuUtaRidgeWins}{5}
\newcommand{\fixGainUtaRidgeUtaAddSmall}{5.2}
\newcommand{\fixGainUtaRidgeUtaAddSmallSE}{0.7}

\newcommand{\fixGainUtaRidgeUtaQSSmall}{3.2}
\newcommand{\fixGainUtaRidgeUtaQSSmallSE}{0.5}

\newcommand{\fixLossLsuOlogAMax}{1.5}
\newcommand{\fixMarginCount}{4}
\newcommand{\fixMarginLsuBestAtFixed}{5}
\newcommand{\fixMarginLsuBestChebMax}{0.8}
\newcommand{\fixMarginLsuBestChebMin}{\minus 0.5}
\newcommand{\fixMarginLsuBestOlogAMax}{0.9}
\newcommand{\fixMarginLsuBestOlogAMin}{\minus 1.5}
\newcommand{\fixMarginLsuGainMax}{0.2}

\newcommand{\fixMarginUtaGainMax}{3.2}
\newcommand{\fixMlarge}{80}
\newcommand{\fixMsecond}{40}
\newcommand{\fixMsmall}{20}
\newcommand{\fixOlogAFits}{1800}

\newcommand{\fixRhoCount}{5}
\newcommand{\fixRhoLsuRange}{1.3}
\newcommand{\fixRhoUtaRidgeRange}{0.2}
\newcommand{\fixRidgeOlogARange}{1.4}

\newcommand{\fixSquareByRhoMax}{1.7}
\newcommand{\fixSquareByRhoMin}{0.0}
\newcommand{\fixSquareByRhoQSSmallMax}{1.7}
\newcommand{\fixSquareByRhoQSSmallMin}{0.6}

\newcommand{\flipsCleanLsuJLarge}{0.771}
\newcommand{\flipsCleanLsuLarge}{0.801}
\newcommand{\flipsCleanOlogALarge}{0.868}

\newcommand{\flipsCleanUtaLarge}{0.893}
\newcommand{\flipsCleanUtaMMLarge}{0.817}
\newcommand{\flipsCleanUtaRegLarge}{0.894}

\newcommand{\flipsGainLsuUtaMMMin}{\minus 1.6}

\newcommand{\flipsGainLsuUtaMax}{\minus 5.1}
\newcommand{\flipsGainLsuUtaMin}{\minus 9.2}

\newcommand{\flipsGainUtaRegUtaMin}{\minus 2.9}

\newcommand{\flipsLossMax}{9.2}

\newcommand{\flipsLossOlogAUtaMax}{2.5}
\newcommand{\flipsLossOlogAUtaMin}{1.8}

\newcommand{\floorClsLargeSpan}{between 0.691 and 0.729}

\newcommand{\floorLsuLargeSpan}{0.979 at each of the three floors}

\newcommand{\floorLsuSmallSpan}{0.858 at each of the three floors}
\newcommand{\hubConsistentLossMax}{1.0}

\newcommand{\hubFlipsLossMax}{2.0}

\newcommand{\hubKnotSmallestShare}{65}
\newcommand{\hubKnotSmallestShareFlips}{62}
\newcommand{\hubMainFlipsLossMax}{0.3}
\newcommand{\hubMainFlipsWithin}{0.4}
\newcommand{\hubMainLossMax}{1.0}
\newcommand{\iterMax}{33}
\newcommand{\iterMean}{5.3}
\newcommand{\kktAboveTarget}{32}

\newcommand{\kktFits}{33300}
\newcommand{\kktInnerAboveTarget}{738}

\newcommand{\kktMaxAll}{3.0e-08}
\newcommand{\kktMaxInner}{7.6e-08}
\newcommand{\kktMaxProfile}{1.3e-09}
\newcommand{\kktMaxStability}{8.7e-11}

\newcommand{\majEraAcc}{0.181}
\newcommand{\majEraMae}{1.78}
\newcommand{\majEslAcc}{0.277}
\newcommand{\majEslMae}{1.23}
\newcommand{\majLevAcc}{0.403}
\newcommand{\majLevMae}{0.72}
\newcommand{\majSwdAcc}{0.399}
\newcommand{\majSwdMae}{0.63}
\newcommand{\mindistMin}{2e-09}
\newcommand{\mmFromSecondCells}{7}
\newcommand{\mmFromSecondLsuWinM}{160}
\newcommand{\mmFromSecondLsuWins}{1}
\newcommand{\mmFromSecondMax}{0.9}

\newcommand{\mpgMissing}{6}
\newcommand{\nCev}{1728}
\newcommand{\nCpu}{209}
\newcommand{\nEra}{1000}
\newcommand{\nEsl}{488}
\newcommand{\nLev}{1000}
\newcommand{\nMpg}{392}
\newcommand{\nSwd}{1000}

\newcommand{\ologAuditChangedMax}{0.1}
\newcommand{\ologAuditFits}{900}
\newcommand{\ologAuditGapMax}{8.6e-06}
\newcommand{\ologAuditGradMax}{1.9e-02}
\newcommand{\ologAuditLower}{3}
\newcommand{\ologLimAccDiffMax}{1e-16}
\newcommand{\ologLimEraShare}{47}
\newcommand{\ologLimFits}{4100}
\newcommand{\ologLimOtherShareMax}{4.3}
\newcommand{\ologLimOut}{252}
\newcommand{\ologLimSwdShare}{28}
\newcommand{\ologMainCells}{41}
\newcommand{\ologMainLossCells}{1}
\newcommand{\ologMainWinCells}{31}
\newcommand{\openmlAccMax}{0.698}
\newcommand{\openmlCells}{12}
\newcommand{\openmlCrossedMax}{97}
\newcommand{\openmlCrossedMin}{38}

\newcommand{\openmlGainLsuLsuJMax}{3.6}

\newcommand{\openmlGainLsuOlogALosses}{11}

\newcommand{\openmlGainLsuOlogAMin}{\minus 1.8}

\newcommand{\openmlGainLsuOlogLosses}{10}
\newcommand{\openmlGainLsuOlogMax}{0.1}
\newcommand{\openmlGainLsuOlogMin}{\minus 2.9}

\newcommand{\openmlGainLsuRelMax}{1.0}

\newcommand{\openmlGainLsuUtaLosses}{3}

\newcommand{\openmlGainLsuUtaRegMax}{0.9}
\newcommand{\openmlGainLsuUtaRegMin}{\minus 0.7}
\newcommand{\openmlGainLsuUtaRegWins}{2}
\newcommand{\openmlGainLsuUtaRidgeLosses}{7}
\newcommand{\openmlGainLsuUtaRidgeMax}{0.4}
\newcommand{\openmlGainLsuUtaRidgeMin}{\minus 0.9}

\newcommand{\openmlGainOlogAUtaWins}{11}

\newcommand{\openmlGainUtaRegUtaLosses}{3}
\newcommand{\openmlGainUtaRegUtaMax}{0.8}
\newcommand{\openmlGainUtaRegUtaMin}{\minus 1.4}

\newcommand{\openmlGainUtaRegUtaWins}{1}

\newcommand{\openmlMaxHalfWidth}{0.015}
\newcommand{\openmlRecalShare}{65}
\newcommand{\pCev}{6}
\newcommand{\pCpu}{6}
\newcommand{\pEra}{4}
\newcommand{\pEsl}{4}
\newcommand{\pLev}{4}
\newcommand{\pMpg}{6}
\newcommand{\pSwd}{10}

\newcommand{\plAddLargeBestDiff}{\minus 0.2}
\newcommand{\plAddLargeBestDiffSE}{0.1}

\newcommand{\plAddLargeM}{80}

\newcommand{\plAddSmallAcc}{0.787}
\newcommand{\plAddSmallBestDiff}{\minus 1.8}
\newcommand{\plAddSmallBestDiffSE}{0.4}

\newcommand{\plAddSmallRidgeRange}{0.5}

\newcommand{\plQsLargeBestDiff}{0.3}
\newcommand{\plQsLargeBestDiffSE}{0.1}
\newcommand{\plQsLargeBothBestDiff}{0.00}
\newcommand{\plQsLargeBothBestDiffSE}{0.03}
\newcommand{\plQsLargeLikBestRidgeExp}{\minus 4}

\newcommand{\plQsMidBestDiff}{0.1}
\newcommand{\plQsMidBestDiffSE}{0.2}
\newcommand{\plQsMidBothBestDiff}{\minus 0.19}
\newcommand{\plQsMidBothBestDiffSE}{0.14}

\newcommand{\plQsMidLsuBelow}{8}

\newcommand{\plQsMidM}{80}
\newcommand{\plQsMidMargins}{10}

\newcommand{\plQsSmallAcc}{0.847}
\newcommand{\plQsSmallBestDiff}{1.4}
\newcommand{\plQsSmallBestDiffSE}{0.5}
\newcommand{\plQsSmallBothBestDiff}{1.4}
\newcommand{\plQsSmallBothBestDiffSE}{0.5}

\newcommand{\plQsSmallLsuAboveSet}{0.3}

\newcommand{\plQsSmallLsuBelowSet}{0, 0.05, 0.75, 0.9}

\newcommand{\plQsSmallLsuSig}{5}

\newcommand{\plQsSmallMargins}{10}
\newcommand{\plQsSmallRidgeRange}{1.1}

\newcommand{\profAddSmallAbsBestAcc}{0.770}

\newcommand{\profAddSmallRelBestAcc}{0.760}

\newcommand{\profQsLargeAbsBestAcc}{0.986}

\newcommand{\profQsLargeRelBest}{0.90}
\newcommand{\profQsLargeRelBestAcc}{0.978}
\newcommand{\profQsLargeRelFirst}{0.702}

\newcommand{\profQsSmallAbsBestAcc}{0.861}

\newcommand{\profQsSmallAbsRange}{4.6}
\newcommand{\profQsSmallRelBest}{0.80}
\newcommand{\profQsSmallRelBestAcc}{0.857}
\newcommand{\profQsSmallRelFirst}{0.680}
\newcommand{\profQsSmallRelInnerRange}{7.3}

\newcommand{\profQsSmallRelRange}{17.7}

\newcommand{\profQsSmallUtaBestAcc}{0.826}

\newcommand{\profQsSmallUtaRidgeBestAcc}{0.857}

\newcommand{\profQsSmallUtaRidgeRange}{8.2}
\newcommand{\qEra}{9}
\newcommand{\qEsl}{9}
\newcommand{\qLev}{5}
\newcommand{\qSwd}{4}
\newcommand{\qsAllCatSmall}{89}
\newcommand{\qsClsFloorMax}{25}
\newcommand{\qsClsFloorMin}{24}
\newcommand{\qsClsFreeMax}{0.895}
\newcommand{\qsClsFreeMin}{0.740}
\newcommand{\qsClsKLarge}{0.913}

\newcommand{\qsClsLarge}{0.706}

\newcommand{\qsClsOneMax}{0.704}
\newcommand{\qsClsOneMin}{0.678}
\newcommand{\qsClsSmall}{0.671}
\newcommand{\qsCountFour}{100}
\newcommand{\qsCountOne}{404}
\newcommand{\qsCountThree}{200}
\newcommand{\qsCountTwo}{300}
\newcommand{\qsFoldsSmall}{3.7}

\newcommand{\qsGainLsuAcenSecond}{0.9}
\newcommand{\qsGainLsuAcenSecondSE}{0.4}
\newcommand{\qsGainLsuAcenSmall}{0.3}
\newcommand{\qsGainLsuAcenSmallSE}{0.5}

\newcommand{\qsGainLsuChebLosses}{1}
\newcommand{\qsGainLsuChebMax}{0.0}
\newcommand{\qsGainLsuChebMin}{\minus 0.4}

\newcommand{\qsGainLsuChebSmall}{0.0}
\newcommand{\qsGainLsuChebSmallSE}{0.6}

\newcommand{\qsGainLsuLsuOneMax}{0.0}
\newcommand{\qsGainLsuLsuOneMin}{\minus 0.1}

\newcommand{\qsGainLsuOlogALarge}{0.2}
\newcommand{\qsGainLsuOlogALargeSE}{0.1}

\newcommand{\qsGainLsuOlogASecond}{\minus 0.9}
\newcommand{\qsGainLsuOlogASecondSE}{0.3}
\newcommand{\qsGainLsuOlogASmall}{\minus 0.9}
\newcommand{\qsGainLsuOlogASmallSE}{0.5}

\newcommand{\qsGainLsuOlogMax}{\minus 0.3}
\newcommand{\qsGainLsuOlogMin}{\minus 2.2}

\newcommand{\qsGainLsuRelMax}{1.1}

\newcommand{\qsGainLsuUtaBeyondMax}{0.7}

\newcommand{\qsGainLsuUtaMMSmall}{2.2}
\newcommand{\qsGainLsuUtaMMSmallSE}{0.7}

\newcommand{\qsGainLsuUtaRegLarge}{0.1}
\newcommand{\qsGainLsuUtaRegLargeSE}{0.1}

\newcommand{\qsGainLsuUtaRegSmall}{3.1}
\newcommand{\qsGainLsuUtaRegSmallSE}{0.7}

\newcommand{\qsGainLsuUtaRidgeSmall}{0.9}
\newcommand{\qsGainLsuUtaRidgeSmallSE}{0.4}
\newcommand{\qsGainLsuUtaRidgeWins}{5}
\newcommand{\qsGainLsuUtaSecond}{1.5}
\newcommand{\qsGainLsuUtaSecondSE}{0.5}
\newcommand{\qsGainLsuUtaSmall}{2.4}
\newcommand{\qsGainLsuUtaSmallSE}{0.7}

\newcommand{\qsGainOlogAUtaSmall}{3.3}
\newcommand{\qsGainOlogAUtaSmallSE}{0.6}

\newcommand{\qsGainUtaRegUtaMax}{0.6}
\newcommand{\qsGainUtaRegUtaMin}{\minus 0.7}

\newcommand{\qsGainUtaRegUtaWins}{2}

\newcommand{\qsGainUtaRidgeUtaSmall}{1.5}
\newcommand{\qsGainUtaRidgeUtaSmallSE}{0.7}

\newcommand{\qsLimitOne}{0.295}
\newcommand{\qsLimitThree}{0.684}
\newcommand{\qsLimitTwo}{0.461}
\newcommand{\qsLsuKLarge}{0.988}

\newcommand{\qsLsuLarge}{0.984}
\newcommand{\qsLsuOlogAWinSizes}{320}
\newcommand{\qsLsuSmall}{0.836}
\newcommand{\qsMlarge}{320}
\newcommand{\qsMsecond}{40}
\newcommand{\qsMsmall}{20}
\newcommand{\qsN}{1004}

\newcommand{\qsOlogALeadSizes}{40}

\newcommand{\qsStratSmall}{56}

\newcommand{\qsUtaSmall}{0.812}

\newcommand{\regTrivial}{2}
\newcommand{\regVsUtaCells}{41}
\newcommand{\regVsUtaLosses}{7}
\newcommand{\regVsUtaMax}{0.8}
\newcommand{\regVsUtaMin}{\minus 2.9}
\newcommand{\regVsUtaWins}{3}

\newcommand{\rerunHours}{1.5}
\newcommand{\rerunJobs}{13}
\newcommand{\shapeAddAtAbsMax}{0.7}
\newcommand{\shapeAddGainMax}{3.0}
\newcommand{\shapeCells}{6}

\newcommand{\shapeFlipsLossMax}{11.8}

\newcommand{\shapeMSmall}{20}
\newcommand{\shapeMaxHalfWidth}{0.024}
\newcommand{\shapeReps}{100}
\newcommand{\simAddLsuLarge}{0.955}

\newcommand{\simAddOlogALarge}{0.959}

\newcommand{\simAddOlogLarge}{0.805}

\newcommand{\simCells}{20}

\newcommand{\simGainLsuLsuOneMax}{0.4}
\newcommand{\simGainLsuLsuOneMin}{\minus 0.1}

\newcommand{\simGainLsuOlogALosses}{15}
\newcommand{\simGainLsuOlogAMax}{\minus 0.3}
\newcommand{\simGainLsuOlogAMin}{\minus 2.5}

\newcommand{\simGainLsuRelMax}{2.5}
\newcommand{\simGainLsuRelMin}{0.3}
\newcommand{\simGainLsuRelWins}{18}

\newcommand{\simGainLsuUtaAddSecond}{2.3}
\newcommand{\simGainLsuUtaAddSecondSE}{0.5}
\newcommand{\simGainLsuUtaAddSmall}{2.8}
\newcommand{\simGainLsuUtaAddSmallSE}{0.6}

\newcommand{\simGainLsuUtaMMWins}{15}
\newcommand{\simGainLsuUtaMax}{2.8}
\newcommand{\simGainLsuUtaMin}{\minus 0.1}

\newcommand{\simGainLsuUtaRegMax}{2.3}

\newcommand{\simGainLsuUtaRegWins}{13}

\newcommand{\simGainLsuUtaRidgeWins}{11}
\newcommand{\simGainLsuUtaWins}{13}

\newcommand{\simGainOlogAUtaWins}{18}

\newcommand{\simGainUtaRegUtaLosses}{3}
\newcommand{\simGainUtaRegUtaMax}{0.6}
\newcommand{\simGainUtaRegUtaMin}{\minus 1.8}

\newcommand{\simGainUtaRidgeUtaAddSecond}{0.6}

\newcommand{\simGainUtaRidgeUtaAddSmall}{1.0}

\newcommand{\simLossLsuOlogAMax}{2.5}
\newcommand{\simMaxHalfWidth}{0.022}
\newcommand{\simMlarge}{160}
\newcommand{\simMsecond}{40}
\newcommand{\simMsmall}{20}

\newcommand{\simNonmonLsuLarge}{0.703}

\newcommand{\simNonmonSvmLarge}{0.808}

\newcommand{\simScenariosNoFlips}{5}

\newcommand{\stabAddLsuSmall}{3.2}

\newcommand{\stabAddOlogASmall}{3.8}

\newcommand{\stabAddUtaMMSmall}{6.0}

\newcommand{\stabAddUtaRidgeSmall}{3.2}
\newcommand{\stabAddUtaSmall}{13.5}

\newcommand{\stabCellsAll}{6}
\newcommand{\stabGainLsuOlogAMax}{0.4}
\newcommand{\stabGainLsuOlogAMin}{\minus 0.7}
\newcommand{\stabGainLsuOlogASig}{3}
\newcommand{\stabGainLsuUtaMMMax}{\minus 0.1}
\newcommand{\stabGainLsuUtaMMMin}{\minus 2.8}

\newcommand{\stabGainLsuUtaMax}{\minus 0.4}
\newcommand{\stabGainLsuUtaMin}{\minus 10.4}
\newcommand{\stabGainLsuUtaRidgeAddSmall}{0.0}
\newcommand{\stabGainLsuUtaRidgeAddSmallSE}{0.0}
\newcommand{\stabGainLsuUtaRidgeMax}{0.0}
\newcommand{\stabGainLsuUtaRidgeMin}{\minus 0.4}
\newcommand{\stabGainLsuUtaRidgeQsSmall}{\minus 0.2}
\newcommand{\stabGainLsuUtaRidgeQsSmallSE}{0.1}
\newcommand{\stabGainLsuUtaRidgeSig}{5}
\newcommand{\stabGainLsuUtaSig}{6}

\newcommand{\stabMlarge}{80}
\newcommand{\stabMsmall}{20}
\newcommand{\stabQsLsuLarge}{0.2}
\newcommand{\stabQsLsuSmall}{2.5}

\newcommand{\stabQsOlogASmall}{2.5}

\newcommand{\stabQsUtaMMSmall}{3.7}
\newcommand{\stabQsUtaRidgeLarge}{0.5}
\newcommand{\stabQsUtaRidgeSmall}{2.7}
\newcommand{\stabQsUtaSmall}{5.6}

\newcommand{\sumGainLsuUtaMax}{2.8}
\newcommand{\sumGainLsuUtaMin}{1.5}

\newcommand{\timeLsuMax}{7}
\newcommand{\timeOlogAMax}{779}
\newcommand{\timeOlogAMin}{89}
\newcommand{\timeUtaMax}{33}

\newcommand{\uciMLarge}{100}
\newcommand{\uciMSecond}{50}
\newcommand{\uciMSmall}{20}
\newcommand{\uciMaxHalfWidth}{0.016}
\newcommand{\uciReps}{100}

\newcommand{\vFits}{42300}
\newcommand{\vHours}{13}
\newcommand{\vRunHours}{1.0}

\newcommand{\vaddGainHubLsuMax}{\minus 0.1}
\newcommand{\vaddGainHubLsuMin}{\minus 0.7}

\newcommand{\vaddGainHubLsuSmall}{\minus 0.7}
\newcommand{\vaddGainHubLsuSmallSE}{0.3}

\newcommand{\vaddGainHubUtaSmall}{2.1}
\newcommand{\vaddGainHubUtaSmallSE}{0.6}

\newcommand{\vfixFits}{9600}

\newcommand{\vfixFlipsHubQuarterLsuMax}{14.6}
\newcommand{\vfixFlipsHubQuarterLsuMin}{2.6}

\newcommand{\vfixFlipsHubWholeLsuMax}{3.7}
\newcommand{\vfixFlipsHubWholeLsuMin}{0.1}

\newcommand{\vfixFlipsUtaRidgeLsuMax}{17.8}
\newcommand{\vfixFlipsUtaRidgeLsuMin}{4.6}

\newcommand{\vflipsGainHubLsuMax}{9.4}
\newcommand{\vflipsGainHubLsuMin}{4.8}

\newcommand{\vflipsGainHubOlogAMax}{2.9}
\newcommand{\vflipsGainHubOlogAMin}{1.5}

\newcommand{\vflipsGainHubUtaMax}{0.4}
\newcommand{\vflipsGainHubUtaMin}{\minus 0.3}

\newcommand{\vflipsGainLsuAtLsuMax}{\minus 1.3}
\newcommand{\vflipsGainLsuAtLsuMin}{\minus 3.7}

\newcommand{\vflipsGainUtaAtUtaMax}{\minus 2.3}
\newcommand{\vflipsGainUtaAtUtaMin}{\minus 3.5}

\newcommand{\vflipsMaeGainLsuAtLsuMax}{3.1}
\newcommand{\vflipsMaeGainLsuAtLsuMin}{1.2}

\newcommand{\vflipsMaeGainUtaAtUtaMax}{2.8}
\newcommand{\vflipsMaeGainUtaAtUtaMin}{2.3}

\newcommand{\violNegKnownMax}{100}
\newcommand{\violNegKnownMin}{18}
\newcommand{\violUnorderedMax}{100}
\newcommand{\violUnorderedMin}{87}
\newcommand{\vmisCells}{16}
\newcommand{\vmisGainHubLsuLosses}{5}

\newcommand{\vmisGainHubLsuMax}{0.5}
\newcommand{\vmisGainHubLsuMin}{\minus 1.0}
\newcommand{\vmisGainHubLsuWins}{1}

\newcommand{\vopenmlGainHubLsuWins}{6}
\newcommand{\vopenmlGainHubLsuWinsBH}{6}

\newcommand{\vopenmlGainHubUtaMax}{1.3}
\newcommand{\vopenmlGainHubUtaMin}{\minus 0.4}

\newcommand{\vopenmlGainLsuAtLsuWins}{1}

\newcommand{\vopenmlGainUtaAtUtaMax}{1.3}
\newcommand{\vopenmlGainUtaAtUtaMin}{\minus 0.2}
\newcommand{\vopenmlGainUtaAtUtaWins}{7}
\newcommand{\vopenmlGainUtaAtUtaWinsBH}{6}
\newcommand{\vopenmlMaeGainLsuAtLsuLosses}{2}

\newcommand{\vopenmlMaeGainUtaAtUtaLosses}{7}

\newcommand{\vqsGainHubLsuLosses}{2}

\newcommand{\vqsGainHubLsuMax}{0.0}
\newcommand{\vqsGainHubLsuMin}{\minus 0.6}

\newcommand{\vqsGainHubLsuSmall}{\minus 0.4}
\newcommand{\vqsGainHubLsuSmallSE}{0.2}

\newcommand{\vqsGainHubUtaSmall}{2.0}
\newcommand{\vqsGainHubUtaSmallSE}{0.7}

\newcommand{\vshapeAddGainHubLsuLosses}{3}

\newcommand{\vshapeAddGainHubLsuMax}{\minus 0.1}
\newcommand{\vshapeAddGainHubLsuMin}{\minus 1.0}

\newcommand{\vshapeAddGainHubUtaMax}{2.5}
\newcommand{\vshapeAddGainHubUtaMin}{0.5}

\newcommand{\vshapeAddGainLsuChebMin}{\minus 2.5}

\newcommand{\vshapeAddGainLsuOlogAMin}{\minus 3.4}

\newcommand{\vshapeAddGainLsuUtaMMSixSmall}{0.4}
\newcommand{\vshapeAddGainLsuUtaMMSixSmallSE}{0.5}

\newcommand{\vshapeAddGainLsuUtaMMWins}{2}

\newcommand{\vshapeAddGainLsuUtaMax}{3.0}
\newcommand{\vshapeAddGainLsuUtaMin}{0.9}

\newcommand{\vshapeFlipsGainHubLsuMax}{12.4}
\newcommand{\vshapeFlipsGainHubLsuMin}{2.7}

\newcommand{\vshapeFlipsGainHubUtaLosses}{1}

\newcommand{\vshapeFlipsGainHubUtaMax}{0.6}
\newcommand{\vshapeFlipsGainHubUtaMin}{\minus 2.0}

\newcommand{\vshapeFlipsGainLsuAtLsuLosses}{5}

\newcommand{\vshapeFlipsGainLsuOlogAMin}{\minus 6.9}

\newcommand{\vshapeFlipsGainLsuUtaMin}{\minus 11.8}

\newcommand{\vshapeFlipsGainUtaAtUtaLosses}{6}

\newcommand{\vshapeSixAddGainLsuChebLosses}{2}

\newcommand{\vshapeSixAddGainLsuOlogALosses}{3}

\newcommand{\vshapeSixFlipsGainLsuOlogALosses}{3}

\newcommand{\vshapeSixGainMMRidgeSmall}{0.5}
\newcommand{\vshapeSixGainMMRidgeSmallSE}{0.6}
\newcommand{\vshapeThreeAddGainLsuChebLosses}{1}

\newcommand{\vshapeThreeAddGainLsuOlogALosses}{1}

\newcommand{\vshapeThreeFlipsGainLsuOlogALosses}{3}

\newcommand{\vuciCells}{9}
\newcommand{\vuciCevCounts}{1210, 384, 69, 65}

\newcommand{\vuciCevLsuSmall}{0.724}

\newcommand{\vuciCevOlogASmall}{0.755}

\newcommand{\vuciCpuCounts}{51, 53, 52, 53}

\newcommand{\vuciCpuLsuSmall}{0.590}

\newcommand{\vuciCpuOlogASmall}{0.611}

\newcommand{\vuciGainHubLsuMax}{0.4}
\newcommand{\vuciGainHubLsuMin}{\minus 1.2}

\newcommand{\vuciGainLsuAtLsuMax}{0.6}
\newcommand{\vuciGainLsuAtLsuMin}{\minus 0.4}

\newcommand{\vuciGainLsuAtLsuWinsBH}{0}

\newcommand{\vuciGainLsuOlogAMax}{0.4}
\newcommand{\vuciGainLsuOlogAMin}{\minus 3.1}

\newcommand{\vuciGainLsuUtaCevSmall}{0.5}
\newcommand{\vuciGainLsuUtaCevSmallSE}{0.5}
\newcommand{\vuciGainLsuUtaCpuSmall}{1.4}
\newcommand{\vuciGainLsuUtaCpuSmallSE}{0.5}
\newcommand{\vuciGainLsuUtaLosses}{1}

\newcommand{\vuciGainLsuUtaMMLosses}{1}

\newcommand{\vuciGainLsuUtaMMWins}{4}

\newcommand{\vuciGainLsuUtaMax}{1.6}

\newcommand{\vuciGainLsuUtaMpgSmall}{0.8}
\newcommand{\vuciGainLsuUtaMpgSmallSE}{0.5}

\newcommand{\vuciGainLsuUtaWins}{5}
\newcommand{\vuciGainLsuUtaWinsBH}{5}

\newcommand{\vuciGainUtaAtUtaMax}{0.5}
\newcommand{\vuciGainUtaAtUtaMin}{\minus 0.2}

\newcommand{\vuciGainUtaAtUtaWins}{1}
\newcommand{\vuciGainUtaAtUtaWinsBH}{1}

\newcommand{\vuciMpgCounts}{92, 104, 93, 103}

\newcommand{\vuciMpgLsuSmall}{0.708}

\newcommand{\vuciMpgOlogASmall}{0.718}

\begin{document}

\begin{center}
{\Large On the choice of loss function in preference disaggregation for multiple criteria sorting with small reference sets\par}
\vspace{1.2em}
{\large Yuqun Zhou$^{1}$, Yuxi Piao$^{2}$, Fangyuan Shi$^{3}$ and Jiufan Wang$^{4}$}\par
\vspace{0.4em}
{\itshape $^{1}$Department of Industrial and Systems Engineering, University of Wisconsin--Madison, Madison, WI, USA}\par
{\itshape $^{2}$The Ohio State University, Columbus, OH, USA}\par
{\itshape $^{3}$Georgia Institute of Technology, Atlanta, GA, USA}\par
{\itshape $^{4}$City University of New York, New York, NY, USA}\par
\vspace{0.4em}
Corresponding author: Yuqun Zhou, \texttt{pkzyq1994@gmail.com}. ORCID: 0009-0003-6665-4525\par
\vspace{0.4em}
Running title: Loss functions in value-based sorting\par
\vspace{0.4em}
Acknowledgments: This research received no specific grant from any funding agency in the public, commercial or not-for-profit sectors.\par
\end{center}

\begin{abstract}
Given examples of assignments to categories, UTADIS (UTilit\'es Additives DIScriminantes) derives an additive value function and cut-off points, minimising total absolute underachievement. We consider instead minimising total squared underachievement, or maximising a proportional odds likelihood, for an equivalent value function. A margin is specified from each side of all cut-off points for both UTADIS and the problem with squares. A further variant adds ridge penalties to sums of absolute underachievement. For QS university rankings, 4 other published examples and 6 sets of simulated data, least squares is on average \qsGainLsuUtaSmall{} points better than UTADIS using \qsMsmall{} references for QS (5 criteria, 4 categories), and up to about \simGainLsuUtaMax{} points better in simulations, without labelling errors. There is no improvement with substantial labelling error, and it is up to \flipsLossMax{} points worse if 10\% of observations are misclassified by 2 or more categories. However, for QS and the additive simulation, whose rules lie within the model family, similar performance is achievable without squaring, using the Chebyshev centre, with little effect from tuning. Adding ridge terms improves the absolute formulation to near best performance, but generally only for the smallest sizes. Tuned, proportional odds is significantly more accurate than least squares over \ologMainWinCells{} out of \ologMainCells{} cases, less accurate in \ologMainLossCells{} case. Squaring shortfalls only up to a fixed amount, otherwise taking them absolutely, is within \hubMainLossMax{} accuracy points of least squares without mislabelling, and \hubMainFlipsWithin{} of UTADIS with main-simulation mislabelling. Imposing penalties at all cut-offs reduces performance with mislabelling, but improves accuracy with absolute shortfalls against all published sets of judgements. Some results for additional data sets are given. Properties of the sum of squares objective indicate when it approximates least-squares fitting of category centres; this has a simple solution which is inappropriate here, and little improvement is possible with constraints.
\end{abstract}


\noindent Table~\ref{tab:notation} collects the notation.

\clearpage
\begin{table}[!tp]
\scriptsize
\renewcommand{\arraystretch}{0.66}
\caption{List of symbols used. As indicated, lower case bold denotes a vector.}\label{tab:notation}
\begin{tabular}{@{}>{\raggedright\arraybackslash}p{0.27\textwidth}p{0.68\textwidth}@{}}
\toprule
Symbol & Meaning \\
\midrule
$n$, $g_j$, $[\alpha_j,\beta_j]$ & number of criteria; criterion $j$; its scale \\
$\gamma_j$, $\gamma$, $\xi_j^l$, $N$ & number of sub-intervals of $g_j$; its value when it is the same for all criteria; the breakpoints of $g_j$; $N=\sum_j\gamma_j$ \\
$\phi_{jl}(a)$, $\ph(a)$, $\Ph$ & coverage share of sub-interval $l$ of $g_j$; stacked vector; $m\times N$ matrix of the reference set \\
$\vv$, $v_{jl}$, $\Delta_N$ & vector of value increments; its components; unit simplex of $\R^N$ \\
$u_j$, $U(a)$, $\omega_j$ & marginal value function; comprehensive value $\ph(a)\T\vv$; trade-off weight of $g_j$ \\
$q$, $C_k$, $\prec$, $b_k$, $\boldsymbol b$ & number of categories; category $k$; the preference order of the categories; limits $0=b_0<\dots<b_q=1$; vector of limits \\
$\cw$, $\pi_k$, $\Delta_q^\varepsilon$, $\varepsilon$ & category widths; $\pi_k=b_k-b_{k-1}$; widths not below $\varepsilon$ that sum to one; smallest width \\
$m$, $a_i$, $y_i$, $y$, $m_k$ & number of reference alternatives; alternative $i$; its category; the vector of these categories; number of them in $C_k$ \\
$Z^-$, $Z^+$, $\bar Z$; $T^-$, $T^+$ & maps from widths to the lower limit, upper limit and centre of the category of each reference alternative (rows $Z^-_i$, $Z^+_i$, $\bar Z_i$); maps from widths to the lower and upper zone ends \\
$\cc$, $c_i$, $h_i$; $c$, $h$ & vector of centres; centre and half-width of the category of $a_i$; of a generic interval (Lemma~\ref{lem:identity}) \\
$\delta$, $d$, $\tau$ & absolute margin; the same as a fraction of the mean half-width, $\delta=d/(2q)$; factor such that $1-\tau$ is the relative margin \\
$\underline t_i$, $\overline t_i$, $\ell_i$ & lower and upper end of the target zone of $a_i$; its zone loss \\
$\operatorname{Hu}_\zeta$, $\zeta$ & Huber function \eqref{eq:huber}; its knot \\
$\sigma_i^-$, $\sigma_i^+$ & UTADIS shortfalls of $a_i$ from the lower and upper limit of its category \\
$\rho$, $\eta$, $\mathcal M_0$ & ridge weights on $\vv$ and on $\cw-q^{-1}\one$; set of minimisers of the criterion without ridge terms \\
$(\mathrm K)$, $(\mathrm J)$; $(\mathrm K^{\mathrm c})$, $(\mathrm J^{\mathrm c})$ & the problems with known and with unknown limits; their centred cases ($\tau=0$, $\delta=0$) \\
$\tht$, $\vartheta$, $\Theta$, $\Psi$, $\Psi^{\mathrm c}$ & parameter $(\vv,\cw)$; a generic point $(\vv',\cw')$; feasible set; objective of $(\mathrm J)$; its centred case \\
$G$, $F$, $E$; $\Ph_F$, $G_F$ & Gram matrix $\Ph\T\Ph+\rho I$; face (positive increments); widths above $\varepsilon$; the columns of $\Ph$ in $F$ and $\Ph_F\T\Ph_F+\rho I$ \\
\bottomrule
\end{tabular}
\end{table}

\begin{table}[!tp]
\scriptsize
\renewcommand{\arraystretch}{0.66}
\begin{tabular}{@{}>{\raggedright\arraybackslash}p{0.27\textwidth}p{0.68\textwidth}@{}}
\multicolumn{2}{@{}l}{Table~\ref{tab:notation}, continued}\\
\toprule
Symbol & Meaning \\
\midrule
$W_F$, $\ww_F$; $W$, $\ww$; $\mu$, $\mu_F$ & face matrix and vector of \eqref{eq:Ww}, and the same for $F=\{1,\dots,N\}$; multiplier of the normalisation, and its value on the face $F$ \\
$\Pi_F$, $S$, $\rr$, $\tilde\rr_E$, $\nu$ & quantities of the Schur-complement solution \eqref{eq:schur}--\eqref{eq:kstar}, with $\tilde\rr_E=\rr_E-\varepsilon S_{E,E^c}\one$ \\
$\tht^\star$, $\vv^\star$, $\cw^\star$, $b^\star_k$; $\vv^\circ$, $\cw^\circ$ & solution of $(\mathrm J)$ or of a subproblem, and its parts; solutions without sign or width constraints \\
$\tht^\dagger$, $\vv^\dagger$, $\cw^\dagger$, $b^\dagger_k$ & a parameter that reproduces all reference assignments (Proposition~\ref{prop:compatible}, Example~\ref{ex:collapse}) \\
$V$, $\Ph_V$, $T_V$, $\boldsymbol a_V$ & set of the violated zone ends; their rows of coverage shares, maps from widths to the ends, constants $\pm\delta$ \\
$\mathcal B(\tht)$, $\mathcal A(\tht)$, $Q(\cdot\,;\tht)$ & alternatives below and above their zone; quadratic model \eqref{eq:Q} \\
$\tht^t$, $\hat\tht^t$, $\chi_t$, $\boldsymbol p^t$ & iterate $t$ of Algorithm~\ref{alg:rls} ($\tht^0$ its start); minimiser of the quadratic model, step length and direction \\
$Q_t$, $H_t$, $L$, $\varsigma$ & the model $Q(\cdot\,;\tht^t)$; its Hessian; a bound on $\|H_t\|$ and Lipschitz constant of $\nabla\Psi$; constant of strong convexity $2\min\{\rho,\eta\}$ \\
$P_\Theta$, $\Psi_0$, $R$, $\boldsymbol\lambda$ & projection onto $\Theta$; criterion without ridge terms; the ridge terms $\|\vv\|^2+\|\cw-q^{-1}\one\|^2$; multipliers of the sign constraints \\
$\tilde\vv$ & coefficients $G^{-1}\Ph\T\cc$ of the ridge regression of the centres on $\Ph$ \\
$z_i$, $\hat k(z;\boldsymbol b)$, $n_z$, $\Lambda_k(x)$, $e_k(x',x)$ & fitted value of $a_i$; category given to value $z$; number of distinct fitted values; value function and stage cost of the recalibration recursion, in which $x_k$ is the number of distinct fitted values below $b_k$ and $z_{(1)}<\dots<z_{(n_z)}$ are those values in increasing order \\
$\one$, $\zero$, $I$, $\upsilon$, $\operatorname{diag}$ & vector of ones; vector of zeros; identity matrix; a generic real number; diagonal matrix with the given diagonal \\
$(\cdot)_+$, $\R^N_+$, $\operatorname{supp}$, $\preceq$ & positive part $\max\{\cdot,0\}$; non-negative orthant; set of non-zero components; Loewner order of symmetric matrices \\
$A^{c}$, $|A|$, $\boldsymbol x_A$, $M_{A,B}$ & complement of an index set $A$ (as $F^c$, $E^c$); its number of elements; the components of a vector $\boldsymbol x$ in $A$ (as $\vv_F$, $\cw_E$, $\rr_E$, $\boldsymbol\lambda_F$); the block of a matrix $M$ with rows in $A$ and columns in $B$ (as $S_{E,E^c}$, $\Ph_{V,F}$; $S_A=S_{A,A}$) \\
$L_1$, $L_p$ & distance or norm of order $1$ and of order $p$ (not the constant $L$) \\
$b^\circ$, $\mathrm E$, $\Pr$ & unconstrained population solution of centred least squares for the limit (Proposition~\ref{prop:population}); expectation; probability \\
$f(b)$, $\mathcal V$, $\tht^{\infty}$ & population value of $\Psi^{\mathrm c}$ as a function of the limit $b$ (Example~\ref{ex:collapse}, Proposition~\ref{prop:population}); a neighbourhood of $\tht^\star$ and a limit point of the iterates (proof of Theorem~\ref{thm:conv}) \\
$\tht_\rho$, $\tht_0$, $\bar\tht$, $\tht^\ast$ & in Proposition~\ref{prop:vanishing} and its proof: minimiser of $\Psi_0+\rho R$, the solution of $(\mathrm J)$; the element of $\mathcal M_0$ nearest to $(N^{-1}\one,q^{-1}\one)$; a point of $\mathcal M_0$; a limit point of $\tht_\rho$ as $\rho\to0$ \\
$\rho'$, $\eta'$, $F_\zeta$, $f$, $\tht_\zeta$ & in Proposition~\ref{prop:huber} and its proof: ridge weights of the limiting absolute criterion; the criterion with Huber shortfalls; that absolute criterion (its minimiser is again written $\tht^\ast$); the minimiser with Huber shortfalls \\
$\mathbf 1\{\cdot\}$ & indicator: 1 if the condition holds, 0 otherwise (proof of Proposition~\ref{prop:population}) \\
\bottomrule
\end{tabular}
\end{table}

\clearpage
\section{Introduction}
\label{sec:intro}

Given a number of criteria, multiple criteria sorting \citep[or rating;][]{colorni2021rating} sorts alternatives into a number of ordered categories. It is widely used: to assess the default risk of individuals and companies \citep{angilella2015sme,doumpos2005replicating,doumpos2001creditrisk,doumpos2011bankrating,zopounidis1999utadis,belahcene2023part1,belahcene2024tour}, to classify potential suppliers \citep{segura2017supplier}, chemical reactions according to the dangers of the process \citep{kadzinski2018nanomaterials}, heart patients into classes requiring different intensities of observation \citep{doumpos2016heart}, and the modes of component failures according to risk \citep{li2026costsensitive} to give only a few examples; many others are collected by \citet{benamor2023review}. In these problems, some or all of the quantities needed to specify a model to sort alternatives are not known to, or otherwise supplied by, the decision maker. A reference set of alternatives is instead sorted by him, and the required quantities are estimated from this \citep{doumpos2011review,jacquet2001twenty,siskos2016uta}. Various techniques aim to minimise the required number of examples, which are provided individually \citep{kadzinski2021active,koksalan2009interactive,li2025incremental,ozpeynirci2018interactive}. The degree of determination of the parameters will obviously vary with the size of the reference set \citep{kadzinski2017expressiveness,vetschera2010robustness}, and various aspects of choosing one from several admissible models, given a sufficiently small reference set, have been studied \citep{beuthe2001comparative,doumpos2014robust,greco2011representative,wojcik2023representative}.

If the model consists of comparing an additive value function with limits, UTADIS (UTilit\'es Additives DIScriminantes) is the basic procedure. It solves a linear programme to minimise total shortfall from some margin within each reference alternative's assigned category; other criteria maximise total distance from the category limits or minimise misclassifications \citep{doumpos2004pairwise,kadzinski2025expressiveness,zopounidis1999utadis,zopounidis2002review}. Various authors add a regularisation term to the absolute, a hinge or a pairwise loss function \citep{doumpos2007regularized,doumpos2007additive,liu2019regularization,liu2020framework}, or include 0-1 variables to choose the relevant criteria, giving a mixed integer programme \citep{zhou2025criteria}. Alternatively, all admissible value functions may be retained, in robust \citep{greco2008ror,greco2010sorting} or stochastic \citep{kadzinski2013stochastic} ordinal regression. In this case various procedures yield one representative value function, deleting and sometimes correcting inconsistent examples if applicable \citep{doumpos2014robust,greco2011representative,mousseau2003inconsistencies,wojcik2023representative,zhang2025lexicographic}. Various procedures outside the context of decision aiding learn an unknown score and thresholds delimiting ordered categories, using a variety of convex losses including the square of a hinge function \citep{chu2007svor,gutierrez2016survey,mccullagh1980,pedregosa2017consistency}. Proportional odds models have been extended to include an additive \citep{hastie1987proportional} or more generally monotone function of explanatory variables \citep{espinosa2019constrained}. The squared hinge loss may be used to penalise deviation from upper and lower bounds of a prescribed interval \citep{hocking2013interval}. Additive models have been fitted by least squares to various problems involving ordered discrete responses, and in particular to derive utilities from conjoint experiments \citep{kruskal1965monanova,deleeuw1976additive,wittink1981alternative}.

It seems natural that methods from both areas, decision aiding and statistics, should be applicable, each approaching it in its own way; \citet{eriskin2021preference} gives a broad overview of both and suggests combining them. We consider a simple case of such an amalgam, applying methods from each area to learn the same value model.

There are many published comparisons of various criteria. They fall broadly into a number of classes, which are described briefly in Section~\ref{sec:literature} with references, and collected together in Table~\ref{tab:priorcomp}. Several criteria have been applied and compared by simulation to learn other types of preference model from rankings: least squares, linear programme, logit and monotone regression in the case of an additive model in conjoint analysis \citep{wittink1981alternative}. Various linear programmes, non-linear programmes and statistics-based procedures have likewise been compared to solve the problem of discriminating between two groups \citep{freed1986evaluating,joachimsthaler1988four,stam1989lp} and $L_p$ criteria, including the cases $p=1,2$, discussed above, reviewed \citep{stam1997lp}. UTADIS has also been compared with procedures based on a different model: logit, probit and discriminant analysis \citep{zopounidis1999failure}, majority rule and a log-likelihood for a Choquet integral \citep{sobrie2019mrsort}. A number of variants use the absolute deviation, but differ in associated aspects, for example various choices when constructing the UTADIS model \citep{doumpos2004experimental} and various methods of selecting a single model, once a set has been determined \citep{beuthe2001comparative,doumpos2014robust,wojcik2023representative}. Finally, the criterion has been varied along with another feature or features of the method: model, in the context of interval regression \citep{nguyen2025interval}; model and criterion, to allow for time-dependent criteria \citep{li2025temporal}; regularisation term and classification rule, in addition to using a sum of squares within categories \citep{liu2021interacting}; type of answer obtained, comparing various forms of Bayesian and probabilistic ordinal regression with linear and non-linear programmes \citep{ru2023probabilistic}.

Those appearing most relevant here are as follows. \citet{liu2021interacting} report better classification using their sum of squares criterion with regularisation than UTADIS, applying them to problems generating monotone classifications, and attribute this to regularisation. \citet{ru2023probabilistic} likewise find better classification with their likelihood-based methods, with reference sets drawn from various rankings of \citRuAltMin{} to \citRuAltMax{} alternatives. In neither case, however, is the effect of the loss function isolated. That is, respectively, all aspects of the loss functions, regularisation and method of allocating alternatives to categories vary over the comparisons; and comparing models, that is, probability distributions over all admissible parameter values, to functions taking a single value. We aim here to isolate its effect, for the two cases of interest. Both loss functions are applied, with a common rule for the margin and range of values tried, to learn the value model. We also introduce methods including regularisation as in the sum-of-squares case and various rules to select one model from a set of admissible ones, when applied to the absolute deviations, so allowing the consequences of each feature to be separated. For completeness we include one procedure using a likelihood to derive a single value function.

The sum of squares has been less widely applied. As noted above, it corresponds to one case of the general class considered by \citet{pedregosa2017consistency}; can be used if there are no constraints on the scores, or if there are upper and lower bounds to be estimated as in interval regression \citep{hocking2013interval}. It has been applied within decision aiding to estimate a Choquet integral, minimising sums of squares of differences of estimated scores and values prescribed to separate the categories \citep{grabisch2008kappalab,meyer2005sorting}. If restricted to the additive case this amounts to estimating linear value functions and implicit category limits by this method. Of course other forms of value function could be input. One recent example using regularisation estimates linear utilities, allowing each variable to have an interval of possible coefficients, to match intervals corresponding to the categories to which examples are assigned. The category limits are pre-assigned and slacks added to meet them; only the regularisation term consists of sums of squares \citep{sun2026interval}.

There are a number of features of the sum-of-squares loss function of potential interest. Firstly, it leads to sums of squares over those alternatives not satisfying the required margin. Thus if it is known which these are, estimation reduces to a ridge regression. This may take various forms, including regression to category mid-points, as discussed below. Secondly, it thus inherently includes small regularisation terms ensuring a unique solution, compared to a linear programme which may have an infinity of answers, in the absence of some additional rule. Similar terms may be introduced into the problem with absolute deviations, allowing their effect to be separately assessed, allowing for their relatively greater impact in that case. Thirdly, regression of the assigned category mid-point would appear a simple method to derive a value model. There are a number of analogous procedures: least-squares estimation of capacities given scores \citep{grabisch2008kappalab}; concentration around the mean of all values assigned to one category as one element of an overall sum of squares criterion \citep{liu2021interacting}; various procedures regressing an ordered response \citep{zhang2016lssvor}; selection of one value within an interval of acceptable values \citep{cheng2023interval}; minimisation by linear programme of distances of estimated scores from means for each group, here with absolute values \citep{lam1996groupmean}. Regression of a discrete ordered variable on explanatory variables has been criticised in favour of probit and logit regression since at least \citet{winship1984ordinal}. The procedure would give a unique, explicit value function, similarly to restricted least squares \citep{liew1976inequality}. However, no property of monotonicity, or of increasing limits, would be imposed; and as in all cases of regression on midpoints or similar summary figures of grouped observations \citep{stewart1983grouped}, there would be no reason to expect the true underlying function, that is, one consistent with all observations, to be recovered.

More generally, we consider a family of loss functions, equal to the distance, or its square, from some interval within the category to which the alternative is assigned. The bounds of the interval are at a fixed distance $\delta$ from each limit, plus a fraction of the width of the category. Thus least-squares UTADIS corresponds to $\delta$ only; there would also be a version with a proportion of the width only, and regression of category centres, if the interval was reduced to a point. All value functions considered are monotone increasing in each variable, and all limits increasing. The main results are:

\begin{enumerate}[leftmargin=*]
\item On a range of data sets (the QS World University Ranking and 4 published data sets) and simulated examples, using the same convention for the allowable distance from limits and trying a similar range, allowing re-estimation of the limits, least squares gives an improvement in accuracy of between \sumGainLsuUtaMin{} and \sumGainLsuUtaMax{} over UTADIS, with \qsMsmall{} and \qsMsecond{} examples. This applies to the QS (Quacquarelli Symonds) data and to one set of simulations where the sorting rule is additive, hence exactly representable by the model, with 4 or 5 criteria and 4 categories; it decreases for increasing number of examples. No improvement is apparent with the 4 published data sets of judgments by human assessors. For the simulation of substantial mislabeling (more than one category), UTADIS with optimum margin gives better accuracy; adding maximisation of margins, that is solving a second programme given the estimates of the value function, does not. Adding regularisation \citep{doumpos2007regularized} gives results \magn{\regVsUtaMin} worse to \regVsUtaMax{} better than UTADIS, significantly so in \regVsUtaLosses{} out of \regVsUtaCells{} cases in favour of UTADIS and in \regVsUtaWins{} in its own favour. All of the above relates, of course, to our implementations and the particular data sets, none with a reference set assigned by a real decision maker.
\item \emph{To investigate causes of this difference.} If the true rule of classification can be represented by the model, there will generally be many solutions with few examples. Hence some central solution should perform well, without any sum of squares term. In particular, we find no significant loss of accuracy from choosing the Chebyshev centre, comparing similar numbers of examples, for the QS data set and for the additive simulation, with similar tuning. There are, however, \cenCand{} candidates to be tuned compared to 24, and hence some of this difference could be due to less exhaustive tuning; with no tuning differences are small. The good performance of central points of the set of admissible functions is thus in line with that of a variety of methods of this type reported by \citet{wojcik2023representative}. All centres they consider outperform most other methods on at least some measures, although the analytic centre gave best overall results and was consistently better than the Chebyshev. There seems no obvious reason for this discrepancy. Similarly, for classification, the analytic centre of all consistent classifiers gave better results than the centre of the maximum-radius inscribed sphere \citep{trafalis2002analytic}. All our centres are zero-margin and lie in the subspace defined by the equalities (Section~\ref{sec:design}), and this may explain some of the discrepancy. The novelty here is comparison with least squares and regularised absolute values, with the same examples and tuning procedure. As regularisation tends to 0 there is a limit, namely closest to all increments equal and all widths equal (Proposition~\ref{prop:vanishing}), so regularisation imposes a further choice of solution. It recovers much, but not all, of the improvement due to sum of squares in some instances, only for the smallest numbers of examples; for greater numbers sum of squares always does better, although the difference varies with amount of regularisation. Hence we hesitate to ascribe any fixed amount of the difference to the loss function. In the main experiments, maximum margin performs worse, and the analytic centre, untuned, poorly, for the smallest reference sets. Least squares is less affected than the absolute-shortfall methods by deletion of one example, with fixed parameters, significantly so compared to UTADIS and its variant with maximum margin for all cases and, to a lesser extent and in fewer cases, than absolute values with regularisation. Allowing a margin proportional to width makes no difference.
\item \emph{Comparison of the procedure deriving from use of an ordinal likelihood with the above.} There is no minimum distance to be achieved from any limit, no requirement that category widths should be positive, but a multiplier on all distances, and regularisation. It has \cenCand{} candidates to tune compared to 24. So the comparison here is less direct. Over all cells in the main set of experiments it outperforms least squares, significantly, for \ologMainWinCells{} of the \ologMainCells{} cells, and is significantly inferior only when $m=\qsLsuOlogAWinSizes$ for QS. It is superior to UTADIS apart from substantial mislabelling. With no tuning and correct labelling rule it performs similarly to least squares. It performs better with no tuning in the additive simulation, for the 2 smallest numbers of examples, but not on QS, where it was significantly better after tuning. One might conjecture that tuning over more possible solutions works against least squares, but we have done no experiments to investigate this possibility. The estimated limits may extend beyond 0-1 after rescaling, with some of the data sets including mislabeling. We do not consider various types of regularisation to penalise differences between pairs, nor probabilistic approaches.
\item \emph{Other criteria, and other data sets.} If each individual shortfall is squared up to some maximum value, and taken as its absolute value above this \citep{huber1964robust}, the resulting loss functions perform no worse than \hubConsistentLossMax{} point from LS-UTADIS if the labelling is according to some rule compatible with the models considered, and generally performs as well as UTADIS, no worse than \hubFlipsLossMax{} points (\hubMainFlipsLossMax{} in the main simulation), if the mislabelling is very great, against losses of up to \flipsLossMax{} points, and \shapeFlipsLossMax{} with 6 criteria and 5 categories, for LS-UTADIS. Requiring positive margins from all other cut-offs \citep{chu2007svor,pedregosa2017consistency} does not improve performance if correctly labelled, but deteriorates it, for either loss function, under very great mislabelling. With absolute values, it performs better than UTADIS against all 4 published data sets consisting of judgements, significantly for \vopenmlGainUtaAtUtaWins{} out of \openmlCells{} and not worse for any. For 3 other published sets of data, one of which is labelled by a monotone rule not representable by any value model considered here with category widths of at least 0.01, LS-UTADIS performs much better, significantly so for \vuciGainLsuUtaWins{} of \vuciCells, and worse only for \vuciGainLsuUtaLosses. It outperforms UTADIS, by up to \shapeAddGainMax{} points, if correctly labelled, with 3 criteria, 3 categories, or 6 criteria, 5 categories. Proportional odds applied to an additive value function does not perform significantly worse, for any of these data sets.
\end{enumerate}

Sections~\ref{sec:model}--\ref{sec:solutions} give properties of sums of squares to be used to interpret and discuss the results. They are straightforward applications of known results on least-squares and related problems. Each member of this class of problem reduces to centred least squares over the set of examples violating the margin; closed forms may be given for all points on the relevant face of the polyhedron of solutions, and a Newton method, with finite number of iterations if a general position condition is satisfied, gives solutions with a measure of lack of optimality. If the allowed margin is greater than half a category width, so the 2 bounds are inverted, the sum of squares tends to pull all values for that category to its centre; this does not arise with absolute values. This occurs with some frequency, as described in Section~\ref{sec:experiments}. Regression of centres, for which an explicit formula exists, may not be non-negative everywhere or have increasing limits, and so may not define an acceptable value model; nor would it in general give 0 loss for the true generating model. In the special case of one criterion and 2 categories, the limiting value as $m\to\infty$ depends on the mean value and the proportion exceeding the cut-off, and on nothing else about the distribution of values (Proposition~\ref{prop:population}). Subject to suitable restrictions to ensure this, in our experiments it classifies poorly. Sufficiently small multiples of the ridge terms added to the sum of absolute shortfalls will yield, as solution, one closest to equal widths and increments, among those minimising the sum; and hence the closest compatible one, if compatible. With squares, there will always be some positive loss, unless $(N^{-1}\one,q^{-1}\one)$ itself gives zero loss (Proposition~\ref{prop:vanishing}). It can be seen from Proposition~\ref{prop:huber} that Huber's function yields the LS-UTADIS answer as long as all its shortfalls fall below the knot; the answer varies continuously with the knot and tends to an answer for UTADIS with ridge terms as the knot tends to 0, provided that the coefficients of the ridge terms also tend to 0; and adding penalties for all cut-offs changes the answer only in respect of alternatives more than 1 category away from their assigned positions. Hence regression of centres does not seem a sensible way to derive a sorting function.

So some aspects of the behaviour observed are explained, particularly properties of regression of centres. However, nothing more definitive can be said about the beneficial effect of some procedures selecting central answers, except that they include regularisation; and nothing at all about use of a likelihood. It follows immediately that the value of parameters is determined only by those examples outside their allowed intervals, which applies equally to absolute deviations plus regularisation. We are not aware of any suggestion in the literature of using regression of centres, but note that the availability of a simple formula might make this tempting, and there are various related procedures, as above. We make no attempt to characterise sampling properties of any of these estimates or to place confidence intervals on individual classifications. Sections~\ref{sec:model}--\ref{sec:solutions} follow the summary of relevant literature, describing the model, criteria, and properties and method of solution of the sums-of-squares problem. Experimental results are given in Section~\ref{sec:experiments}, concluded in Section~\ref{sec:conclusion}, and relevant proofs given in Appendix~\ref{app:proofs}.

\section{Related work}
\label{sec:literature}

The following areas of work seem relevant: methods to disaggregate preferences to derive value-based sorting models, and related methods originally developed for discrimination; extensions using regularisation and other techniques more usual in statistics; least-squares estimation of other aggregation functions; previous work comparing various fitting criteria; estimation of thresholds separating categories in ordered responses; least-squares estimation of additive models with ordered responses; least squares with constraints and piecewise quadratic loss functions; and other sorting models. We briefly review each area, indicating broadly the scope of results obtained. Their relationship to each other is summarised schematically in Table~\ref{tab:comparison}, and comparisons of fitting criteria previously reported in Table~\ref{tab:priorcomp}.

\paragraph{Disaggregation to derive value-based sorting models.}
A number of variants of goal programming have been applied to problems of separating two populations by a threshold \citep{freed1981goal}; various choices of norm to measure separation have been considered and discussed elsewhere \citep{stam1997lp}. An additive value function, piecewise linear in each variable, is derived by linear programming to fit overall evaluations given by a decision maker, the UTA family of procedures \citep{jacquet1982uta,jacquet2001twenty,siskos2016uta}. The possibility of a range of optimal answers has been noted from the outset, with various proposed rules to select one, compared in terms of subsequent correct classification \citep{beuthe2001comparative}. UTADIS applies similar ideas to the problem of sorting, minimising total amounts by which examples transgressed the relevant limits; it may require a minimum distance within each category or minimise number wrongly classified \citep{doumpos2004pairwise,zopounidis1999utadis,zopounidis2002review}. In one case \citep[monitoring of heart patients;][]{doumpos2016heart} the authors consider a number of variants, in order to minimise first the number of misassigned reference alternatives and then their total distance from their correct positions. So there seems scope for choice in practice. Instead of estimating one model, robust and stochastic approaches retain the whole set of value functions consistent with the information given, deriving classifications certain or possible respectively \citep{corrente2013ror,greco2008ror,greco2010sorting}, or indices of acceptability of each classification \citep{kadzinski2013stochastic}. Many extensions are possible: hierarchical structure \citep{corrente2017hierarchy}; a number of models applicable depending on circumstances \citep{kadzinski2020contingent}; some variables monotone, others free \citep{kadzinski2020partial}; more than one decision taken simultaneously \citep{kadzinski2021multidecision}; various forms, including polynomials and splines, for value functions \citep{sobrie2018utapoly}; interaction with the decision maker to supply further examples \citep{kadzinski2021active,koksalan2009interactive,ozpeynirci2018interactive} and reducing the number required when asking for comparisons of pairs \citep{huber2025bayesian,zhao2026fitradeoff}; a probabilistic procedure to sequentially estimate value function and limits, trading off cost of further information against errors of classification \citep{ulu2024bayesian} and uneven distribution of examples between categories \citep{liu2018imbalanced}. The richness and reliability of the derived models for varying amounts of information given have been investigated experimentally \citep{kadzinski2017expressiveness,vetschera2010robustness,kadzinski2025expressiveness} and a number of reviews are available \citep{belahcene2023part2,belahcene2023part1,belahcene2024tour}.

Various methods have therefore been proposed to single out one model. For example, to maximise agreement with assignments deemed certain \citep{greco2011representative}, the analytic centre for rankings \citep{bous2010acuta}, a variety of optimisation problems to find robust models \citep{doumpos2014robust} or a wide comparison of \citWojProcedures{} methods aiming to find various types of desirable model, for example maximising discrimination between categories, close to the average, central, simple or robust \citep{wojcik2023representative}. Several types of central point of the polyhedron of admissible models were generally successful in all aspects investigated, in addition to those specifically designed to be robust. As mentioned above, the best overall, and better than the Chebyshev centre for all variants considered, was the analytic centre. There are analogous results in other contexts. For a given ordering of the criteria, the mean of all admissible sets of weights was best among several options tested by \citet{barron1996decision}; and for classification, the analytic centre of all classifiers agreeing with the given training data was better than the SVM, that is, than the centre of the largest inscribed sphere, in some initial tests reported by \citet{trafalis2002analytic}. Examples may of course conflict, and it is possible to detect and remove or amend these \citep{mousseau2003inconsistencies} or allow for them as one stage of several in sequence \citep{zhang2025lexicographic}. Alternatively, a maximum margin solution, as a variant of UTADIS or included in some learning procedures, builds the requirement into the criterion. For a Choquet integral a criterion to find the minimum variance model consistent with information given, a kind of maximum entropy principle, has been proposed \citep{kojadinovic2007minvar} and is included in Kappalab as a convex quadratic programme. Sum of squares with a margin is thus similar in that there is one objective, and the regularisation terms select, as they tend to 0, the closest model to all increments equal and widths equal.

\paragraph{Other methods based on regularisation.}
\citet{doumpos2007regularized} replaced the requirement that the value function add to 1 with regularisation. That is, minimise the sum of all increments, plus a set of multiples, one for each category, of total absolute shortfall from some margin beyond upper and lower free limits. This is still a linear programme. On simulations, with \citDzRef{} reference alternatives, accuracy exceeded that of UTADIS, to an increasing extent as more error was introduced into the `true' class. Conditions on the multiples are discussed, but no recommendation is possible; many alternatives to the absolute error loss are mentioned as worthy of study.

\citet{doumpos2007additive} learned a variety of models, all additive, using a quadratic loss analogous to that for support vector machines, that is, with margins. Various methods, statistical and otherwise, of disaggregating a decision rule are compared in \citet{doumpos2011review}. Several linear pieces per criterion were allowed, not restricted to be monotone and without specified limits, in additive models learned by \citet{liu2019regularization}. Every alternative of any 1 category was required to exceed every alternative of any lower category by a margin of at least unity; amounts of shortfall from this were all penalised in proportion to their magnitude, and the first increment for each criterion and all subsequent differences were all subject to a squared penalty. Appropriate multiples were determined by cross-validation. So this is equivalent to a support vector machine, with linear logits, over all differences of pairs of reference alternatives. Any other alternative would be classified by greatest overall consistency with such differences. Comparisons by cross-validation were made against UTADIS, on \citLiuUniv{} real examples (universities) in \citLiuClasses{} classes, and various simulated sets ranging in size from \citLiuSimMin{} to \citLiuSimMax{}.

A variety of monotone additive models were fit by \citet{liu2020framework}, given a set of examples and their classes. Here shortfall occurs, linearly, between every pair of reference examples; so there is encouragement for large gaps between distinct classes, but small ones between examples in the same class. All parameters are subject to an overall squared penalty, determined by cross-validation. Upper and lower limits are free. Alternating direction multiplier method is applied. Several classes or `bands' of universities according to the QS ranking provide examples, principally for reference.

Criteria may interact, in \citet{liu2021interacting}. For a monotone, piecewise linear value function, the preprint describes learning to minimise the total squared deviation between all pairs of examples in each category, subject to specified margins between means of consecutive categories, together with a further penalty, determined by cross-validation. With free category limits, this would tend to bring all values for 1 category near their centre. Thus it has some features of our problem, except that centres of intervals rather than arbitrary points would result, dividing $[0,1]$ into a number of sub-intervals as for centred least squares. Outperforming UTADIS on all monotone problems tried, including several published data sets used in this paper, is claimed, from the preprint, for the squared error criterion; this would be expected, because of greater regularisation. Of course, many other features also differ from any version of UTADIS, namely the whole form of the loss function, any regularisation and how its constant is determined and rule for classifying further examples, in addition to large numbers of examples. We wish to investigate the effect of introducing only a penalty term, to a loss of total absolute shortfall from limits (Section~\ref{sec:experiments}).

Choice of a subset of available criteria, using 0-1 decision variables, is also described \citep{zhou2025criteria}. All the procedures of \citet{li2025temporal} allow criteria varying over time. At least one of the procedures in \citet{li2025temporal}, from reading the preprint only, involves various margins, and so appears closely related to maximising an appropriate likelihood, based on the category or categories to which each alternative could reasonably be expected. It includes linear slacks for all pairs of examples, and the limits would be found after solving. In the other, a recurrent neural net is constrained to be monotone, and maximises sum over examples of minus logs of appropriate class probabilities. Probabilities of classes are given by differences of consecutive values of standard normal distribution functions at a set of cut-off points, which are further parameters. Various comparisons on real data by cross-validation are reported.

\citet{ru2023probabilistic} describe \citRuMethods{} methods, Bayesian and non-Bayesian, to learn from examples which may all be certain as to class, or some uncertain; more is said later. One of several likelihoods based on observations of choices, rankings and so on, is that of \citet{ru2022bayesian} based on all pairwise comparisons, with an additive value. After choosing priors, the posterior distribution is sampled by Metropolis--Hastings algorithms. Similarly, \citet{liu2023contingent} assume a Dirichlet process distribution over the set of value functions and cut-offs, and so compute posterior probabilities over all admissible models given the reference set. Comparison data are represented both by a linear model and by Bradley--Terry, with Bayesian and regularised MLEs found, by \citet{grillo2025online}, who also find bounds for expected regret. Each possible category in Deck of Cards can be similarly represented by an appropriate cumulative link model, and the maximum of its posterior found exploiting suitable properties \citep{grillo2026bdor}. \citet{sun2026interval} propose to represent an unknown interval of utilities for each example, assuming only linearity and intervals for all coefficients, so as to fall within chosen intervals representing the assigned classes; a regularised quadratic programme with linear slacks for departures from these intervals. All coefficients would be penalised by sum of squares. Probabilities would be given by proportions of each interval lying within each pre-assigned interval. A data set is discussed and compared with UTA.

Allowance for uncertainty to give a robust classifier is made in another support vector machine type loss \citep{wu2023drsvm}, and hybrid models are used by \citet{guo2021hybrid} and \citet{ghaderi2021patterns}. All parameters of a number of analogous models with limits and total scores, as for UTA, UTADIS and others, are learned using the AdamW algorithm, that is, using various forms of neural network \citep{martyn2023deep}. Thus for example ANN-UTADIS learns all break-points and other parameters of monotone piecewise-linear marginals, with a UTADIS objective, that is, total shortfall of overall scores from intervals representing classes to which examples should belong. \citMkSets{} binary problems, ranging in size from \citMkAltMin{} to \citMkAltMax{} examples, with \citMkShareMin{}, \citMkShareMid{} and \citMkShareMax{} per cent to define the model, each with \citMkReps{} repetitions, were tried. Comparators included choquistic and simple logistic regressions, kernel logistic regression, simple decision rules, decision trees, UTADIS assuming \citMkSegments{} pieces per marginal, or general marginals, and \citMkMrLearners{} others learning models with a majority decision rule. All were superior to their equivalent UTADIS models except for ERA. A Choquet integral is used to form part of the linear function determining logits in \citet{tehrani2012choquistic} and extended to several classes in order by \citet{tehrani2013ordinal}. Thus it represents an ordinal likelihood, with a value function constrained to be monotone and adding to 1, reducing to a weighted sum if capacities are constrained to be additive, and so on similar terms to our various likelihoods. It is therefore analogous to the least squares fitting programme included in Kappalab.

Many authors have recently used non-monotone, generally piecewise-linear marginals and learned them, typically using substantial numbers of examples, with a variety of techniques \citep{li2025nonmonotonic}; allowed interactions, learned with an `attention' procedure, with decision according to fixed cut-off points \citep{gao2026attention}; represented examples each with several alternative profiles \citep{gao2026multiview}. Similarities and contrasts between these latter methods and others learning limits and value functions are discussed in \citet{hullermeier2024part1,hullermeier2024part2}, illustrating the diversity of methods. In particular, note absence of limits, use of cross-validation to choose a regularisation constant and sizeable numbers of examples for \citet{liu2020framework}, and absence of explicit limits, the model being parameterised by margins as in a standard SVM, for \citet{doumpos2007additive}. The robust and hybrid machines each change model and method of fitting. ANN-UTADIS of \citet{martyn2023deep} changes only the form of marginal functions and the method of fit; hence, as for \citet{liu2021interacting}, of interest but confounding the question of criterion. It is included in Table~\ref{tab:priorcomp}. None of these, however, appears to set out to investigate the best function of the criteria to use, where there are cut-offs, and so none gives any indication of this. When numbers of examples may be very small, use of differences of pairs instead of actual target values and piecewise linearity, in an additive model, was recommended in a summary, leading to regularised quadratic programs \citep{wu2026scarce}. Nothing further was given about the form of regularisation (for example, whether of fitted values, that is, as a squared loss, or of parameters) and no comparator. Each is thus quite distinct from varying only the loss function for values within classes, with monotone additive model, which we consider, sometimes with quite small numbers of examples.

\paragraph{Least squares fitting of aggregate functions.}
Several variants of least squares to determine capacities, that is, monotone and adding to 1, of Choquet integrals from assorted forms of information, have been described, for example, in \citet{grabisch2008kappalab} and implemented in Kappalab. The data may be only overall scores; or, in programs for learning to sort according to TOMASO \citep{marichal2005tomaso,meyer2005sorting}, it appears from the code that arbitrary target values for each example may be set and least squares differences of these and actual Choquet integrals minimised. Values of examples from neighbouring categories are constrained to differ by at least a constant. Minimising further over the arbitrary target values is clearly equivalent to summing distances, squared, from intervals separated as required. Thus it is analogous to sum of upper and lower shortfalls from unspecified limits. Restricting to additive capacities, it is clearly equivalent to least-squares fitting of the UTADIS straight-line marginals, without penalties or specified end points or minimum widths. So there is some justification for a least-squares program using a margin to learn an aggregation from examples. General piecewise linear marginals could be used if inputs were proportions over sub-intervals, as here. So differences consist principally of specified limits, specified end points, minimum widths, values assigned in absence of examples from some class, inclusion of ridge terms and margins. We are unaware of published comparisons with small sample sizes, which are possible with our methods of Sections~\ref{sec:model} and~\ref{sec:solutions}. Hence the special case of least squares UTADIS with only 1 sub-interval, included in our results, is fairly similar.

\paragraph{Comparison of various criteria.}
It is not surprising that there should be many comparisons of methods, particularly for discrimination. There are various papers comparing procedures within each class of method for sorting. In particular, \citet{eriskin2021preference} discusses differences, including nomenclature, between them. He compares various methods within each group; this paper is concerned only with criteria, with fixed form of model. It seems that comparison between least squares, linear programming and others would begin in this context, that is, learning utility functions to explain consumer responses. On simulated data, least squares, LINMAP \citep{srinivasan1973linmap}, logit and another monotone procedure were compared \citep{wittink1981alternative}, least squares being preferred if utilities were reasonably compensatory, and linear program if 1 factor dominated others. Since only rankings are fitted, no question of classes or limits arises. 2-group discrimination was compared for a number of linear programs, by simulation \citep[including non-linear]{freed1986evaluating,stam1989lp}. Discriminant functions, linear and quadratic, logit and a linear program were all satisfactory on normally and otherwise distributed examples \citep{joachimsthaler1988four}. A variety of choices of norm are discussed in \citet{stam1997lp}.

To learn various weighted means and other simple aggregations, \citet{beliakov2009lp} used linear programs with total absolute error, analogous to least squares; advantage with outliers seemed likely. As seen in Section~\ref{sec:sim}, some forms of criterion for UTADIS have this property, with appropriate margin, but not all which involve totals of absolute errors. Many features of fitting UTADIS were investigated for their effect on resulting classification accuracy and variability, with many repetitions \citep{doumpos2004experimental}. Several other methods were compared, with real data divided into learning and testing subsamples, by \citet{zopounidis1999failure}, and their hierarchical version with others \citep{doumpos2001creditrisk}. Methods to select one or more good solutions to the UTA problem after solution have been compared \citep{beuthe2001comparative}, and ways of obtaining an overall representative single model \citep{doumpos2014robust,wojcik2023representative}. Generalisation was little affected by choice of support vector machine \citep{koshiba2003l1l2}. Hinge losses can be shown to have rates of convergence comparable to logit but better than square loss \citep{rosasco2004loss}, though this should be distinguished from least squares on target values.

Numerous methods for fitting responses over an ordered set of categories are reviewed in \citet{gutierrez2016survey}, of which several make some use of ordering, with improved accuracy and reliability compared to fitting them as unordered. Some theoretical guarantees of good properties are shown \citep{pedregosa2017consistency} for \citPedWays{} ways of generalising to allow cut-off points, with various choices of function. A new approximation to square loss was more successful than least squares for \citPedBetter{} of \citPedData{} problems, \citPedSig{} significantly. Others, including some analogous to hinge losses, were not directly compared. These \citPedWays{} methods are \citPedWays{} of many possibilities discussed briefly in \citet{pedregosa2017consistency}, whose names they adopt.\footnote{See Rennie and Srebro, Loss functions for preference levels: regression with discrete ordered labels, IJCAI workshop, 2005. We have not included this in our references since we could find no reference to it on DBLP nor a DOI, as we require for all others; and nothing we say relies on it.} For various loss functions applicable with cutoff points, \citet{yamasaki2024remarks} show that several lead to many estimates taking only a small number of possible values. Methods to find values lying within given intervals have been compared \citep{nguyen2025interval}, as have those in \citet{li2025temporal}, which are quite different. Decision by majority of suitable rules was about as accurate as either UTADIS or choquistic regression \citep{sobrie2019mrsort}.

Ideally one would wish to compare criteria when all other features of model to be learned and decision rule are the same, of which only \citet{ru2023probabilistic} qualify. Starting from prior distributions, posteriors are sampled (MCMC) over all value functions and limits to give the observed decisions on all examples; alternatively, uncertain observations, expressed as various degrees of belief in each class, are allowed. Each was compared with suitable mathematical programs, over several published data sets. Thus a likelihood and optimization methods are contrasted, but both methods seek full distributions over all models, hence giving degrees of belief in classes of other alternatives. In contrast, all procedures here yield a unique solution from certainly classified examples. 1 arm is standard likelihood for an ordered response, giving a maximum likelihood estimate, the rest are various criteria based on shortfalls, but with common ranges of margins and ridge penalty, data and so on. Uncertain observations are separately represented as distributions; certainly known observations impose constraints, and the modal class over samples taken as decision. Over \citRuTables{} published data sets, with \citRuAltMin{}--\citRuAltMax{} observations each, classes were generated randomly, and typically \citRuShareMin{}--\citRuShareMax{} per cent selected to be known, with about the same number uncertain. Thus similar sizes of learning sample are used. No quantities were selected by cross-validation, nor any criteria involving squares tried. Both sampling procedures were superior to either optimization procedure, irrespective of amount known about classes. Hence this is not informative about squared versus absolute error, with little information, nor about simple maximisation of a standard likelihood. However, as noted, there are other differences between procedures in essentially all comparisons. All assume the same form of additive utility function and vary only the type of criterion, but with rankings; mostly only linear value function and 1 cut-off point in the discrimination studies; variations of UTADIS only, with other models in some comparisons; choice of representation from set of good solutions only varied in UTA comparisons; no normalisation constraint, large samples in many ML comparisons; learning procedure only changed in ANN-UTADIS. Thus none compare square versus absolute shortfall loss with identical other assumptions. Summary details are given in Table~\ref{tab:priorcomp}, with only summaries available to us in several cases.

\paragraph{Learning values and cut-off points.}
Standard methods to fit values and several cut-offs include proportional odds \citep{mccullagh1980} and reduction to several binary decisions \citep{frank2001simple}, and various forms of support vector machines \citep{chu2007svor}, some compared in \citet{gutierrez2016survey}. \citet{tutz2022ordinal} defines wider classes of which this is an example, with logit link. \citPedWays{} extensions allowing cut-off points were investigated theoretically in \citet{pedregosa2017consistency}. In one, only cut-offs on either side of actual class appear in some function of extent to which a margin condition is violated; several choices are discussed. All cut-off points are fitted to give a convex program. In the other, all occur, each example being charged once per cut-off point between its value and any class. Without any restrictions on values, it was shown these approximate 0-1 and absolute errors respectively, under mild conditions. So with suitable functions, margins and restriction to monotone additive value functions, these reduce exactly to UTADIS, and to least squares on limits less margin. Adding the requirement of limits in $[0,1]$, this is least squares UTADIS; using proportions of widths gives the obvious extension. Learned cut-offs should respect dominance relationships between examples \citep{bendavid1995monotone,cano2019monotonic,kotlowski2013isotonic}; rules can be used to ensure this \citep{greco2001rough,slowinski2009rough}. There is of course no reason why additive scores could not be compatible with sets of such rules. An additive model of this kind, with piecewise linear marginals, can be derived under some conditions from rough sets approximation for \citDzClasses{} classes \citep{dembczynski2006additive}, related to UTA and UTADIS and fitted by a support vector machine; margin \citDzMargin, with total linear penalty on violations, and squared sum of remaining parameters. Various forms of ordinal likelihood with additive scores, sometimes with some restrictions to ensure monotonicity, include no restrictions, other than additivity \citep{hastie1987proportional}; some factors constrained monotone, all on ordinal scales, in proportional odds model \citep{espinosa2019constrained} and monotonicity of smooth terms in many other responses \citep{pya2015scam}. Hence proportional odds with the sum over criteria of marginals of our score is a special case; any rescaling to sum to unity gives a value function with various choices for the range of limits. Many likelihoods based on such a response have been mentioned above, to learn from various types of response, and by various methods. It is used to fit the final neural net in \citet{li2025temporal}. We have included it as 1 alternative procedure, without suggesting it is desirable, since it did best after suitable tuning.

\paragraph{Additive models, least squares.}
A number of authors fitted additive functions to ordered responses, with unknown increasing function of response, starting with \citet{kruskal1965monanova} and extending to other types of independent variables and responses \citep{deleeuw1976additive,young1976regression}. Hence all dependent values would be assigned most suitable values, constrained only to preserve ordering. It follows, as above, that least squares and centred versions correspond to deviations from such best-fitting ordered values and all equal within classes respectively. Several methods of learning additive utilities to fit rankings were compared, on simulated data \citep{wittink1981alternative} and questionnaires \citep{jain1979comparison}; and fitting monotone functions seemed desirable given an unknown true function \citep{cattin1984conjoint}. Fitting by least squares would give monotonicity of all marginals \citep{bacchetti1989isotonic}. Some methods using least squares criteria have been proposed for classification into unordered categories \citep{suykens1999lssvm}, and 1 into ordered categories, attracting all to a centre value per category with cut-off points determined subsequently \citep{zhang2016lssvor}, similar to centring except that any values are allowed. None however has all features of the model here.

\paragraph{Methods of solution.}
It is straightforward to express the answer given active constraints, and there are several finite algorithms to find these (\citealp{bro1997nnls,liew1976inequality,lawson1995}; \citealp[eliminating a number of parameters]{lovell1963seasonal}; \citealp{nocedal2006}). They can also be viewed as modifications of Newton's method \citep{hintermuller2002pdas}. Answers are continuous functions of the input data but can have points of non-differentiability \citep{bemporad2002explicit}. Functions comprising a number of separate quadratics, such as squared hinge loss, can similarly be minimised, applying extended Newton methods to active terms, with finite convergence \citep{keerthi2005newton,mangasarian2002newton,sun1997piecewise}; also, using features of first-order conditions, with simple bounds \citep{li1993newton}. An insensitivity region around 0 may be introduced \citep{smola2004svr}. Similarly, predictions may be attracted to intervals, with margins and squared amounts by which they are exceeded, on both sides \citep{hocking2013interval}, of which least squares UTADIS would be a special case with correct intervals and unrestricted values. A number of theoretical results has been established in this general context of interval targets, for example, about consistency \citep{cheng2023interval}, relative performances of methods \citep{nguyen2025interval}, and various bounds \citep[sometimes uniform over all values consistent with intervals]{pukdee2025interval}. So losses of this sort, avoiding attraction to centres of intervals, are well founded, but always assuming these known and changing other features. We therefore extend only to allow all limits unknown but restricted as required. A mistake in a key equation is pointed out \citep{dolgopolik2024hessian}, though immaterial to our argument. Attraction to centres of intervals has otherwise been shown inappropriate \citep{stewart1983grouped}. We therefore rely mainly on standard results; the main novelty being to move all end-points, each common to 2 adjacent categories, and hence the whole insensitive interval, with various constraints giving rise to sub-problems.

\paragraph{Other approaches to sorting.}
As with the models considered above, it is possible to learn ELECTRE TRI and other majority rule methods given a set of examples of assignments to categories, using various mathematical programming, heuristic or satisfiability techniques \citep[see, for example,][]{belahcene2018sat,dias2002robust,doumpos2019electretri,kadzinski2022characteristic,leroy2011mrsort,minoungou2022singlepeaked,mousseau2001weights,mousseau1998inferring,sobrie2019mrsort}. The disaggregation procedure has been tested on various learning sets for one form of ELECTRE TRI \citep{madhooshiarzanagh2021electre}; an approach based on distances to centroids, with no limits, has been proposed by \citet{karasakal2021distance}; \citet{tuzcu2026noncompensatory} considers an example and concludes that compensation between criteria may be undesirable. However, we wish only to cast doubt on regression onto centres of categories, that is, on fitting these centres as functions of overall scores, and not on using distances to centroids to determine an ordering. All outranking methods involve a non-compensatory aggregation of the criteria, resulting in very combinatorial problems to solve to learn them; it therefore seems reasonable to limit ourselves to comparison with approaches yielding an overall score, using various standard classification methods \citep{breiman2001rf,cortes1995svn}.

\begin{table}[tp]
\centering
\renewcommand{\baselinestretch}{1}\selectfont
\caption{Earlier comparisons of fitting criteria and of methods that fit an ordinal assignment model. Sizes are given where the source states them.}
\label{tab:priorcomp}
\scriptsize
\setlength{\tabcolsep}{3pt}
\renewcommand{\arraystretch}{0.62}
\begin{tabular}{@{}>{\raggedright\arraybackslash}p{0.19\textwidth}>{\raggedright\arraybackslash}p{0.21\textwidth}>{\raggedright\arraybackslash}p{0.30\textwidth}>{\raggedright\arraybackslash}p{0.25\textwidth}@{}}
\toprule
Study & Model & Compared & What else changes, or is held fixed \\
\midrule
\citet{wittink1981alternative} & additive part-worths, rank-order data & least squares, linear programme, logit, monotone regression & no categories, limits or margin \\
\citet{freed1986evaluating}; \citet{joachimsthaler1988four}; \citet{stam1989lp} & linear discriminant & linear and nonlinear programming formulations; Fisher's, logistic and linear-programming discriminants & one cut-off; no marginal value functions \\
\citet{beliakov2009lp} & weighted means and related functions & least absolute deviations and least squares & given scores are fitted; no categories or limits \\
\citet{doumpos2004experimental} & additive value with limits & settings of the UTADIS development process & one method; performance and stability by simulation \\
\citet{zopounidis1999failure}; \citet{doumpos2001creditrisk} & additive value with limits, or hierarchical discrimination; statistical models & UTADIS and variants of it, or one multicriteria method; discriminant analysis, logit, probit & the statistical comparators have another model \\
\citet{beuthe2001comparative} & additive value function (ranking) & post-optimality rules of UTA & absolute loss kept; the selection among its solutions varies \\
\citet{doumpos2014robust}; \citet{wojcik2023representative} & additive value with limits & procedures that select one compatible model & loss kept; the selection varies \\
\citet{doumpos2007regularized} & additive value with limits & UTADIS; UTADIS with the normalisation replaced by a penalty on the increments & absolute loss kept; reference sets of \citDzRef{} \\
\citet{liu2019regularization} & additive value; no limits in the proposed method & UTADIS; pairwise hinge loss with a quadratic regulariser & assignment rule and tuning change; \citLiuUniv{} or more alternatives \\
\citet{liu2021interacting} & additive value; no limits in the proposed method & UTADIS; quadratic within-category criterion with a tuned regulariser & criterion, regulariser, tuning and assignment rule change together \\
\citet{sobrie2019mrsort} & majority-rule, Choquet and additive value models & a majority-rule model; choquistic regression; UTADIS & the model changes with the method \\
\citet{nguyen2025interval} & interval regression, several model families & two-sided hinge, absolute and squared; a likelihood-based model & intervals are given; the model family changes \\
\citet{martyn2023deep} & threshold-based models, among them an additive value with limits & UTADIS by linear programming and by a network with learned breakpoints; logistic-type models; rules; trees & loss of UTADIS kept; parametrisation and training change; \citMkAltMin{} to \citMkAltMax{} alternatives \\
\citet{li2025temporal} & value models for temporal criteria & margin-based quadratic programme; network trained by an ordinal likelihood & the model changes with the criterion \\
\citet{ru2023probabilistic} & additive value with limits & Bayesian and stochastic ordinal regression; programmes minimising misclassification & a distribution over models against point solutions; reference sets from \citRuAltMin{} to \citRuAltMax{} alternatives \\
This paper & additive value with limits & absolute and squared shortfall; absolute shortfall with ridge terms; Huber shortfall; all-threshold absolute and squared shortfalls; proportional-odds likelihood & shortfalls: margin, grid, recalibration and reference sets in common; likelihood: value model and reference sets in common \\
\bottomrule
\end{tabular}
\end{table}

\begin{table}[tp]
\centering
\renewcommand{\baselinestretch}{1}\selectfont
\caption{Approaches to learning an ordinal assignment model from examples. Columns: (1) returns a monotone additive value function with limits on its scale (``set'': all such functions compatible with the examples; ``distr.'': a distribution over them; ``rescaled'': limits after division by a positive constant, not confined to $[0,1]$; ``implicit limits'': separated scores of an aggregation model in place of limits; ``no limits'': a value function without category limits); (2) defined for inconsistent reference sets without a preliminary repair; (3) fit and preference for a margin from the limits in a single problem (``2nd stage'': the margin is maximised in a second problem); (4) the criterion is a sum of squared deviations or squared shortfalls; (5) how the cited source computes its solution: ``yes'', an explicit solution on the optimal face or a finite method built on such solutions; ``QP'', a quadratic programme passed to a solver; ``iter.'', an iterative method not stated to terminate finitely; ``none'', no algorithm; a dash where (4) is ``no''. Columns (4) and (5) record what the source states. Rows are works whose formulation was checked in the source or its preprint (Section~\ref{sec:literature}).}
\label{tab:comparison}
\scriptsize
\setlength{\tabcolsep}{3pt}
\renewcommand{\arraystretch}{0.74}
\begin{tabular}{@{}p{0.50\textwidth}ccccc@{}}
\toprule
Approach & (1) & (2) & (3) & (4) & (5) \\
\midrule
UTADIS programmes \citep{zopounidis1999utadis,doumpos2004pairwise} & yes & yes & yes & no & -- \\
Robust ordinal regression for sorting \citep{greco2010sorting} & set & no & no & no & -- \\
Representative models \citep{greco2011representative,wojcik2023representative} & yes & no & 2nd stage & no & -- \\
Bayesian and subjective stochastic ordinal regression for sorting \citep{ru2023probabilistic} & distr. & no & yes & no & -- \\
Pairwise regularised learning of value functions \citep{liu2019regularization,liu2020framework} & no limits & yes & no & no & -- \\
Regularised UTADIS \citep{doumpos2007regularized} & rescaled & yes & yes & no & -- \\
Interval utility regression, fixed limits \citep{sun2026interval} & no & yes & no & no & -- \\
Least-squares sorting with a Choquet capacity \citep{grabisch2008kappalab,meyer2005sorting} & implicit limits & yes & yes & yes & QP \\
Quadratic within-category criterion \citep{liu2021interacting} & no limits & yes & no & yes & QP \\
Proportional odds, linear index \citep{mccullagh1980} & no & yes & no & no & -- \\
Support vector ordinal regression \citep{chu2007svor} & no & yes & yes & no & -- \\
Proportional odds with additive or monotone effects \citep{hastie1987proportional,espinosa2019constrained} & rescaled & yes & no & no & -- \\
Additive model from rough approximations, two classes \citep{dembczynski2006additive} & yes & yes & yes & no & -- \\
Threshold surrogates with squared hinge \citep{pedregosa2017consistency} & no & yes & yes & yes & none \\
Interval regression with squared hinge, fixed ends \citep{hocking2013interval} & no & yes & yes & yes & iter. \\
Monotone least-squares scaling \citep{kruskal1965monanova,deleeuw1976additive} & no & yes & no & yes & iter. \\
Restricted least squares on category centres \citep{liew1976inequality} & no & yes & no & yes & yes \\
Least-squares UTADIS (as studied here) & yes & yes & yes & yes & yes \\
\bottomrule
\end{tabular}
\end{table}

In Table~\ref{tab:comparison} we have collected information about the type of model returned and the nature of the fitting criterion, for those approaches that it was possible to unambiguously define from the papers. The various forms described in each line of the table are all referred to in that line. All disaggregation procedures yield a sorting model, and involve solving a linear program or, more generally, a convex program. Proportional odds, with all effects assumed to be additively monotone in the criteria, involves maximum likelihood estimation, and yields a sorting model only if all rescaled limits fall between 0 and 1, as they did not always in our experiments, usually where there was considerable labelling noise (see Section~\ref{sec:design}). Fitting by least squares, with aggregation by a Choquet integral, returns an overall score assumed monotone and normalised, and limits implicit in the solution; in the particular case of an additive Choquet integral, it returns least-squares UTADIS with linear marginal value functions. The various forms of regression using surrogates for the unknown true limits all involve least squares, but none returns a sorting model. Regression on the centres of the categories is as described, but does not provide a sorting model. So least-squares UTADIS shares the features of returning a sorting model and of a squared fitting function with other entries in Table~\ref{tab:comparison}, and appears to combine several familiar elements. The main novelty of this work is hence the set of comparisons made in Section~\ref{sec:experiments}, supported by an investigation of properties of fitting by a squared function (Sections~\ref{sec:model} and~\ref{sec:solutions}).

\section{Additive models to represent sorting; further functions of \texorpdfstring{$\cw$}{pi} and \texorpdfstring{$\vv$}{v}}
\label{sec:model}

In this section we describe the class of sorting models to be fitted, and define a number of related functions to be minimised in the fitting procedure. Symbols used are listed in Table~\ref{tab:notation}.

\subsection{Model and definitions}
\label{sec:setting}

Assume there are $n$ criteria $g_1,\dots,g_n$ to all be maximised, over known ranges from $\alpha_j$ to $\beta_j$. As in UTA \citep{jacquet1982uta} and subsequent work \citep[for example,][]{siskos2016uta}, break points divide the range of each $g_j$ into sub-intervals: $\alpha_j=\xi_j^0<\xi_j^1<\dots<\xi_j^{\gamma_j}=\beta_j$, not necessarily regular, and there are in total $N=\sum_j\gamma_j$ sub-intervals. Given an alternative $a$,
\begin{equation}\label{eq:coverage}
\phi_{jl}(a)=\min\Bigl\{1,\max\Bigl\{0,\frac{g_j(a)-\xi_j^{l-1}}{\xi_j^{l}-\xi_j^{l-1}}\Bigr\}\Bigr\},\qquad l=1,\dots,\gamma_j ,
\end{equation}
the proportion of $j$'s $l$-th sub-interval which is covered by values less than $g_j(a)$. Collecting all these proportions gives the vector $\ph(a)$, of dimension $N$, all elements in $[0,1]$. Then $u_j(g_j(a))=\sum_{l}v_{jl}\phi_{jl}(a)$ defines a marginal value function, with $v_{jl}$ the additional value attributed to $g_j$ taking values throughout the $l$-th sub-interval. Overall value is
\begin{equation}\label{eq:U}
U(a)=\ph(a)\T\vv,\qquad \vv\in\Delta_N:=\{\vv\in\R^N:\vv\ge \zero,\ \one\T\vv=1\}.
\end{equation}
Non-negativity of $\vv$ ensures non-decreasing marginal value functions, required for a sorting model where all criteria are to be preferred increasing \citep{belahcene2023part1}. The sum-to-one restriction gives overall values of 0 and 1 to respectively the worst and best possible alternatives, and the sum $\omega_j=\sum_l v_{jl}$ can be regarded as the relative importance of criterion $j$. Increments, rather than slopes on each sub-interval, are used to ensure $\phi_{jl}$ in $[0,1]$ irrespective of scaling of each $g_j$. Various forms of smooth monotonic functions could of course be used \citep{pya2015scam,sobrie2018utapoly} but lose this simple interpretation of $v_{jl}$.

Assume there are $q$ categories $C_1,\dots,C_q$ in order of preference, delimited by $b_0=0<b_1<\dots<b_q=1$. Alternative $a$ is then allocated to category $C_k$ if $b_{k-1}\le U(a)<b_k$, and to $C_q$ if $U(a)=1$. Let $\pi_k$ be the width of $C_k$, that is, $b_k-b_{k-1}$. Then $b_k=\sum_{s\le k}\pi_s$. We take $\cw$ to lie in $\Delta_q^{\varepsilon}$, that is, all components at least $\varepsilon$, where $0<\varepsilon<1/q$, and sum to 1, so that all $b_k$ are strictly increasing. Requiring simply $b_k-b_{k-1}$ greater than some minimum, as in \citet{zopounidis1999utadis}, would be equivalent. Defining categories by lengths has the advantage that the centre and half-width are both simply linear functions.

A set of $m$ reference alternatives $a_i$ is assumed to have been allocated by the decision maker to categories $y_i$. Let $\ph_i=\ph(a_i)$, and let the rows of $\Ph$ be the transposes of these. Define $m\times q$ 0-1 matrices $Z^-$, $Z^+$ such that $Z^-_{ik}$ is unity if $k<y_i$ and 0 otherwise, and similarly $Z^+_{ik}$ if $k\le y_i$. Hence $Z^-_i\cw$ and $Z^+_i\cw$ are respectively the lower and upper bounds on values of alternatives falling in the same category as $a_i$. Let $c_i$ be the midpoint, $\bar Z_i\cw$, of these bounds, and $h_i$ their half-distance. We shall be concerned both with the case of given $b_k$ and with estimating these as well. The latter would normally apply, except, for example, if categories were predefined according to ranges of some score.

\subsection{Zone losses}
\label{sec:zone}

\begin{figure}[t]
\centering
\includegraphics{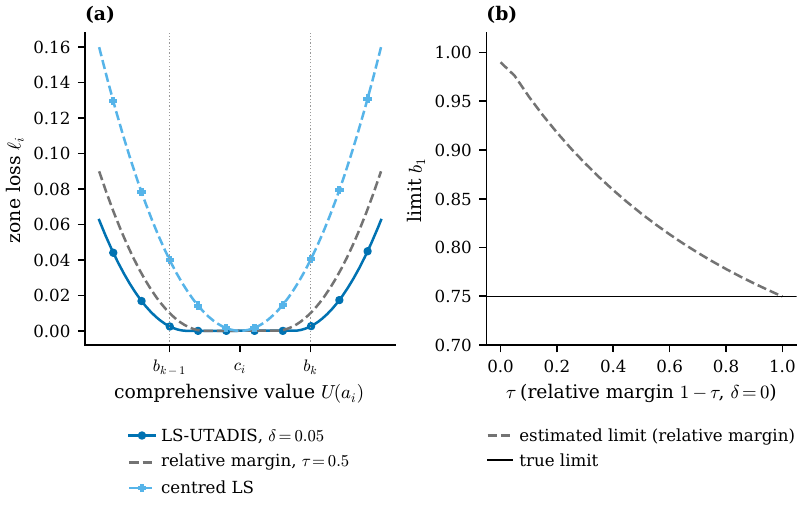}
\caption{The zone-loss family. (a) Loss of one reference alternative as a function of its comprehensive value, for the three named members: least-squares UTADIS vanishes on the category less a margin $\delta$ at each end, the relative-margin variant on the category shrunk by $\tau$, and centred least squares only at the centre. (b) Limit estimated in Example~\ref{ex:collapse} along the relative-margin path: it moves from the boundary of the scale under centring to the true value as $\tau\to1$.}
\label{fig:zone}
\end{figure}

The fitting procedure requires that the overall value of each reference alternative should differ from each of the two relevant limits by at least a fixed amount $\delta$, any less being penalised. Thus define $\sigma_i^-=\pos{b_{y_i-1}+\delta-U(a_i)}$ and $\sigma_i^+=\pos{U(a_i)-b_{y_i}+\delta}$, $(\upsilon)_+$ denoting $\max(\upsilon,0)$, and UTADIS minimises their sum \citep{doumpos2004pairwise,zopounidis1999utadis}. Equivalently, it is an immediate-thresholds model with the hinge function, as defined in \citet{pedregosa2017consistency}. We propose instead to square each term, retaining the immediate thresholds definition for each. This is natural as an extension of UTADIS, and \citet{pedregosa2017consistency} show that, if the scores can take arbitrary values, it is a consistent estimator relative to the 0-1 function; that is, it is equivalent to counting errors of allocation, which we take as a primary performance statistic and use in choosing parameters. Using all thresholds would be appropriate if the desired statistic were total distance from true category, also calculated in our experiments; it would impose a penalty at each limit between the true and allocated categories of any alternative allocated well out. Results comparing it with others are given in Section~\ref{sec:v08}. More generally, let there be allowed a minimum distance from either boundary, absolute and relative to the half-width $h_i$, so acceptable values for $a_i$ lie between
\begin{equation}\label{eq:zone}
\underline t_i(\cw)=T^-_i\cw+\delta\ \mbox{ and }\ \overline t_i(\cw)=T^+_i\cw-\delta ,
\end{equation}
where $T^-_i=\tfrac{1+\tau}{2}Z^-_i+\tfrac{1-\tau}{2}Z^+_i$ and similarly for $T^+$. These may equally be written $c_i-\tau h_i+\delta$, $c_i+\tau h_i-\delta$, and are clearly linear in $\cw$. Let
\begin{equation}\label{eq:loss}
\ell_i(\vv,\cw)=\pos{\underline t_i(\cw)-\ph_i\T\vv}^{2}+\pos{\ph_i\T\vv-\overline t_i(\cw)}^{2}.
\end{equation}
Provided $\delta<\tau h_i$, so these are in the right order, this is clearly the square of the distance of $\ph_i\T\vv$ from this interval. If $\delta>\tau h_i$, any value between the two (reversed) values is penalised relative to both, any others relative to one, and the minimum, positive, is attained at $c_i$. Hence any reference alternative with true value between these reversed values, so from a category of width less than $2\delta$, would effectively be forced to the centre by the constant $\delta$. It seems reasonable that $\delta$ should be no more than a proportion of a typical category half-width, so we write it as $d(2q)^{-1}$, $d$ being a proportion of the average category half-width, and note the frequency of reference alternatives exceeding both values in \eqref{eq:zone} for the estimated parameters. With additional terms $\rho\|\vv\|^2$ and $\eta\|\cw-q^{-1}\one\|^2$, to be minimised over $\vv$, and also over $\cw$ if these are not known, we have problems $(\mathrm K)$ and $(\mathrm J)$ respectively:
\begin{align}
(\mathrm K)\quad &\min_{\vv\in\Delta_N}\ \sum_{i=1}^m \ell_i(\vv,\cw)+\rho\|\vv\|^2 ,\label{eq:K}\\
(\mathrm J)\quad &\min_{\vv\in\Delta_N,\ \cw\in\Delta_q^\varepsilon}\ \sum_{i=1}^m \ell_i(\vv,\cw)+\rho\|\vv\|^2+\eta\|\cw-q^{-1}\one\|^2 .\label{eq:J}
\end{align}
Let $\tht=(\vv,\cw)$, $\Theta=\Delta_N\times\Delta_q^\varepsilon$ and $\Psi$ denote the whole expression to be minimised in $(\mathrm J)$. Problem $(\mathrm K)$ arises by fixing $\cw$. We name three cases:
\begin{itemize}[leftmargin=*]
\item \emph{least-squares UTADIS}, for $\tau=1$, $\delta>0$: as above, this can equally be regarded as immediate thresholds, with the square of the hinge, embedded in an overall value model. We compare this directly with UTADIS in Section~\ref{sec:experiments}.
\item \emph{relative margin}: a constant proportion $1-\tau$ of each half-category width, for $\delta=0$ and $0<\tau<1$.
\item \emph{centred least squares}, for $\delta=\tau=0$: minimise the sum over $i$ of $(\ph_i\T\vv-c_i)^2$.
\end{itemize}
These are illustrated in Figure~\ref{fig:zone}. Note that UTADIS omits one of the two terms of equation~\eqref{eq:loss} for the top and bottom categories; we retain both, to include the centred problem as a special case, and show that it makes little difference (Section~\ref{sec:experiments}). There are various reasons to prefer a square. In particular, given the limits to the acceptable region for each alternative, it leads to least-squares problems on subsets of the reference set, allowing the explicit solutions and other properties established in Section~\ref{sec:solutions}. It also allows UTADIS, centred least squares and various forms of regression on limits to be regarded as special cases of a single general form, so properties can be investigated as functions of $\tau$ and $\delta$. Penalties on the same quantities $\sigma_i^\pm$ include obviously the original absolute function for UTADIS and, with an added penalty on $\sum v_{jl}$, regularised UTADIS \citep{doumpos2007regularized}; various forms including hinge and logistic functions arise in support vector machine approaches and as models of ordered responses \citep{chu2007svor,doumpos2007additive,mccullagh1980}; and an insensitive zone of fixed width around the true value, as in support vector regression \citep{smola2004svr}. A disadvantage of squaring is that outliers, that is, reference alternatives with values very different from their true category, have undue influence; this is considered further in Section~\ref{sec:recal}. The desirable properties of threshold surrogates with a squared hinge function are shown by \citet{pedregosa2017consistency}. Various approaches penalise discrepancies between pairs of overall scores \citep[for example,][]{liu2020framework} and so do not involve the limits.

It is thus of interest to establish under what circumstances regression on centres of categories arises as a limit of other penalties.

\begin{lemma}\label{lem:identity}
If $c\pm h$ are the ends of an interval of centre $c$ and half-width $h$, then for all $\upsilon\in\R$, $(\upsilon-c+h)^2+(\upsilon-c-h)^2=2(\upsilon-c)^2+2h^2$.
\end{lemma}

Summing the squares of the distances from each bound on the true category of each reference alternative, we have
\begin{equation}\label{eq:two-limit}
\sum_i\bigl[(\ph_i\T \vv-b_{y_i-1})^2+(\ph_i\T \vv-b_{y_i})^2\bigr]=2\|\Ph \vv-\bar Z\cw\|^2+\tfrac12\sum_{k=1}^q m_k\pi_k^2 ,
\end{equation}
where $m_k$ is the number of reference alternatives allocated to category $C_k$. Hence, with known limits, minimising this expression is equivalent to centred least squares: regression of the category centres onto $\Ph$, with non-negative elements summing to 1. Without these restrictions this is standard \citep{liew1976inequality}. If the limits are unknown, the second term penalises large widths, more heavily for categories containing many reference alternatives. Unbalanced reference sets would thus be expected to have some effect on estimated widths, quite separately from the effect discussed by \citet{liu2018imbalanced}; problem $(\mathrm J)$ explicitly includes a penalty on unequal widths.

Lemma~\ref{lem:identity} applies equally to any pair of bounds in this range. In particular, for those in equation~\eqref{eq:zone},
\begin{equation}\label{eq:two-ends}
\sum_i\bigl[(\ph_i\T \vv-\underline t_i)^2+(\ph_i\T \vv-\overline t_i)^2\bigr]=2\|\Ph \vv-\bar Z\cw\|^2+\tfrac12\sum_{k=1}^q m_k(\tau\pi_k-2\delta)^2 ,
\end{equation}
reducing to \eqref{eq:two-limit} if $\tau=1$, $\delta=0$. Thus if all values, whether inside or outside the interval, were penalised by their squared distances from its 2 ends, all the variants would be equivalent to finding centres, exactly if the $b$'s are known, approximately otherwise, differing only in pulling the widths towards $2\delta/\tau$. It is therefore the one-sided penalties which distinguish them.

We next establish conditions under which the problems $(\mathrm K)$ and $(\mathrm J)$ are regular. No restriction on the size of the reference set $m$ is then necessary.

\begin{proposition}\label{prop:convex}
For all $\tau\in[0,1]$ and $\delta\ge0$, $\Psi$ is convex and continuously differentiable. Hence $(\mathrm J)$ will have at least one minimising solution, which will be unique if $\rho,\eta>0$. Similarly $(\mathrm K)$ will have a unique solution if $\rho>0$, whatever $m$ and $N$. For centred least squares there will also be a unique solution if $\rho=0$ provided $\operatorname{rank}\Ph=N$.
\end{proposition}

Convexity of various threshold functions in both the scores and thresholds has been noted \citep[for example,][]{pedregosa2017consistency}; it is of interest here that it continues to apply when thresholds are defined in terms of widths and there may be a margin depending on the widths. Uniqueness is of course desirable to ensure a single model selected, but no more than that. Denote the set of all solutions, that is, minimisers of the function without the final square terms, by $\mathcal M_0$. This will be a non-empty closed convex set. If there exists any parameter keeping all reference alternatives within the prescribed region inside their categories, then $\mathcal M_0$ consists of all such parameters, with minimum value of the fitting function 0. Otherwise there may be no such parameter even if all regions to be satisfied are proper sub-intervals of the categories: for example, two reference alternatives with identical values for all criteria, allocated to different categories, could not both satisfy a requirement of distance $\delta$ from limits. In such a case the minimum value attained would be positive.

\begin{proposition}\label{prop:vanishing}
Consider now equal coefficients $\rho=\eta$ on the additional 2 terms. Then as $\rho$ tends to 0, the solution $\tht_\rho$ of $(\mathrm J)$ tends to that point $\tht_0$ in $\mathcal M_0$ closest to $(N^{-1}\one,q^{-1}\one)$. Also:
\begin{enumerate}[label=(\roman*),leftmargin=*]
\item for absolute rather than squared shortfalls, with corresponding $\mathcal M_0$, again $\tht_\rho$ tends to $\tht_0$, indeed equals it for all $0<\rho\le\bar\rho$ for some positive $\bar\rho$;
\item for squared shortfalls, if some value of $\tht$ gives zero loss, no $\tht_\rho$ will lie in $\mathcal M_0$, unless $(N^{-1}\one,q^{-1}\one)$ lies in it, in which case this will be $\tht_\rho$ for all $\rho>0$.
\end{enumerate}
\end{proposition}

The simple arguments of Appendix~\ref{app:proofs} apply equally to using the absolute values of $\sigma_i^\pm$ with the same added terms, that is, with $\mathcal M_0$ then defined as all solutions to that problem. Thus, for the small positive values of these coefficients normally to be expected in an application, the minimiser of the whole expression would be just outside $\mathcal M_0$ since any small displacement towards $(N^{-1}\one,q^{-1}\one)$ would reduce the last 2 terms to first order but increase the original term only to second order; Proposition~\ref{prop:vanishing}(ii) covers this case. Clearly this argument breaks down if the original terms are taken absolutely; in this case minimisation over $\mathcal M_0$ of the remaining 2 terms is subject only to linear equalities and inequalities. Hence there exist associated Lagrange multipliers and minimisation over all of $\Theta$, for small enough positive values of the coefficients, gives a point in $\mathcal M_0$ \citep{friedlander2007exact}. In particular, if any point gives all zeros for the original terms, the 2 problems have the same set of minimisers $\mathcal M_0$, so have common limits as the coefficients tend to 0, but the absolute version attains this limit, while the squared one gives points tending to it, with 1 or more small positive terms. As its name implies, the second vector in the limit of Proposition~\ref{prop:vanishing} corresponds to all categories of equal width. The first corresponds to equal values of $v_{jl}$ for all $j$ and $l$: equal increments, implying greater overall weight attached to any criterion with more intervals. So these terms provide some regularisation, and resolve any ambiguity in selection of a solution, analogous to various methods of selecting, for example, most discriminant solutions from a linear program \citep{beuthe2001comparative,greco2011representative,wojcik2023representative}, but no choice of level of regularisation is involved, in contrast to \citet{doumpos2007regularized} and \citet{liu2019regularization}. We take both to be $10^{-4}$ except in one experiment (Section~\ref{sec:fixed}). As the sum of losses increases linearly with $m$ but the regularisation terms do not, the latter will become relatively less important as $m$ increases. We take $\varepsilon=0.01$, so not only avoiding empty categories but ruling out any category of less than 1\% width; the effect of this choice is considered in Section~\ref{sec:margin}.

The following summarises the effect of the two parameters defining the required distance from the limits.

\begin{proposition}\label{prop:family}
Taking $\rho=\eta=0$:
\begin{enumerate}[label=(\roman*),leftmargin=*]
\item if $\tau=\delta=0$, $\Psi(\tht)$ is the sum of squares of deviations of the overall values of reference alternatives from their category centres;
\item if $\tau=1$, $\Psi(\tht)$ is the sum of squares of all $\sigma_i^\pm$, and so is 0 iff all reference alternatives are within their categories, at least $\delta$ from both limits. If $\delta=0$, this means correct allocation of all, except possibly some on true limits;
\item if $\delta=0$ and $\tau<1$, $\ell_i=0$ iff the overall value of $a_i$ is at least $(1-\tau)h_i$ from each limit delimiting its category;
\item $\ell_i$, and hence the minima of $(\mathrm K)$ and $(\mathrm J)$, are non-increasing in $\tau$ and non-decreasing in $\delta$.
\end{enumerate}
\end{proposition}

So all variants require all reference alternatives to be strictly inside their allocated categories. Maximising distances from limits, and finding most discriminant and central representatives, similarly \citep{doumpos2014robust,greco2011representative,wojcik2023representative,zopounidis2002review}. The difference is in the requirement imposed if a category is narrow: constant distance for all, or constant proportion of the width. Our experiments indicate a preference for the former.

\subsection{Huber and all-threshold shortfalls}
\label{sec:huber}

We next describe 2 variants of LS-UTADIS, also compared empirically in Section~\ref{sec:v08}. Both thus involve an additive value model, margins from the bounds on each assigned category and 2 ridge terms. They differ from it, respectively, in applying a different function to each positive distance, or in taking positive distances from all, not just 2, bounds. Let $\operatorname{Hu}_\zeta$ denote Huber's \citeyearpar{huber1964robust} function, truncated to non-negative arguments, that is,
\begin{equation}\label{eq:huber}
\operatorname{Hu}_\zeta(s)=\begin{cases}s^2,&0\le s\le\zeta,\\ 2\zeta s-\zeta^2,&s>\zeta.\end{cases}
\end{equation}
If each term $(\sigma_i^\pm)^2$ in LS-UTADIS is replaced by $\operatorname{Hu}_\zeta(\sigma_i^\pm)$, distances less than $\zeta$ contribute as before, but greater ones only linearly, so that an alternative very far from its category exerts only a fixed amount of influence on $\tht$. With $\zeta$ equal to $\tfrac14$, $\tfrac12$ and 1 times the average width$/2$, $=1/(2q)$, we refer to the resulting problems as Huber shortfalls. If the margin is 1 and $\zeta=2$, this is the function proposed by \citet{zhang2004sgd}, a smooth approximation to a hinge loss function. For $\zeta=1$ this is double the smoothed hinge function used by \citet{rennie2005fast} for all cutoff points, in their model for ordered data, learning the positions of all cutoff points; the knot would then coincide with the margin. This corresponds to $d=0.25$ and $0.5$ in our parameterisation, since $\zeta$ takes the value $\delta$. In general, however, we wish the knot to depend on the average width of the intervals into which the scale is divided, and so to be independent of the margin.

Alternatively, each alternative may be required to achieve margins from all internal cut-offs. Thus $a_i$ would additionally have terms
\[
\pos{b_k+\delta-U(a_i)},\ \ k=1,\dots,y_i-2;\quad \pos{U(a_i)-b_k+\delta},\ \ k=y_i+1,\dots,q-1.
\]
It is shown in \citet{pedregosa2017consistency} that, with suitable functions, minimising these is consistent for the mean absolute error of assignment, just as the original is for 0-1 error. They are used, in the form of constraints, by \citet{chu2007svor}. We consider these with squared terms plus 2 ridge terms, and as a linear program with absolute values.

\begin{proposition}\label{prop:huber}
\begin{enumerate}[label=(\roman*),leftmargin=*]
\item All these problems are convex and differentiable, except the last, a linear program, and have unique minima over $\Theta$ if $\rho$ and $\eta$ are positive.
\item The minimiser is a continuous function of $\zeta$. It equals that for LS-UTADIS if none of the latter's terms exceeds $\zeta$, in particular if $\zeta$ is at least $1+\delta$. If the coefficients of the ridge terms are $2\zeta\rho'$ and $2\zeta\eta'$, it tends, as $\zeta$ tends to 0, to that for UTADIS with coefficients of the ridge terms $\rho'$ and $\eta'$, within $(2m\zeta/\min(\rho',\eta'))^{1/2}$. If instead $\rho$ and $\eta$ are fixed positive constants, the limit is $(N^{-1}\one,q^{-1}\one)$.
\item There will be additional positive terms for $a_i$ compared to LS-UTADIS if and only if it falls, ignoring margins, outside the range of categories adjacent to its own, that is, if and only if $U(a_i)<b_{y_i-2}+\delta$ or $U(a_i)>b_{y_i+1}-\delta$.
\end{enumerate}
\end{proposition}

These problems thus reduce to those of LS-UTADIS and UTADIS respectively if there are only 2 categories. Hence we would expect little variation from the answer to LS-UTADIS until $\zeta$ drops below the maximum of its positive terms, after which continuous variation, tending to an answer with ridge terms for UTADIS only if these coefficients decrease to 0. Our coefficients do not decrease, being fixed at $10^{-4}$, so no conclusion can be drawn from the above bound, for the sizes of $\zeta$ considered. Obviously only very badly misallocated alternatives will give any difference between these and the corresponding immediate-threshold criteria. It would be straightforward to extend Algorithm~\ref{alg:rls} to the problem with Huber shortfalls, adding appropriate linear terms, but we have preferred to use general conic programming solvers, as described in Section~\ref{sec:design}.

\section{Nature of the solutions}
\label{sec:solutions}

We next obtain an explicit expression for the solution of centred least squares, show that any other member of the above family can be similarly solved on an appropriate subset of the reference set, and describe an algorithm. For simplicity we write $(\mathrm K^{\mathrm c})$, $(\mathrm J^{\mathrm c})$ and $\Psi^{\mathrm c}$ for the corresponding quantities in the centred case. All proofs are given in Appendix~\ref{app:proofs}.

\subsection{Centred least squares}
\label{sec:centred}

Write $G=\Ph\T\Ph+\rho I$ and suppose:

\begin{assumption}\label{ass:rank}
Either $\rho>0$ or $\operatorname{rank}\Ph=N$, so $m\ge N$.
\end{assumption}

Given any subset $F$ of the set of indices $1,\dots,N$ of elements of $\vv$, denote by $\Ph_F$ the sub-matrix of corresponding columns of $\Ph$, $G_F$ the corresponding sub-matrix of $G$, and define
\begin{equation}\label{eq:Ww}
W_F=G_F^{-1}-\frac{G_F^{-1}\one\one\T G_F^{-1}}{\one\T G_F^{-1}\one},\qquad
\ww_F=\frac{G_F^{-1}\one}{\one\T G_F^{-1}\one}.
\end{equation}
Then, with known limits, the solution can be written down on the set $F$ of positive elements.

\begin{theorem}\label{thm:known}
Under Assumption~\ref{ass:rank}, there is a unique solution $\vv^\star$ of $(\mathrm K^{\mathrm c})$. Where $F$ is the set of indices of positive components of $\vv^\star$, and with $\cc=\bar Z\cw$,
\begin{equation}\label{eq:vstar}
\vv^\star_F=W_F\Ph_F\T \cc+\ww_F,\qquad \vv^\star_{F^{c}}=\zero .
\end{equation}
For any subset $F$, the corresponding $\vv$ given by \eqref{eq:vstar} and zeros elsewhere will be the solution iff
\begin{equation}\label{eq:cert}
\vv_F\ge\zero\qquad\text{and}\qquad \bigl(G\vv-\Ph\T\cc\bigr)_j\ \ge\ \mu_F\quad\text{for all } j\notin F,
\end{equation}
where $\mu_F=(1-\one\T G_F^{-1}\Ph_F\T \cc)/(\one\T G_F^{-1}\one)$. The solution is therefore a continuous, piecewise linear function of $\cc$.
\end{theorem}

This is simply the well-known solution and conditions for least squares subject to elements summing to a constant and non-negative \citep{lawson1995,nocedal2006}. Note in particular the simple expression for the solution once $F$ is known, and the condition allowing easy checking whether a given $F$ is correct. $F$ will generally be found by adding and dropping single variables, as in standard active-set methods, and will be found exactly in a finite number of iterations.

\begin{corollary}\label{cor:free}
Denote by $\vv^{\circ}$ the solution allowing negative elements, that is, equation~\eqref{eq:vstar} with $F$ the full index set. Then $\vv^\circ$ will be the required solution $\vv^\star$ iff $\vv^\circ\ge\zero$. Otherwise $\vv^\star$ will include some zero elements, and the minimum value attained will exceed that for the problem without the positivity requirement.
\end{corollary}

This may alternatively be written $\vv^\circ=\tilde\vv+(1-\one\T\tilde\vv)\,\ww$, where $\tilde\vv=G^{-1}\Ph\T\cc$ are the coefficients on regression of $\cc$ on $\Ph$ (that is, least squares if $\rho=0$), and $\ww$ as in \eqref{eq:Ww} with $F$ all indices. So $\vv^\circ$ is simply the least restricted solution, shifted in the direction $\ww$ to ensure that all increments sum to 1. It follows from Corollary~\ref{cor:free} when this simpler expression can be applied, and Section~\ref{sec:experiments} indicates how frequently not.

If the limits are unknown, similar expressions result after eliminating $\vv$.

\begin{theorem}\label{thm:unknown}
If $\rho,\eta$ are positive there will be a unique solution $(\vv^\star,\cw^\star)$ to $(\mathrm J^{\mathrm c})$. Let $F$, $E$ denote the sets of indices of positive elements of $\vv^\star$ and $\cw^\star-\varepsilon\one$ respectively. With
\begin{equation}\label{eq:schur}
\Pi_F=\Ph_FW_F\Ph_F\T,\qquad S=\bar Z\T(I-\Pi_F)\bar Z+\eta I,\qquad \rr=\bar Z\T\Ph_F\ww_F+\tfrac{\eta}{q}\one ,
\end{equation}
it is easy to check that 0 is less than or equal to $\Pi_F$, which in turn is less than or equal to $I$, and $S$ greater than or equal to $\eta I$, in the Loewner order, and that
\begin{equation}\label{eq:kstar}
\begin{aligned}
\cw^\star_E&=S_E^{-1}\bigl(\tilde \rr_E+\nu\one\bigr),&\qquad
\nu&=\frac{1-\varepsilon|E^c|-\one\T S_E^{-1}\tilde \rr_E}{\one\T S_E^{-1}\one},\\
\cw^\star_{E^c}&=\varepsilon\one,&\qquad
\vv^\star_F&=W_F\Ph_F\T \bar Z\cw^\star+\ww_F ,
\end{aligned}
\end{equation}
where $\tilde \rr_E=\rr_E-\varepsilon S_{E,E^c}\one$.
\end{theorem}

The form of the elimination is well known \citep{lovell1963seasonal}. $S$ may be regarded as the appropriate Schur complement after eliminating $\vv$, and $(I-\Pi_F)$ subtracts the part of $\cc$ reproducible by some overall value function with support $F$. Of more interest here are the inequalities on the eigenvalues of these two matrices, valid for all $F$, and the following.

\begin{corollary}\label{cor:ordered}
For all $\tau$ and $\delta$ the $b_k$ resulting from the solution of $(\mathrm J)$ will differ by at least $\varepsilon$. Correspondingly defined $\cw^\circ$ with $E$ and $F$ all indices, that is, $\cw^\circ=S^{-1}(\rr+\nu\one)$, need not have all positive elements, and hence may give $b$'s that are not increasing.
\end{corollary}

\subsection{Other members of the family}
\label{sec:general}

Define, given any $\tht\in\Theta$, $\mathcal B(\tht)$ to be the set of reference alternatives with values less than or equal to the lower value \eqref{eq:zone} of the acceptable region, and $\mathcal A(\tht)$ those exceeding the upper value. Then, given any other $\vartheta=(\vv',\cw')$, define
\begin{equation}\label{eq:Q}
\begin{aligned}
Q(\vartheta;\tht)={}&\sum_{i\in\mathcal B(\tht)}\bigl(\ph_i\T \vv'-\underline t_i(\cw')\bigr)^2+\sum_{i\in\mathcal A(\tht)}\bigl(\ph_i\T \vv'-\overline t_i(\cw')\bigr)^2\\
&+\rho\|\vv'\|^2+\eta\|\cw'-q^{-1}\one\|^2 .
\end{aligned}
\end{equation}
This is thus similar to the centred problem, except applied only to those reference alternatives exceeding one or other limit, towards that limit, and with targets affine functions of $\cw$, involving rows of $T^-$ and $T^+$ and constant terms $\pm\delta$. If the two values \eqref{eq:zone} are reversed, an alternative may exceed both and appear in both sums; if not, it may exceed neither and appear in neither. Clearly, for the centred problem, all reference alternatives will appear in exactly one sum, and so $Q(\vartheta;\tht)=\Psi^{\mathrm c}(\vartheta)$.

\begin{theorem}\label{thm:fixed}
For all $\rho,\eta>0$, $\tau\in[0,1]$ and $\delta\ge0$, $\tht^\star$ is the unique solution of $(\mathrm J)$ iff it minimises $Q(\vartheta;\tht^\star)$.
\end{theorem}

It follows that an expression analogous to Theorem~\ref{thm:unknown} gives the solution:

\begin{corollary}\label{cor:general}
Given a solution $\tht^\star=(\vv^\star,\cw^\star)$ to $(\mathrm J)$, with positive values of $\rho$ and $\eta$, form the matrices $\Ph_V$ and $T_V$, and 1-vector $\boldsymbol a_V$, from rows indexed by all $i$ in $\mathcal B(\tht^\star)$, followed by those in $\mathcal A(\tht^\star)$, consisting of $\ph_i\T$, $T^-_i$, $T^+_i$ and $+\delta$, $-\delta$. As before let $F$ and $E$ denote the sets of positive elements of $\vv^\star$ and $\cw^\star-\varepsilon\one$, and define $W_F$, $\ww_F$ from \eqref{eq:Ww} with $G_F=\Ph_{V,F}\T\Ph_{V,F}+\rho I$, and similarly $\Pi_F$. Then \eqref{eq:kstar} applies, with
\begin{align*}
S&=T_V\T(I-\Pi_F)T_V+\eta I,\\
\rr&=T_V\T\Ph_{V,F}\ww_F-T_V\T(I-\Pi_F)\boldsymbol a_V+\tfrac{\eta}{q}\one,
\end{align*}
and $\vv^\star_F=W_F\Ph_{V,F}\T(T_V\cw^\star+\boldsymbol a_V)+\ww_F$.
\end{corollary}

Of course this is not directly calculable since the relevant sets of rows are unknown a priori. For problems involving a squared hinge function it is standard to obtain a finite number of iterations each involving a Newton step, that is, solving the relevant least-squares problem \citep{keerthi2005newton,mangasarian2002newton}. Theorem~\ref{thm:fixed} indicates the applicability of such an approach here, except that the least-squares problem involves inequalities as well as the sum constraint, solved as in Theorems~\ref{thm:known} and~\ref{thm:unknown}. Hence the following procedure (Algorithm~\ref{alg:rls}) to determine the relevant subset of reference alternatives.

\begin{algorithm}[t]
\caption{Re-partitioned least squares (RLS) for $(\mathrm J)$}\label{alg:rls}
\begin{algorithmic}[1]
\Require $\Ph$, $y$, $\tau$, $\delta$, $\rho,\eta>0$, $\varepsilon$
\State $\tht^0\gets(N^{-1}\one,\ q^{-1}\one)$
\For{$t=0,1,2,\dots$}
  \State determine $\mathcal B(\tht^t)$ and $\mathcal A(\tht^t)$
  \State $\hat\tht^t\gets\argmin_{\vartheta\in\Theta}Q(\vartheta;\tht^t)$ \Comment{explicit on its face (Theorem~\ref{thm:unknown}); face by active-set search}
  \If{$\hat\tht^t=\tht^t$} \Return $\tht^t$ \EndIf
  \State $\chi_t\gets\argmin_{\chi\in[0,1]}\Psi\bigl(\tht^t+\chi(\hat\tht^t-\tht^t)\bigr)$ \Comment{root of a non-decreasing piecewise-linear function}
  \State $\tht^{t+1}\gets\tht^t+\chi_t(\hat\tht^t-\tht^t)$
\EndFor
\end{algorithmic}
\end{algorithm}

\begin{assumption}\label{ass:nondeg}
For no $i$ does the solution $\tht^\star$ of $(\mathrm J)$ take a value exactly equal to either of the limits \eqref{eq:zone}.
\end{assumption}

The following shows that this procedure must converge.

\begin{theorem}\label{thm:conv}
For $\rho,\eta>0$, denote by $\tht^t$ the $t$-th iterate of Algorithm~\ref{alg:rls}, $\tht^\star$ the solution of $(\mathrm J)$. Then
\begin{enumerate}[label=(\roman*),leftmargin=*]
\item $\Psi(\tht^{t+1})<\Psi(\tht^t)$, unless $\tht^t=\tht^\star$;
\item $\tht^t$ converges to $\tht^\star$;
\item subject to Assumption~\ref{ass:nondeg}, terminates at $\tht^\star$ after a finite number of steps. In particular, one step suffices for the centred problem.
\end{enumerate}
\end{theorem}

This follows a result of \citet{sun1997piecewise} on convergence of a similar algorithm, without inequality constraints, to the unique global solution for a broad class of functions. Extension to include the present inequality constraints, handled by active-set steps \citep{hintermuller2002pdas}, is proved in Appendix~\ref{app:proofs}, with Assumption~\ref{ass:nondeg} necessary only to ensure termination. The main virtues of the algorithm are that it highlights the least-squares nature of each step and provides a measure of lack of optimality of each solution. It is not significantly faster than simply solving the linear program for UTADIS (Section~\ref{sec:time}).

\subsection{Disadvantages of centring}
\label{sec:cost}

The following shows that centring will tend to give a stable estimate, but will not give zero loss even for a correct set of reference allocations.

\begin{proposition}\label{prop:compatible}
Suppose there exists $\tht^\dagger$ such that all reference alternatives fall within their allocated categories, and let $\rho=\eta=0$. Then:
\begin{enumerate}[label=(\roman*),leftmargin=*]
\item if $\tau=1$ and $\delta=0$, $\Psi(\tht^\dagger)=0$, as will be any solution of the minimisation of $\Psi$. If $\delta>0$, $\Psi(\tht^\dagger)=0$ iff all reference alternatives exceed both relevant limits of their categories by at least $\delta$.
\item if $\tau=\delta=0$, $\Psi^{\mathrm c}(\tht^\dagger)$ will equal the sum over $i$ of squared distances of the overall values from their category centres, so will be non-zero unless all are exactly at these centres.
\end{enumerate}
\end{proposition}

Of course $\tht^\dagger$ may still minimise the centred problem, and give the minimum value, despite the positive value. For instance, if the values of the reference alternatives in Example~\ref{ex:collapse} below were uniformly distributed over $[0,1]$, with true limit $\tfrac12$, the expected value of the criterion would have derivative $(2b-1)/4$ and be minimised at the true value. But no assurance can be given that the minimum will converge to $\tht^\dagger$ as $m$ increases; this is analogous to results on the inconsistency of least-squares fitting to mid-points \citep{stewart1983grouped}, and to more general results on regression where observations are known only to lie in intervals \citep{cheng2023interval}. Here of course the intervals depend on the unknown parameters. Example~\ref{ex:collapse} shows a considerable discrepancy can result. It is possible to derive explicit expressions for some simple cases in the limit of infinite numbers of observations.

\begin{proposition}\label{prop:population}
Suppose a single criterion defined on $[0,1]$, with 1 sub-interval, so $U(a)=g_1(a)$, 2 categories with true cutoff $b^\dagger$, and reference alternatives independently drawn, correctly allocated, from a distribution of $g_1(a)$ over its range of possible values. For any 1 reference alternative, mean centred loss as a function of cutoff point $b$ would therefore be minimized, over the range $\varepsilon$ to $1-\varepsilon$, as closely to $b^\circ=2\,\mathrm E[g_1(a)]-\Pr(g_1(a)\ge b^\dagger)$ as possible. If $b^\dagger$ lies between $\varepsilon$ and $1-\varepsilon$ there will thus be agreement with the true value in the limit only if $b^\dagger=b^\circ$. The distribution of values affects this only via its mean value and proportion exceeding $b^\dagger$.
\end{proposition}

\begin{example}\label{ex:collapse}
Let there be one criterion over range $[0,1]$ divided into one sub-interval only, two categories with true limit $\tfrac34$, and assume $\varepsilon<\tfrac14$ so the true solution is admissible. Let values of the reference set be uniformly distributed over $[\tfrac12,1]$, correctly allocated. Writing $\cw=(b,1-b)$, the expected value of the centred loss function would be
\[
f(b)=\int_{1/2}^{3/4}\Bigl(\upsilon-\frac b2\Bigr)^2d\upsilon+\int_{3/4}^{1}\Bigl(\upsilon-\frac{1+b}2\Bigr)^2d\upsilon,\qquad f'(b)=\frac{b-1}{4},
\]
so decreasing throughout, giving $b^\star=1-\varepsilon$. Thus almost all alternatives would be allocated to $C_1$, with error $\tfrac12-2\varepsilon$. This is the asymptotic result as $m\to\infty$; for any finite $m$ there would be some variation about this. Clearly the reason is the very uneven distribution of members of $C_1$ over its range. As expected, with $\tau=1$, $\delta=0$ the function is zero at $b=\tfrac34$, which is admissible for all such $\varepsilon$, and so very different from the centred estimate.
\end{example}

Applying these expressions to the situation of Example~\ref{ex:collapse}, $\mathrm E[g_1(a)]$ would be $\tfrac34$ and $\Pr(g_1(a)\ge\tfrac34)$ would be $\tfrac12$, so $b^\circ$ would equal 1, giving estimated limit $1-\varepsilon$. Note that the corresponding $f$ would be an integral over length $\tfrac12$, with uniform density 1, and so would give half the mean loss of Proposition~\ref{prop:population}. If all values were equally likely over $[0,1]$ and the true limit $\tfrac12$, then $b^\circ$ would equal $\tfrac12$, giving agreement as in the earlier example.

Varying $\tau$ thus provides a continuous range of estimates, as shown in Figure~\ref{fig:zone}. An absolute margin would not include any element of centring.

We next consider the effect of incorrectly labelled reference alternatives.

\subsection{Estimation of limits}
\label{sec:recal}

Clearly with a squared function a very erroneous reference alternative may pull the estimated limits considerably. So if labelling is suspected to be unreliable, we propose to re-estimate the limits, given the estimated value function, to minimise total distance of reference alternatives from their true categories. That is,
\begin{equation}\label{eq:recal}
\min_{0\le b_1\le\dots\le b_{q-1}}\ \sum_{i=1}^m\bigl|\hat k(z_i;\boldsymbol b)-y_i\bigr|,\qquad z_i=\ph_i\T\vv^\star ,
\end{equation}
where $\hat k(z_i;\boldsymbol b)$ denotes the category containing value $z_i$, given the limits. Any monotone non-decreasing sequence of categories corresponding to the ordered set of values $z_i$ can be represented by suitable limits, so this is simply isotonic regression, with absolute error, of the categories onto the values \citep{stout2013isotonic} and a special case of monotone classification \citep{kotlowski2013isotonic}.

\begin{proposition}\label{prop:recal}
Order the set of distinct values taken by the $z_i$ as $z_{(1)}<\dots<z_{(n_z)}$. Let $e_k(x',x)$ denote $\sum|k-y_i|$ over all $i$ with $z_i$ taking one of values $z_{(x'+1)},\dots,z_{(x)}$. Then recursively defining
\begin{align*}
\Lambda_0(0)&=0,\qquad \Lambda_0(x)=\infty\quad(x\ge1),\\
\Lambda_k(x)&=\min_{0\le x'\le x}\bigl\{\Lambda_{k-1}(x')+e_k(x',x)\bigr\},\qquad k=1,\dots,q,
\end{align*}
the minimum value of \eqref{eq:recal} is $\Lambda_q(n_z)$, attained by any limits separating consecutive values of $z$ corresponding to the relevant minimising values of $x'$.
\end{proposition}

This standard dynamic programming solution \citep{bellman1961segments} to division into a given number of contiguous groups is $O(qn_z^2)$ and trivial in running time for reasonable sizes of reference set, with the advantage of directly giving suitable limits. There are more efficient solutions to isotonic regression \citep{stout2013isotonic}. We choose to minimise total distance rather than simply number wrongly allocated in order to take account of ordering of categories. Obviously it would be up to cross-validation to decide whether to use recalibrated limits or not. For comparison we also apply this procedure after fitting UTADIS. It thus resembles procedures to find representative models in involving two stages, while otherwise being a single convex program. Obviously the properties guaranteed of the two sets of parameters differ: in particular, the recalibrated values of $\boldsymbol b$ may be less than all values, or greater than all values, taken by reference alternatives, and so some categories never occur, or two or more consecutive equal, again giving unused categories. The overall result would thus be allocation according to a number of cut-off points applied to the estimated value function, and so outside the model of normalised overall value. Results are presented with and without recalibration.

\section{Empirical results}
\label{sec:experiments}

The issues investigated are: the effect on accuracies of using squared rather than absolute shortfalls, other features being unchanged, and how much is attributable to the ridges; whether there will be less variation according to reference set; the penalty from squaring if mislabellings extend over two or more categories; superiority, if any, of the margin as a fraction of the width, over a fixed amount; extent of degradation from centring; comparison of each penalty with the appropriate likelihood, again based on an additive value; and whether Huber or all-threshold shortfalls (Section~\ref{sec:huber}), or further data, change any of these conclusions.

\subsection{Design}
\label{sec:design}

\paragraph{Data.}
QS rankings of universities for 2026 provide 5 scores for each university, published as weighted scores, adding to the total. There are \qsN{} with complete information. Ranks 1--100, 101--300, 301--600, 601 and lower, give \qsCountFour{}, \qsCountThree{}, \qsCountTwo{}, \qsCountOne{} universities; in terms of total score over 100 these correspond to cutoff points \qsLimitOne{}, \qsLimitTwo{} and \qsLimitThree{}. They are not based on any judgement. They derive entirely from the scores allocated, similar to other work based on published assessments \citep{doumpos2005replicating}, and there is considerable dispute over the appropriateness of such league tables as measures \citep{billaut2010shanghai}. A similar approach with universities is taken by \citet{liu2020framework}; assessment is viewed as a MCDM problem by \citet{corrente2019universities}. Each lens $j$ takes values in $[0,100\,\omega_j]$, where the published weights $\omega_j$ are 0.5, 0.2, 0.15, 0.1, 0.05. It is these that act as true weights. Since all marginal value functions are linear, the true rule belongs to the class being fitted, with no error in assignments. All that can be learnt from this data set is the rate of convergence to the correct rules.

All experiments based on actually published data are with a large pool of alternatives, fully evaluated, with ranges of scale chosen independently of the category information, that is, ranges observed, or maximum published in the case of QS, and only categories unknown, for a subset. The 4 further published data sets have all increasing variables, and feature prominently elsewhere in connection with monotone classification \citep{cano2019monotonic}; they are ERA \nEra{} alternatives, \pEra{} criteria, \qEra{} categories; ESL \nEsl, \pEsl, \qEsl; LEV \nLev, \pLev, \qLev; SWD \nSwd, \pSwd, \qSwd. As judgements by people, they may be expected to be poorly fitted by any method (Section~\ref{sec:openml}) and are similarly not referred to any particular DM. It is desirable to have a fairly large number of alternatives assigned to categories to provide a source of reference sets and testing of others. Typically this will be much larger than the number of judgements available to elicit a rule for a particular problem. So all data relate either to rules derived numerically, published sets of judgements, or simulated DMs. A number of populations is simulated, all 4-dimensional uniform, divided into 4 categories with cutoff points at values corresponding to 35, 65 and 85 per cent points. They are generated according to 6 possibilities. Piecewise linear with 3 intervals (\emph{Additive}); marginals with a power function with index log-uniformly distributed over $e^{-1.2}$ to $e^{1.2}$ (\emph{Smooth}); 0.7 additive plus 0.3 minimum of 2 lenses (\emph{Interaction}); 1 lens with a single peak (\emph{Non-monotonicity}); additive plus normally distributed independent error, standard deviation 0.05, determining category (\emph{Error}); finally, additive with 10 per cent of all alternatives assigned incorrectly to another category 2 or more categories distant (\emph{Mislabel}). All parameters of these generating models vary randomly from population to population. Weights are Dirichlet(2,2,2,2) and increments in each interval of each lens Dirichlet(1,1,1); indices as above. The peaked marginal value function is $u_1(a) = 1-|2g_1(a)-1|$ and the minimum is of the first 2 lenses. Cutoff points are quantiles of the distribution of values without error, calculated for 20\,000 further alternatives. There is no external justification either for quantiles or the parameters of the Dirichlet distributions. Quantiles were chosen to produce categories increasing in size downwards, as for QS. Any incorrect label is equally likely to be any category at least 2 categories distant. Mislabelling according to a fixed distribution conditional on true category is of course only 1 possible form \citep{frenay2014label}. It is clear from Proposition~\ref{prop:huber}(iii) that if the underlying function is correctly learned, this type of error would cause non-zero all-cutoff shortfalls. Only errors equally likely to all categories are allowed if minimising the square function is to give the same accuracy as with correct labels \citep{manwani2013noise}; various alternatives suitable if the probability of error depends on true category have been proposed, for example \citet{garg2020robust}. Most errors would be expected to be to neighbouring categories, see \emph{Error}. Members of the test set may also be incorrectly assigned, but accuracy is calculated relative to the original category. Apart from the first, these represent various ways of deviating from an additive function, following \citet{stewart1996robustness}. Most published simulations of this type of problem involve varying various features of the problem, with assignments by the true model \citep{wojcik2023representative}. Here only 1 set of numbers of criteria and categories has been considered, varying the type and degree of departure from an additive value, except for the 2 further shapes of Section~\ref{sec:further}. The number of possible solutions depends to some degree on $m$ relative to the total numbers of parameters for increments and cutoff points. Apart from Section~\ref{sec:further}, all results with consistent labels relate to 4 or 5 criteria, 4 categories; presumably the sample sizes required before effects become apparent would be larger with more parameters. Various amounts of departure were judged large enough to have a clear effect on accuracy, but small enough to leave an additive approximation reasonably good.

3 published data sets, all available from UCI, are considered in Section~\ref{sec:further}. Car Evaluation (CEV) \citep{bohanec1988car} divides \nCev{} observations into 4 ordered categories according to an a priori rule depending monotonically on each of \pCev{} ordered attributes, price, maintenance, doors, people, boot and safety. No value function of the fitted class appears consistent with this rule (Section~\ref{sec:further}). Frequencies are \vuciCevCounts. Computer Hardware (CPU) \citep{feldmesser1987cpu} comprises \nCpu{} observations and \pCpu{} attributes, cycle time, minimum and maximum memory, cache memory and minimum and maximum number of I/O channels; Fuel economy (MPG) \citep{quinlan1993mpg}, \nMpg{} observations, \pMpg{} attributes (number of cylinders, displacement, power, weight, time to accelerate, and date of model). We excluded \mpgMissing{} observations with undefined power. In both cases we divided into 4 categories according to quartiles of overall performance, as published, and of mpg. Observations corresponding exactly to a dividing value were placed in the higher category. There are thus \vuciCpuCounts{} observations in each category of CPU, and \vuciMpgCounts{} of MPG. Attributes of CEV are numbered $1,2,\dots$ from least desirable, hence price and maintenance decrease with level; the remaining attributes increase: negative values were used for cycle time (CPU), and cylinders, displacement, power and weight (MPG). All would therefore be desirable to increase, over the minimum to maximum observed. As elsewhere, we took equally-spaced subdivisions, allowing breakpoints not corresponding to attribute values. The additive and mislabelling scenarios were generated as before with 3 attributes, 3 categories dividing at the 45th and 80th percentiles; and 6 attributes, 5 categories dividing at 30, 55, 75, 90. Categories again decrease in frequency. With 3 categories, only observations in the highest and lowest categories were eligible to be moved 2 categories, 10\% of all observations being so moved.

\paragraph{Methods.}
With the exception of the final method, all are based on an additive value, and have the options described in Section~\ref{sec:model}: increments sum to 1, $\gamma = 1,2,3$ equal sub-intervals dividing each scale; widths of categories with minimum width $\varepsilon=0.01$; margins of $d/(2q)$ at each boundary for $d = 0.1, 0.25, 0.5, 0.75$; re-estimation of limits.
\begin{itemize}
\item LS-UTADIS.
\item UTADIS, minimizing total shortfall, that is, a linear programme, solved using the dual simplex algorithm implemented in HiGHS \citep{huangfu2018highs}, giving one corner point solution.
\item UTADIS plus additional penalties given by equation~\eqref{eq:J} (UTADIS with ridge). Solved as a QP using Clarabel \envClarabel{}, via CVXPY \envCvxpy{}. All tolerances relating to the gap and feasibility are set to the smallest of $10^{-12}$, $10^{-10}$, $10^{-8}$, below which an optimum is reported; solutions with a larger tolerance are discarded.
\item UTADIS-MM. After minimization as for UTADIS, with $d=0$, maximize margin at all limits subject to the sum of shortfalls no greater than the optimum. Thus choose a model maximizing the minimum distance to a limit, following \citet{greco2011representative}. Of several options discussed in \citet{wojcik2023representative} to choose one solution representing the set of all consistent ones, this was preferred as straightforward to implement with just 1 LP.
\item 2 methods selecting some central solution: Chebyshev center, that is center of maximum radius sphere within the region, determined by LP and attributed by \citet{wojcik2023representative} to \citet{doumpos2014robust}; and analytic center \citep{bous2010acuta} as modified by \citet{greco2011representative} for UTADIS. Both with no margins or re-estimation. That is, find increments summing to 1, widths $\cw\ge\varepsilon\one$, and $\vv$ non-negative, satisfying all inequalities corresponding to assigned categories of the reference set, with no margin (so for end categories just 1 relevant inequality). Normals of constraints are projected onto the subspace defined by the 2 equalities and normalized, to define a maximum Euclidean sphere within that subspace; if the radius is less than $10^{-9}$, it is assumed there are no points strictly internal to the region. The analytic center maximizes the sum of logs of amounts by which each inequality is not binding. Again solved with Clarabel, default tolerances. None of these solutions has been independently checked, nor are solutions to any of the variants of UTADIS, apart from LS-UTADIS, UTADIS with ridge, and the likelihood (Section~\ref{sec:time}). Of course, it may be impossible to represent even the true generating function with the number of intervals used; however, if there is any solution for the reference set, both centers could be calculated. They have only been applied if there is certainly a solution containing the true rule, that is, to QS and Additive, with the true number of sub-intervals. No attempt has been made to implement any of the other sampling methods discussed in \citet{wojcik2023representative}.
\item Variant with no absolute margin, and relative margin $\tau = 0$, $0.25$, $0.5$, $0.75$, $0.9$, $1$.
\item Centred least squares solution subject to restrictions of positive increments and minimum widths, and without these. With $\gamma=1$, the latter reduces to a simple expression. Dividing points would similarly change.
\item With Huber loss on shortfall, and either of the 2 all-cutoff versions described in Section~\ref{sec:huber}. $\zeta$ may take values $\tfrac14$, $\tfrac12$, 1 of $1/(2q)$, all combinations being tried, leading to 72 variations instead of 24. Ties in the results over the divisions were broken in favour of the lowest value of $\zeta$, as ties generally were broken in favour of the earliest-mentioned possibility. All variants with Huber loss on shortfall, and square on all cutoffs, were computed as conic programmes using Clarabel, with the same tight tolerances as for UTADIS with ridge. For each division, the problem was formulated once only, varying the margin. As with UTADIS, the sum of absolute values over all cutoffs is an LP. Any negative solutions, all very small, were replaced by zero, and rescaled to total 1. We have made no independent verification of these results, except that we recovered solutions equivalent to LS-UTADIS when $\zeta$ was greater than all shortfalls, and, for 2 categories, results equivalent to using only immediate cutoffs.
\end{itemize}

As usual for UTA models \citep[for example,][]{jacquet1982uta,siskos2016uta} equal widths of intervals have been assumed. Various choices of quantiles or higher-order polynomials, splines, and so on \citep{sobrie2018utapoly} could of course be made. \citet{stewart1993piecewise} concludes, from extensive simulation, that some departure from linearity would generally be beneficial, and piecewise linear adequate. More intervals were not tried, since 1 corresponds to the actual function for QS, 3 to that assumed for one of the simulations, and greater numbers exceed the information provided by some reference sets.

Thus margins are imposed at all limits including uppermost and lowermost, unlike the version given by \citet{zopounidis1999utadis}; accordingly LS-UTADIS is simply UTADIS with squared deviations. Various aspects of this set-up are the result of failures with others. It was originally intended to use a zone proportional to each category width; however, in preliminary work not part of the reproducible set of results, a fixed margin performed as well or better over most circumstances. Hence LS-UTADIS with this feature was adopted, retaining the other only to introduce centring. In addition, in those experiments, only the upper limit had a margin, fewer margins were tried and limits were not re-estimated. With these changes, squaring appeared more beneficial for QS and beneficial, though less markedly, for the published data sets. Once both solutions had identical options, no advantage remained, motivating the common features of all methods. Re-estimating limits was included following experiments showing this to be detrimental if done jointly with the increments for LEV and SWD. It is desirable to know the separate effects of the ridge and squaring, hence the addition of the former to the original UTADIS solution, although as noted in Section~\ref{sec:fixed} there is some ambiguity for finite values of the parameters.

Also compared is another penalty function on the same set of functions. The standard cumulative likelihood \citep{mccullagh1980} is maximized over $\ph(a)^{\top}\vv$, positive increments, arbitrary scale of $\vv$ and cutoff points increasing, adding a small constant times sum of squares of increments. Subject to division through by sum of increments to give scale 1, this gives an increasing value function. If all rescaled cutoff points fall between 0 and 1, they can be taken as limits on this scale, although this may fail to happen. Cumulative odds models with sums of effects were considered by \citet{hastie1987proportional}. Imposing monotonicity on effects, and maximizing directly, is the approach of \citet{espinosa2019constrained}, and a similar model allowing non-ordinal responses is developed by \citet{pya2015scam}. Our solution differs only in details. Increments are all squares and intervals between successive cutoff points exponentials, allowing use of standard unconstrained maximization. Due to the squares this is no longer a convex problem, so 3 starting values are tried, retaining the maximum. A small amount of regularization, $10^{-3}$, fixed, avoids difficulties should all points be correctly classified with a separating hyper-surface. It will therefore determine the solution returned in such circumstances. Penalties of this form tend to give maximum-margin solutions as their weight tends to 0 \citep{rosset2003margin}, although we have not investigated whether this applies to our problem. Variations of this regularization constant are considered in Section~\ref{sec:fixed}. Methods of inference have been developed assuming all variables ordinal \citep{espinosa2021inference} but not for our case, so none is attempted. There are therefore no margins or restrictions on minimum widths, regularization applies to unscaled increments, and only $\gamma$ chosen under the general rules for cross-validation. After scaling, there is no reason to expect cutoff points away from 0 and 1 or widths exceeding the minimum, nor indeed confined to 0 to 1, that is, some extreme categories may contain no points, analogous to the effect sometimes seen with re-estimated limits (Section~\ref{sec:recal}). We re-ran the \ologLimFits{} optimized solutions of this study using their reference sets, obtaining identical results to those recorded, to within \ologLimAccDiffMax{}. However, \ologLimOut{} had at least 1 limit outside $[0,1]$; \ologLimEraShare{} per cent of all solutions for ERA, \ologLimSwdShare{} per cent SWD, and no more than \ologLimOtherShareMax{} per cent any other set. None of those considered in more detail in Section~\ref{sec:fixed}, \fixOlogAFits{} solutions in total, did so. So with erroneous labels there is a substantial chance that the solution is not an admissible set of limits and categories. Hence this is less directly comparable with other methods than UTADIS with ridge. Only that and LS-UTADIS involve exactly the same terms except for squares on the shortfalls. The usual UTADIS differs in both respects. For brevity it is referred to as proportional-odds. Of various forms of regression applicable to ordered responses \citep{tutz2022ordinal}, only the cumulative version has analogous form to finding a single number to compare with a set of ordered constants; logistic is standard. Another link function, probit, has been used \citep{li2025temporal} but we have not experimented with it and say nothing further about its performance. It is simply the natural choice of likelihood given an ordered set of categories and the relevant single index.

3 methods unrelated to a value function are included for comparison. Proportional odds assuming a linear combination of standardized variables as index, regularized slightly; standard radial-basis function support vector machine, optimizing parameters \citep{cortes1995svn}; random forest, 300 trees \citep{breiman2001rf}. They indicate generally the level achievable given the reference sets, but with no attempt at comparable optimization. Given the origin of these 4 data sets, various forms of monotone classification and ordinal regression by SVM \citep{cano2019monotonic,chu2007svor,kotlowski2013isotonic} would seem appropriate comparisons. However, none yields an estimate of a value function with cutoff points and much has been written comparing such procedures.

We also include 1 other method involving regularization (regularised UTADIS) to choose between possible value functions. Following \citet{doumpos2007regularized}, minimize sum of increments plus a further parameter times total distance from all points being at distance 1 from free cutoff points. So there is no restriction summing to unity or on values at ends of ranges. It is again solved as a LP. In principle there could be a separate parameter for each category, but no suggestion of how they should be chosen; a single value common to all was chosen from 0.1 to 100, as well as $\gamma$ and re-estimation, giving a comparable number of possibilities. This is of course quite different from adding squares to the usual normalized solution of UTADIS.

Various other approaches to aiding decisions have not been considered. In particular, various forms of learning pairwise comparisons \citep{liu2019regularization,liu2020framework} would not estimate limits to categories and hence fall outside our problem; similarly learning within categories to decide others \citep{liu2021interacting}. The approaches of \citet{ru2023probabilistic} involve an underlying prior and some sampling procedure leading to a range of probable functions. All others give a single solution. So clearly the range considered is limited.

\paragraph{Details.}
The linear proportional-odds model is regularized with $10^{-3}$ times the sum of squares of coefficients. Range of constants for the SVM is 1, 10, 100 and widths 0.1, 1 and the program default. Maximization for proportional-odds on additive value is with L-BFGS-B as implemented in SciPy, maximum of 2000 iterations, other defaults, starting with sums of increments 4, 12 and 36. UTADIS-MM allows excess total shortfall up to $10^{-6}$, maximum value of common margin 0.5, and reverts to solution to first problem if necessary. If all increments returned by regularized UTADIS are 0, clearly all values 0 and assigned accordingly. The relevant published sets may be found on OpenML with IDs 1030 (ERA), 1027 (ESL), 1029 (LEV), 1028 (SWD). All variables assumed to be desirable to maximize, with observed minima and maxima. Random seed 20261112. Elapsed time to complete each optimization, including any cross-validation and final predictions, was noted for all 3 main sets of experiments, with several proceeding concurrently. In total this amounts to about \computeHours{} hours, and has no direct relation to either total CPU time or actual time taken. Individual optimizations for the timing comparison and those for investigating variation of likelihood solution (Section~\ref{sec:time}) were similarly measured. Others were not. Running all experiments to produce all final displays from start to finish took \rerunHours{} hours elapsed time with \rerunJobs{} concurrent processes, on the machine detailed below. The experiments reported in Sections~\ref{sec:v08} and~\ref{sec:further} were subsequently carried out on the applicable reference sets, requiring a further \vRunHours{} hours elapsed time (\vHours{} hours over all the fits, by the same method of calculation) and are additional to those so far described as being rerun. No measure was made of total processor time. All entries common to any output file and that originally used for displays in this paper are identical (apart from times). All columns relating to just 1 run have been left blank. All may be seen on request (see availability of data).

For each data set, various $m$, $m$ alternatives selected randomly as reference set, providing at least 2 different categories, remainder forming test set, repeated 100 times. Initial $m$ chosen with reference to the ranges used elsewhere (\citealp[\citWojMin{} to \citWojMax{} per category]{wojcik2023representative}; \citealp[variously \citRuAltMin{} to \citRuAltMax{}]{ru2023probabilistic}), and continued doubling until changes with increasing $m$ were slight. So much larger sizes, such as virtually all of \citLiuUniv{}+ \citep{liu2019regularization} or \citDzRef{} \citep{doumpos2007regularized}, would be outside this range. Simulations all based on initial sample 2160. Value-based solutions retain a width for any category not represented, precluding prediction by others.

All features to be optimized are decided on cross-validation on subsets of reference set \citep{varma2006bias}. Hence 3 values of $\gamma$, 4 of $d$ and re-estimation yes or no, for LS-UTADIS, UTADIS and with ridge; similarly for regularized version, varying the constant instead of $d$; 6 for UTADIS-MM, 36 for the relative version, \cenCand{} of $\gamma$ for proportional-odds and each center, 9 combinations of constants for the SVM. All on corresponding divisions, choosing maximum accuracy or, in case of equality for any of the value-based methods, least mean error of category. No optimization of parameters of proportional-odds on linear combination of variables, or random forest. Equal numbers from each category as far as possible, that is, 5 divisions if minimum frequency at least 5, otherwise that number. If only 1 point of some category, divisions into 5 ignoring categories. Thus for $m=\qsMsmall$ for QS there are on average only \qsFoldsSmall{} divisions, and only \qsStratSmall{} per cent with any stratification; however, all 4 categories occur for \qsAllCatSmall{} per cent of samples. Section~\ref{sec:time} shows how many divisions were in fact used. Clearly such small numbers to choose among a number of possibilities will lead to considerable variability, more so with more possibilities to choose between. There will also be some degree of over-fitting in choosing the best function, its effect being often comparable to the differences between algorithms, according to \citet{cawley2010overfitting}. We did not try imposing equal numbers of functions to compare, nor repeating internal cross-validation. Accordingly all experiments with no reason to expect inconsistent labels are repeated with no choice to make (Section~\ref{sec:fixed}). In addition to correct classification, mean error of assigned category and, where possible, $L_1$ distance of weights from true values are calculated. The former is primary since used for comparison in most other work \citep[for example,][]{kadzinski2025expressiveness,ru2023probabilistic}, and the second is commonly used as an alternative for ordered categories \citep{gutierrez2016survey}. Both treat all errors as equally serious regardless of category. Comparisons are all between corresponding samples. If the $t$-statistic for difference of paired observations exceeds 1.96 in either direction this is counted as significant, and the numbers of instances so classified are given. As well as unadjusted values, for the experiments of Sections~\ref{sec:v08} and~\ref{sec:further} we show those resulting from the procedure of \citet{benjamini1995fdr} to limit the proportion of false positives. These are based on two-sided standard normal probabilities for identical statistics, with adjustment across all cells arising from 1 set of comparisons, that is, 1 table, at 5 per cent. No adjustment has been made to the numbers reported in the other parts of this section. If summing over a large number of cells, an appropriate measure is the expected proportion incorrect, of those classified, which this method is designed to control, not the chance of at least 1 incorrect classification. We have not thought it meaningful to pool over all combinations of $m$ and type of data set, for example by ranking methods \citep{demsar2006statistical}, since there are several of each, chosen deliberately. So these give some indication of circumstances leading to differences, relying on means and standard errors of differences for details. Obviously all inferences relate only to variability due to choice of reference set from each particular set. It has been pointed out that such tests will tend to over-estimate significance of superiority of one method of fitting, if applied to various random subdivisions of 1 data set \citep{dietterich1998tests,nadeau2003inference}. We are not concerned with that question, but simply the expected performance given $m$ randomly chosen from the full set considered. So samples are appropriately independent and standard methods apply. Inferences about other data from same general source can only be made with reference to the simulations, where all features vary randomly. All calculations based on various packages for Python \envPython{} (in particular NumPy \envNumpy, SciPy \envScipy, sklearn \envSklearn, CVXPY \envCvxpy), run on a single machine, \envCpu, \envLogical{} processors, \envMemory{}~GiB. Details are given with all versions in a standard file produced at run time and accompanying code, except that processor, number and memory were subsequently appended, and Clarabel version much later from current install. All may be regenerated with appropriate seeds using the script \texttt{scripts/reproduce.py}.

\subsection{Known additive rule: QS}
\label{sec:qs}

\begin{figure}[t]
\centering
\includegraphics{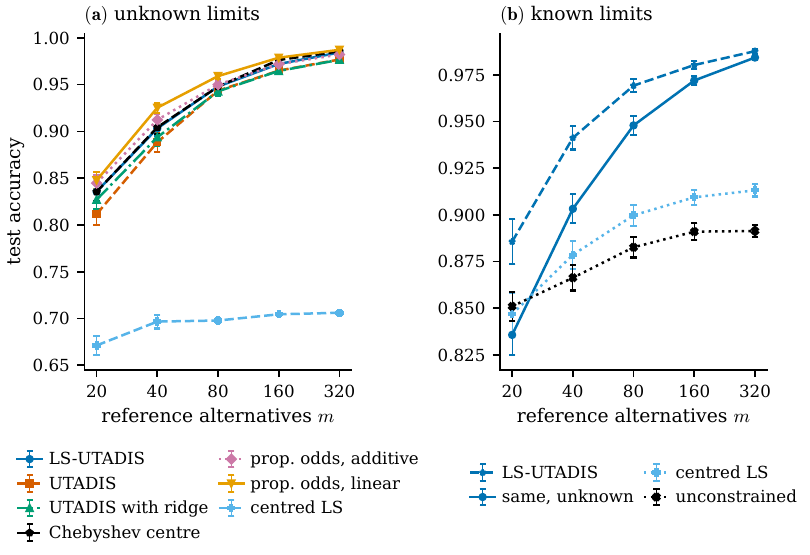}
\caption{QS rankings: test accuracy against the number of reference alternatives, with 95\% confidence intervals over 100 reference sets. (a) Unknown limits. Least-squares UTADIS is above the UTADIS programme, most clearly for the smallest reference sets. The Chebyshev centre of the compatible models, a central choice without squaring, is not significantly less accurate than least-squares UTADIS. Proportional odds on the additive value is close to or above least-squares UTADIS and the UTADIS programme. Centred least squares stays far below. (b) Known limits help least-squares UTADIS; they do not rescue centring (``unconstrained'' is centred least squares without sign constraints).}
\label{fig:qs}
\end{figure}

Results are shown in Table~\ref{tab:qs} and Figure~\ref{fig:qs}. Allowing arbitrary limits, for $m=\qsMsmall$ accuracy is \qsLsuSmall{} for LS-UTADIS compared to \qsUtaSmall{}, mean difference \qsGainLsuUtaSmall{}, SE \qsGainLsuUtaSmallSE{}. Corresponding figures for $m=\qsMsecond$ are \qsGainLsuUtaSecond{} (\qsGainLsuUtaSecondSE), decreasing to \qsGainLsuUtaBeyondMax{} or less. So part is attributable to adding the sums of squares, improving by \qsGainUtaRidgeUtaSmall{} (\qsGainUtaRidgeUtaSmallSE) for $m=\qsMsmall$, with a further improvement of \qsGainLsuUtaRidgeSmall{} (\qsGainLsuUtaRidgeSmallSE) due to squaring. For \qsGainLsuUtaRidgeWins{} values of $m$, LS-UTADIS is significantly better than UTADIS with the additional terms. This is with both set to $10^{-4}$; the sensitivity of this conclusion to their values is considered in Section~\ref{sec:fixed}. UTADIS-MM trails by \qsGainLsuUtaMMSmall{} (\qsGainLsuUtaMMSmallSE) for $m=\qsMsmall$ but is indistinguishable thereafter. Selecting the Chebyshev center, with optimized number of sub-intervals as before, performs virtually identically to LS-UTADIS, with differences from \qsGainLsuChebSmall{} (\qsGainLsuChebSmallSE) for $m=\qsMsmall$ to between \qsGainLsuChebMax{} and \qsGainLsuChebMin{}; it is never significantly inferior, but superior for \qsGainLsuChebLosses{} $m$. The other center is \qsGainLsuAcenSmall{} (\qsGainLsuAcenSmallSE) and \qsGainLsuAcenSecond{} (\qsGainLsuAcenSecondSE) worse for $m=\qsMsmall$ and \qsMsecond. Hence some choice of a representative point from those consistent would be as successful as LS-UTADIS. Regularized UTADIS is on average \qsGainLsuUtaRegSmall{} (\qsGainLsuUtaRegSmallSE) lower for $m=\qsMsmall$, tending to \qsGainLsuUtaRegLarge{} (\qsGainLsuUtaRegLargeSE) by $m=\qsMlarge$. Differences with usual UTADIS range from \qsGainUtaRegUtaMin{} to \qsGainUtaRegUtaMax{} and are significant for \qsGainUtaRegUtaWins{} values of $m$ only, always in its favor.

Proportional odds on additive value is \qsGainOlogAUtaSmall{} points (\qsGainOlogAUtaSmallSE) better than UTADIS at $m=\qsMsmall$, and is no worse than LS-UTADIS up to $m=\qsMsecond$ inclusive. Hence LS-UTADIS falls short of proportional odds by \magn{\qsGainLsuOlogASmall} points (\qsGainLsuOlogASmallSE) and \magn{\qsGainLsuOlogASecond} points (\qsGainLsuOlogASecondSE) at $m=\qsMsmall$ and \qsMsecond, but leads it by \qsGainLsuOlogALarge{} points (\qsGainLsuOlogALargeSE) at $m=\qsMlarge$. Proportional odds is always best, using the linear index, and LS-UTADIS ranges from \magn{\qsGainLsuOlogMax} to \magn{\qsGainLsuOlogMin} points worse. For QS there would be only one parameter, a slope, under this model, for each criterion, in contrast to as many increments as criteria times sub-intervals. Since all models give comparable results for the weights on criteria (Table~\ref{tab:qs}), there seems nothing special about LS-UTADIS in estimating them.

\begin{table}[t]
\centering
\caption{QS rankings: test accuracy (mean over 100 reference sets, with the half-width of the 95\% confidence interval) and, for the methods that return a value function, the $L_1$ error of the trade-off weights. Best mean accuracy per column in bold, separately for unknown and known limits. Abbreviations: LS-UTADIS, least-squares UTADIS; UTADIS-MM, UTADIS with a max-margin second stage; centred LS, centred least squares; SVM (RBF), support vector machine with a Gaussian kernel.}
\label{tab:qs}
\footnotesize
\setlength{\tabcolsep}{3pt}
\resizebox{\textwidth}{!}{%
\begin{tabular}{@{}lccccccc@{}}
\toprule
 & \multicolumn{5}{c}{Accuracy, by number of reference alternatives $m$} & \multicolumn{2}{c}{Weight error} \\
\cmidrule(lr){2-6}\cmidrule(l){7-8}
Method & 20 & 40 & 80 & 160 & 320 & $m=20$ & $m=320$ \\
\midrule
\multicolumn{8}{@{}l}{\emph{Unknown limits}} \\
LS-UTADIS & 0.836 {\scriptsize$\pm$.011} & 0.903 {\scriptsize$\pm$.008} & 0.948 {\scriptsize$\pm$.005} & 0.972 {\scriptsize$\pm$.002} & 0.984 {\scriptsize$\pm$.001} & 0.369 & 0.030 \\
LS-UTADIS, no recalibration & 0.840 {\scriptsize$\pm$.010} & 0.898 {\scriptsize$\pm$.008} & 0.947 {\scriptsize$\pm$.005} & 0.972 {\scriptsize$\pm$.002} & 0.984 {\scriptsize$\pm$.001} & 0.367 & 0.031 \\
Relative margin & 0.825 {\scriptsize$\pm$.011} & 0.903 {\scriptsize$\pm$.008} & 0.945 {\scriptsize$\pm$.004} & 0.970 {\scriptsize$\pm$.002} & 0.982 {\scriptsize$\pm$.002} & 0.388 & 0.033 \\
Centred LS & 0.671 {\scriptsize$\pm$.010} & 0.697 {\scriptsize$\pm$.007} & 0.698 {\scriptsize$\pm$.003} & 0.705 {\scriptsize$\pm$.002} & 0.706 {\scriptsize$\pm$.002} & 0.425 & 0.100 \\
Centred LS, unconstrained & 0.740 {\scriptsize$\pm$.022} & 0.817 {\scriptsize$\pm$.014} & 0.857 {\scriptsize$\pm$.012} & 0.887 {\scriptsize$\pm$.009} & 0.895 {\scriptsize$\pm$.005} & 0.356 & 0.066 \\
Huber shortfalls & 0.832 {\scriptsize$\pm$.010} & 0.898 {\scriptsize$\pm$.009} & 0.947 {\scriptsize$\pm$.005} & 0.972 {\scriptsize$\pm$.002} & 0.984 {\scriptsize$\pm$.001} & 0.372 & 0.031 \\
LS-UTADIS, all thresholds & 0.836 {\scriptsize$\pm$.011} & 0.903 {\scriptsize$\pm$.008} & 0.948 {\scriptsize$\pm$.005} & 0.972 {\scriptsize$\pm$.002} & 0.984 {\scriptsize$\pm$.001} & 0.369 & 0.030 \\
UTADIS & 0.812 {\scriptsize$\pm$.012} & 0.888 {\scriptsize$\pm$.010} & 0.944 {\scriptsize$\pm$.005} & 0.965 {\scriptsize$\pm$.003} & 0.977 {\scriptsize$\pm$.002} & 0.406 & 0.050 \\
UTADIS with ridge & 0.827 {\scriptsize$\pm$.010} & 0.894 {\scriptsize$\pm$.009} & 0.943 {\scriptsize$\pm$.005} & 0.965 {\scriptsize$\pm$.003} & 0.977 {\scriptsize$\pm$.002} & 0.393 & 0.051 \\
UTADIS-MM & 0.814 {\scriptsize$\pm$.012} & 0.897 {\scriptsize$\pm$.008} & 0.947 {\scriptsize$\pm$.005} & 0.973 {\scriptsize$\pm$.003} & 0.985 {\scriptsize$\pm$.002} & 0.367 & 0.029 \\
Regularised UTADIS & 0.804 {\scriptsize$\pm$.013} & 0.889 {\scriptsize$\pm$.010} & 0.942 {\scriptsize$\pm$.005} & 0.970 {\scriptsize$\pm$.003} & 0.983 {\scriptsize$\pm$.002} & 0.392 & 0.034 \\
UTADIS, all thresholds & 0.817 {\scriptsize$\pm$.010} & 0.888 {\scriptsize$\pm$.010} & 0.941 {\scriptsize$\pm$.006} & 0.965 {\scriptsize$\pm$.003} & 0.977 {\scriptsize$\pm$.002} & 0.390 & 0.050 \\
Chebyshev centre & 0.836 {\scriptsize$\pm$.010} & 0.904 {\scriptsize$\pm$.008} & 0.948 {\scriptsize$\pm$.005} & 0.976 {\scriptsize$\pm$.002} & 0.986 {\scriptsize$\pm$.002} & 0.337 & 0.030 \\
Analytic centre & 0.833 {\scriptsize$\pm$.010} & 0.894 {\scriptsize$\pm$.007} & 0.945 {\scriptsize$\pm$.005} & 0.973 {\scriptsize$\pm$.002} & 0.985 {\scriptsize$\pm$.002} & 0.316 & 0.028 \\
Proportional odds, additive value & 0.845 {\scriptsize$\pm$.009} & 0.912 {\scriptsize$\pm$.006} & 0.950 {\scriptsize$\pm$.004} & 0.972 {\scriptsize$\pm$.002} & 0.982 {\scriptsize$\pm$.002} & 0.336 & 0.040 \\
Proportional odds, linear index & \textbf{0.848} {\scriptsize$\pm$.009} & \textbf{0.925} {\scriptsize$\pm$.005} & \textbf{0.959} {\scriptsize$\pm$.003} & \textbf{0.979} {\scriptsize$\pm$.002} & \textbf{0.987} {\scriptsize$\pm$.001} & -- & -- \\
SVM (RBF) & 0.707 {\scriptsize$\pm$.014} & 0.804 {\scriptsize$\pm$.011} & 0.863 {\scriptsize$\pm$.006} & 0.910 {\scriptsize$\pm$.004} & 0.942 {\scriptsize$\pm$.003} & -- & -- \\
Random forest & 0.672 {\scriptsize$\pm$.010} & 0.757 {\scriptsize$\pm$.006} & 0.812 {\scriptsize$\pm$.005} & 0.851 {\scriptsize$\pm$.004} & 0.879 {\scriptsize$\pm$.003} & -- & -- \\
\addlinespace
\multicolumn{8}{@{}l}{\emph{Known limits}} \\
LS-UTADIS & \textbf{0.886} {\scriptsize$\pm$.012} & \textbf{0.941} {\scriptsize$\pm$.006} & \textbf{0.969} {\scriptsize$\pm$.003} & \textbf{0.980} {\scriptsize$\pm$.002} & \textbf{0.988} {\scriptsize$\pm$.001} & 0.276 & 0.025 \\
Relative margin & 0.877 {\scriptsize$\pm$.011} & 0.936 {\scriptsize$\pm$.007} & 0.962 {\scriptsize$\pm$.005} & 0.979 {\scriptsize$\pm$.002} & 0.986 {\scriptsize$\pm$.001} & 0.298 & 0.027 \\
Centred LS & 0.847 {\scriptsize$\pm$.011} & 0.879 {\scriptsize$\pm$.008} & 0.900 {\scriptsize$\pm$.006} & 0.909 {\scriptsize$\pm$.004} & 0.913 {\scriptsize$\pm$.003} & 0.354 & 0.191 \\
Centred LS, unconstrained & 0.851 {\scriptsize$\pm$.008} & 0.866 {\scriptsize$\pm$.007} & 0.883 {\scriptsize$\pm$.005} & 0.891 {\scriptsize$\pm$.004} & 0.891 {\scriptsize$\pm$.003} & 0.397 & 0.250 \\
\bottomrule
\end{tabular}}
\end{table}

\begin{figure}[t]
\centering
\includegraphics{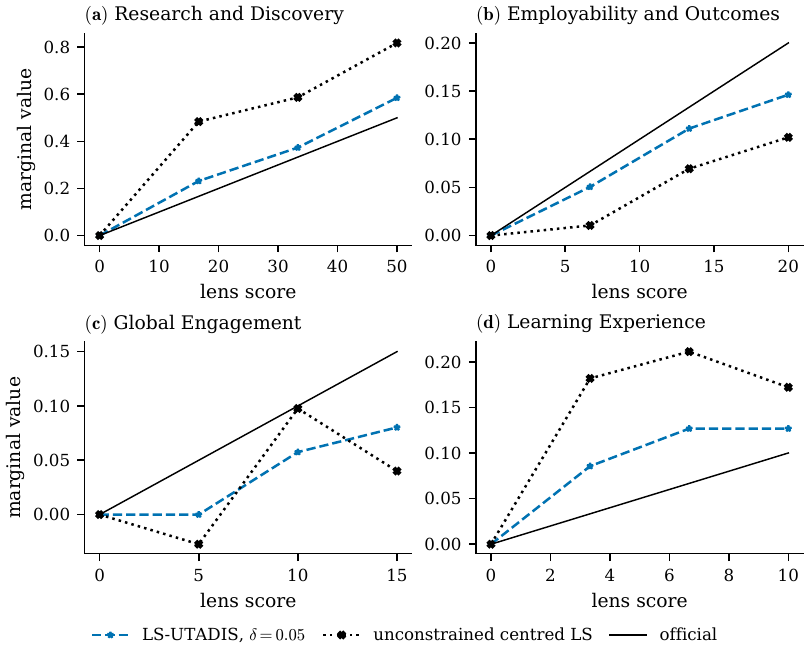}
\caption{Marginal value functions inferred from one reference set of 40 universities (three sub-intervals per criterion, known limits, margin $\delta=0.05$) for the four lenses with the largest weights. The least-squares UTADIS estimate is monotone; centred least squares without sign constraints decreases on some sub-intervals. (a)--(d) One lens each. The figure is an illustration from a single draw; frequencies are in Table~\ref{tab:violations}.}
\label{fig:marginals}
\end{figure}

Various features of the designs appear to be reflected in different patterns of results for centred least squares. Ignoring limits and imposing constraints, results are relatively insensitive to $m$, rising from \qsClsSmall{} ($m=\qsMsmall$) to \qsClsLarge{} ($m=\qsMlarge$), against \qsLsuSmall--\qsLsuLarge{} for LS-UTADIS. With 1 sub-interval on each criterion, results range from \qsClsOneMin{} to \qsClsOneMax{}, and one end is at $\varepsilon$, the lowest permissible value, for \qsClsFloorMin{} to \qsClsFloorMax{} per cent of widths (c.f.\ Example~\ref{ex:collapse}). If the limits are allowed a priori to take any values, results range from \qsClsFreeMin{} to \qsClsFreeMax{} with the limits in ascending order. While better than above, these results should not be interpreted as the ability to correctly sort alternatives: for \violUnorderedMin{} to \violUnorderedMax{} per cent of reference sets, for both QS and the additive scenario, limits are not all ascending, and for \violNegKnownMin{} to \violNegKnownMax{} per cent when true limits are given, 1 or more increments are negative (Table~\ref{tab:violations}). Thus violations of Corollaries~\ref{cor:free} and~\ref{cor:ordered} would be expected to be quite frequent. Imposing constraints thus ensures admissible values of all parameters, but leads to poorer sorting, so it is desirable to impose constraints and not centre. Even given the true limits, results rise only to \qsClsKLarge{}, at $m=\qsMlarge$, against \qsLsuKLarge{} for LS-UTADIS. Figure~\ref{fig:marginals} shows 1 case of Corollary~\ref{cor:free} arising.

\begin{table}[t]
\centering
\caption{Centred least squares without sign and width constraints (one sub-interval per criterion): share of the 100 reference sets in which at least one increment is negative, mean share of negative increments, and share of sets in which the estimated limits are not increasing (a dash where the limits are known).}
\label{tab:violations}
\small
\resizebox{\textwidth}{!}{%
\begin{tabular}{@{}llrccc@{}}
\toprule
Data & Limits & $m$ & Some increment negative & Negative increments & Limits not increasing \\
\midrule
QS rankings & unknown & 20 & 0.29 & 0.062 & 0.92 \\
QS rankings & unknown & 40 & 0.18 & 0.036 & 0.97 \\
QS rankings & unknown & 80 & 0.10 & 0.020 & 1.00 \\
QS rankings & unknown & 160 & 0.03 & 0.006 & 1.00 \\
QS rankings & unknown & 320 & 0.01 & 0.002 & 1.00 \\
QS rankings & known & 20 & 0.81 & 0.170 & -- \\
QS rankings & known & 40 & 0.76 & 0.156 & -- \\
QS rankings & known & 80 & 0.86 & 0.172 & -- \\
QS rankings & known & 160 & 0.95 & 0.190 & -- \\
QS rankings & known & 320 & 1.00 & 0.200 & -- \\
\addlinespace
Additive scenario & unknown & 20 & 0.10 & 0.028 & 0.91 \\
Additive scenario & unknown & 40 & 0.05 & 0.013 & 0.87 \\
Additive scenario & unknown & 80 & 0.03 & 0.007 & 0.87 \\
Additive scenario & unknown & 160 & 0.01 & 0.003 & 0.92 \\
Additive scenario & known & 20 & 0.35 & 0.090 & -- \\
Additive scenario & known & 40 & 0.26 & 0.065 & -- \\
Additive scenario & known & 80 & 0.20 & 0.050 & -- \\
Additive scenario & known & 160 & 0.18 & 0.045 & -- \\
\bottomrule
\end{tabular}}
\end{table}

\subsection{Range over margin and penalty}
\label{sec:margin}

\begin{figure}[t]
\centering
\includegraphics{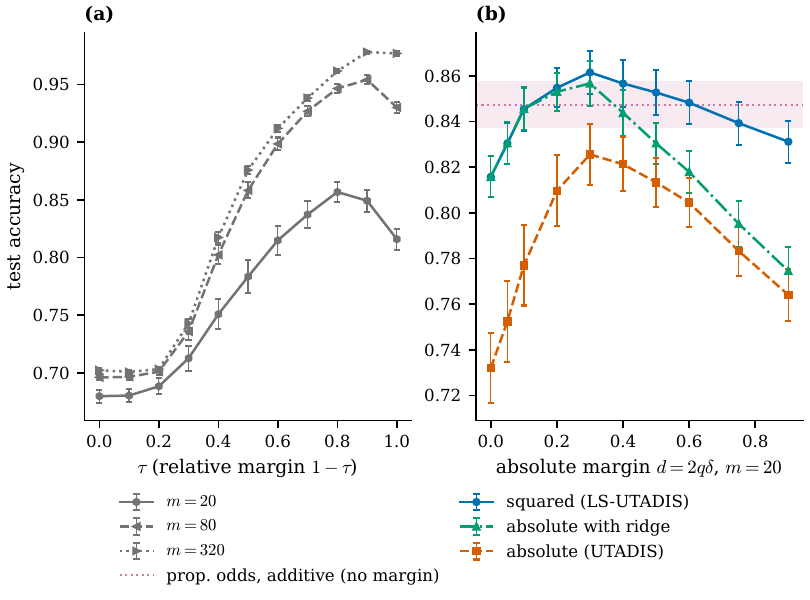}
\caption{Fixed hyperparameters on the QS data (one sub-interval per criterion, unknown limits, no recalibration), with 95\% confidence intervals over 100 reference sets. (a) Relative margin: the accuracy of centring ($\tau=0$) improves little with the size of the reference set and stays far below that of larger $\tau$, and the best $\tau$ moves towards one as the set grows. (b) Absolute margin with 20 reference alternatives: the squared penalty and the absolute penalty with ridge terms peak at nearly the same accuracy, the vertex solution of the linear programme is below both, and the squared penalty loses least when the margin is too large. The horizontal line and band are proportional odds on the additive value, which has no margin, fitted to the same reference sets with its ridge at $10^{-3}$; over a middle range of margins it lies below the squared penalty.}
\label{fig:profile}
\end{figure}

Figure~\ref{fig:profile} shows the range over values of margin and penalty, all else being fixed, and in particular with $\gamma=1$ and without recalibration, to remove this source of fluctuation. For centring, accuracy rises only slightly with $m$, \profQsSmallRelFirst{} to \profQsLargeRelFirst{} ($m=\qsMsmall$ to \qsMlarge), compared to fairly substantial values if the optimal value of $\tau$ (\profQsSmallRelBest{} and \profQsLargeRelBest{}) is used. The maximum accuracy attainable with absolute and relative margins is almost identical: \profQsSmallAbsBestAcc{} and \profQsSmallRelBestAcc{} ($m=\qsMsmall$) and \profQsLargeAbsBestAcc{} and \profQsLargeRelBestAcc{} ($m=\qsMlarge$). However, the amounts varying over the ranges examined, as the corresponding margin parameter varies, are quite different: \profQsSmallAbsRange{} points for $d$ over 0 to 0.9, \profQsSmallRelRange{} points for $\tau$ over 0 to 1 and \profQsSmallRelInnerRange{} points for 0.5 to 1, for $m=\qsMsmall$. These are of course only to be interpreted as ranges over an arbitrary chosen set of values. Broadly similar results apply with 3 sub-intervals per criterion in the additive case, maxima being \profAddSmallAbsBestAcc{} and \profAddSmallRelBestAcc{} ($m=\simMsmall$).

If the optimal values are chosen by cross-validation, LS-UTADIS exceeds the relative margin by up to \qsGainLsuRelMax{} points on QS, by \simGainLsuRelMin{} to \simGainLsuRelMax{} points on other scenarios (significantly for \simGainLsuRelWins{} out of \simCells{} sets) and by up to \openmlGainLsuRelMax{} point on the published data. Some difference might be expected merely from tuning to the best of 24 rather than 36 values, but there is no clear basis for apportioning the effects. Thus there seems no advantage, in terms of resulting accuracy, in the relative margin, whose only attraction is its natural association with centring.

Figure~\ref{fig:profile}b shows ranges of penalty, with fixed margin. Best values at $m=\qsMsmall$ are \profQsSmallAbsBestAcc{}, \profQsSmallUtaRidgeBestAcc{}, \profQsSmallUtaBestAcc{} for the square, regularised absolute, and LP. Hence most of the advantage of the square can be obtained by an absolute penalty with ridge terms if a good margin is chosen, that is with $\rho=\eta=10^{-4}$. From results of Section~\ref{sec:fixed} this conclusion does not appear very sensitive to $\rho=\eta$, unlike comparisons between the penalties, both with ridge. Only for very large margins is there any substantial difference, the square varying less over its range, \profQsSmallAbsRange{} points, than the regularised absolute, \profQsSmallUtaRidgeRange{} points. Little variation results from varying $\varepsilon$: accuracy on QS for $d=0.25$ is \floorLsuSmallSpan{} at $m=\qsMsmall$, and \floorLsuLargeSpan{} at $m=\qsMlarge$. The range for centring, \floorClsLargeSpan{} ($m=\qsMlarge$), suggests this is unlikely to account for its generally poorer results. Penalties on the two end intervals, which are not included in UTADIS, have little effect: LS-UTADIS differs from that without these terms by \simGainLsuLsuOneMin{} to \simGainLsuLsuOneMax{} points in various simulations with no mislabeling, and by \qsGainLsuLsuOneMin{} to \qsGainLsuLsuOneMax{} points on QS.

The likelihood gives one value only, since there is no margin. It has been evaluated for all reference sets arising with the same number of sub-intervals per criterion, and ridge terms $10^{-3}$, as chosen by the tuning, $10^{-4}$ and $10^{-2}$. Convergence was always achieved and all limits lay in $[0,1]$, after multiplication by a constant. On QS it achieves \plQsSmallAcc{} at $m=\qsMsmall$, its best result (Figure~\ref{fig:profile}b). Over the \plQsSmallMargins{} margins considered, it differs significantly from LS-UTADIS on \plQsSmallLsuSig{} occasions: less accurately for $d=\plQsSmallLsuAboveSet$; more accurately for $d=\plQsSmallLsuBelowSet$. For $m=\qsMsmall$, choosing a posteriori the largest of the \plQsSmallMargins{} margins, LS-UTADIS exceeds it on average by \plQsSmallBestDiff{} points (s.e.\ \plQsSmallBestDiffSE). This is not surprising, as it chooses the largest of \plQsSmallMargins{} highly correlated means, so the s.e.\ would tend to underestimate. The same average difference, \plQsSmallBothBestDiff{} points (\plQsSmallBothBestDiffSE), results if it is allowed its best a posteriori choice of ridge term. Choosing all parameters a priori, and using the reference sets described in Section~\ref{sec:fixed}, gives \fixGainLsuOlogAQSSmall{} points (\fixGainLsuOlogAQSSmallSE); with tuning, \qsGainLsuOlogASmall{} points (\qsGainLsuOlogASmallSE), that is, non-significantly different from 0 (Section~\ref{sec:qs}). Thus, for this $m$, there appears to be no benefit in the likelihood, if untuned. Tuning may be detrimental, but all this experiment can say is that no such difference was apparent, and that it would be unlikely to show up over this restricted set of profiles. There are also other differences from experiments involving tuning (Section~\ref{sec:fixed}). For $m=\plQsMidM$, it is significantly higher, at the ridge term chosen by tuning, for \plQsMidLsuBelow{} out of \plQsMidMargins{} margins, but is lower by \plQsMidBestDiff{} points (\plQsMidBestDiffSE) for the most favourable margin, or higher by \magn{\plQsMidBothBestDiff} points (\plQsMidBothBestDiffSE) if allowed its best ridge term a posteriori. At $m=\qsMlarge$ these become \plQsLargeBestDiff{} points (\plQsLargeBestDiffSE) and \plQsLargeBothBestDiff{} points (\plQsLargeBothBestDiffSE) (ridge $10^{\plQsLargeLikBestRidgeExp}$). Greater differences are found additively: best value \plAddSmallAcc{} ($m=\simMsmall$), always significantly exceeding LS-UTADIS. Against LS-UTADIS for its best margin, it is higher by \magn{\plAddSmallBestDiff} points (\plAddSmallBestDiffSE) and \magn{\plAddLargeBestDiff} points (\plAddLargeBestDiffSE) for $m=\simMsmall$ and \plAddLargeM. There would thus appear some advantage in using it, choosing parameters a priori. Changing the ridge term over 3 values changes results by up to \plQsSmallRidgeRange{} points on QS, and \plAddSmallRidgeRange{} points additively, for $m=\qsMsmall$.

\subsection{Incorrectly specified true generating rule, random errors, mislabelled examples}
\label{sec:sim}

\begin{figure}[tp]
\centering
\includegraphics{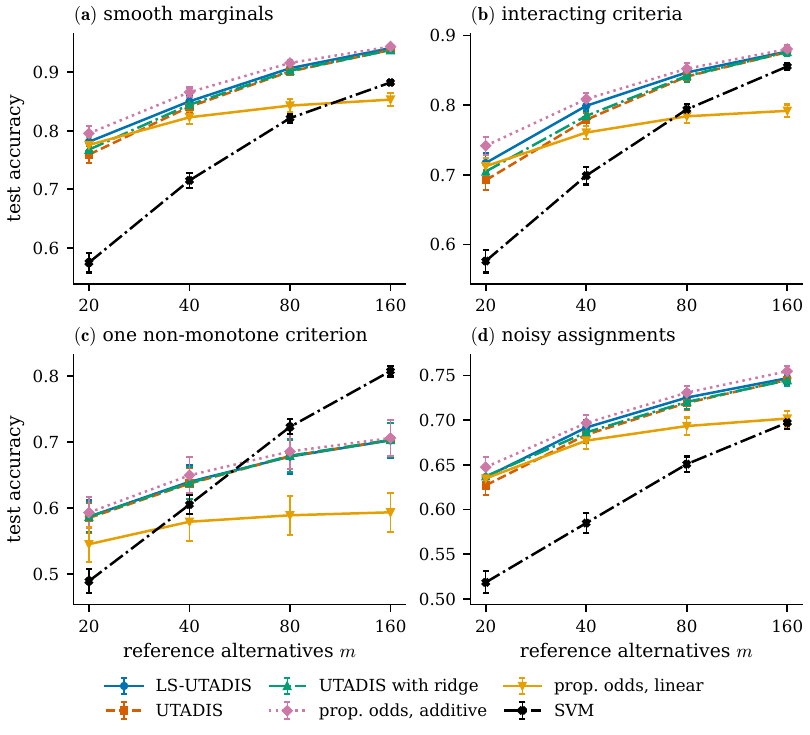}
\caption{Simulations under misspecification and noise: test accuracy against the number of reference alternatives, with 95\% confidence intervals over 100 populations. The additive value methods are close to one another, with proportional odds on the additive value at or above the others. Proportional odds with a linear index is above UTADIS at the smallest size under smooth marginals, interacting criteria and noisy assignments, and falls behind the additive value methods as the reference set grows; with one non-monotone criterion it is behind them from the smallest size. The support vector machine overtakes the additive value methods only when one criterion is non-monotone and the reference set is large. (a) Smooth marginals. (b) Interacting criteria. (c) One non-monotone criterion. (d) Noisy assignments.}
\label{fig:sim}
\end{figure}

Table~\ref{tab:sim} shows some results, with mislabelling of 2 or more categories, when the true marginals are additive, and others, without mislabelling, are shown in Figure~\ref{fig:sim}. For $m=\simMsmall$ and \simMsecond, LS-UTADIS is \simGainLsuUtaAddSmall{} points (\simGainLsuUtaAddSmallSE) and \simGainLsuUtaAddSecond{} points (\simGainLsuUtaAddSecondSE) better than UTADIS, for additive marginals, some of this, \simGainUtaRidgeUtaAddSmall{} points and \simGainUtaRidgeUtaAddSecond{} points, being due to ridge terms, for $\rho=\eta=10^{-4}$. UTADIS-MM is \addGainLsuUtaMMSmall{} points lower ($m=\simMsmall$), but differences of LS-UTADIS are \mmFromSecondMax{} points or less on both types of data in all other cases: there are \mmFromSecondCells{} comparisons, of which \mmFromSecondLsuWins, additively, $m=\mmFromSecondLsuWinM$, is significantly in favour of LS-UTADIS, and none significantly against. Over all \simCells{} comparisons with various other generating rules, LS-UTADIS is significantly higher than UTADIS for \simGainLsuUtaWins, never significantly less, higher than UTADIS with ridge on \simGainLsuUtaRidgeWins{} occasions, never less, and higher than UTADIS-MM on \simGainLsuUtaMMWins{} occasions, never less. All differences are small, from \simGainLsuUtaMin{} to \simGainLsuUtaMax{} points against UTADIS. Regularised UTADIS and UTADIS behave very similarly: LS-UTADIS is higher, never lower, for \simGainLsuUtaRegWins{} out of \simCells{} comparisons, by up to \simGainLsuUtaRegMax{} points, and regularised UTADIS differs from UTADIS by up to \magn{\simGainUtaRegUtaMin} points unfavourably, \simGainUtaRegUtaMax{} points favourably (significantly for \simGainUtaRegUtaLosses{} comparisons, all in favour of UTADIS).

All data sets admit a solution with the correct number of sub-intervals, if additive, so either centre could be found. LS-UTADIS differs from the Chebyshev centre by between \addGainLsuChebMin{} and \addGainLsuChebMax{} points, depending on $m$, non-significantly, while on \addGainLsuChebLosses{} occasion the centre is significantly higher. Differences from the analytic centre are between \addGainLsuAcenMin{} and \addGainLsuAcenMax{} points, none significantly favourable to LS-UTADIS. At $m=\simMsmall$, the Chebyshev centre exceeds UTADIS by \addGainChebUtaSmall{} points (\addGainChebUtaSmallSE). Any failure to fit for a proposed number of sub-intervals, on any subset in cross-validation, has been assumed a misclassification. These include failures to show existence of an interior point to the relevant polyhedron, and failures of the code to determine a solution, totaling \cenChebInnerFail{} and \cenAcenInnerFail{} for the Chebyshev and analytic centres respectively for all data, of which \cenChebQsInnerFail{} and \cenAcenQsInnerFail{} on QS. In view of identical subsets and hence polyhedra, and the fact that an interior always exists on QS, these are presumably due to failure of the algorithm. For \cenFallback{} out of \cenFits{} finally selected numbers of sub-intervals, no solution could be found with all examples, so the next lowest was taken, but all finally selected numbers were successfully fitted.

Both forms of penalty are inferior to the likelihood. Using proportional odds with additive value, it exceeds LS-UTADIS significantly for \simGainLsuOlogALosses{} out of \simCells{} generating rules and $m$'s, and never is lower (differences from \simGainLsuOlogAMin{} to \simGainLsuOlogAMax{} points), and exceeds UTADIS for \simGainOlogAUtaWins. Once marginals are non-linear, proportional odds with linear index deteriorates: best \simAddOlogLarge{} ($m=\simMlarge$), additively, against \simAddLsuLarge{} for LS-UTADIS, \simAddOlogALarge{} for proportional odds.

\begin{table}[t]
\centering
\caption{Simulations: test accuracy (means over 100 populations) in the well-specified additive scenario and under distant mislabelling, where 10\% of the alternatives carry a label at least two categories away from the true one; for the latter, accuracy is measured against the true categories. The centres of the compatible models are defined only for reference sets that the model can reproduce and were run in the additive scenario only. Half-widths of the 95\% confidence intervals are at most \simMaxHalfWidth. Best mean per column in bold. Abbreviations: LS-UTADIS, least-squares UTADIS; UTADIS-MM, UTADIS with a max-margin second stage; centred LS, centred least squares; SVM (RBF), support vector machine with a Gaussian kernel.}
\label{tab:sim}
\small
\setlength{\tabcolsep}{4pt}
\resizebox{\textwidth}{!}{%
\begin{tabular}{@{}lcccccccc@{}}
\toprule
 & \multicolumn{4}{c}{Additive, by $m$} & \multicolumn{4}{c}{Distant mislabelling, by $m$} \\
\cmidrule(lr){2-5}\cmidrule(l){6-9}
Method & 20 & 40 & 80 & 160 & 20 & 40 & 80 & 160 \\
\midrule
LS-UTADIS & 0.752 & 0.852 & 0.919 & 0.955 & 0.625 & 0.701 & 0.767 & 0.801 \\
LS-UTADIS, no recalibration & 0.750 & 0.855 & 0.918 & 0.955 & 0.621 & 0.678 & 0.730 & 0.771 \\
Relative margin & 0.727 & 0.838 & 0.905 & 0.952 & 0.611 & 0.693 & 0.750 & 0.796 \\
Centred LS & 0.585 & 0.629 & 0.648 & 0.653 & 0.517 & 0.522 & 0.524 & 0.523 \\
Huber shortfalls & 0.745 & 0.845 & 0.914 & 0.954 & 0.674 & \textbf{0.768} & 0.842 & \textbf{0.895} \\
LS-UTADIS, all thresholds & 0.750 & 0.852 & 0.919 & 0.955 & 0.612 & 0.677 & 0.739 & 0.764 \\
UTADIS & 0.724 & 0.829 & 0.907 & 0.952 & \textbf{0.677} & 0.764 & 0.842 & 0.893 \\
UTADIS with ridge & 0.734 & 0.835 & 0.907 & 0.952 & \textbf{0.677} & 0.760 & 0.842 & 0.893 \\
UTADIS-MM & 0.731 & 0.843 & 0.913 & 0.952 & 0.617 & 0.696 & 0.771 & 0.817 \\
Regularised UTADIS & 0.730 & 0.830 & 0.906 & 0.952 & 0.648 & 0.765 & \textbf{0.846} & 0.894 \\
UTADIS, all thresholds & 0.734 & 0.828 & 0.904 & 0.952 & 0.653 & 0.734 & 0.811 & 0.858 \\
Chebyshev centre & 0.757 & 0.861 & 0.919 & 0.954 & -- & -- & -- & -- \\
Analytic centre & 0.753 & 0.856 & 0.916 & 0.954 & -- & -- & -- & -- \\
Proportional odds, additive value & \textbf{0.758} & \textbf{0.868} & \textbf{0.922} & \textbf{0.959} & 0.659 & 0.739 & 0.821 & 0.868 \\
Proportional odds, linear index & 0.729 & 0.780 & 0.797 & 0.805 & 0.648 & 0.723 & 0.770 & 0.791 \\
SVM (RBF) & 0.578 & 0.706 & 0.796 & 0.856 & 0.506 & 0.601 & 0.703 & 0.782 \\
Random forest & 0.565 & 0.648 & 0.727 & 0.793 & 0.515 & 0.596 & 0.678 & 0.748 \\
\bottomrule
\end{tabular}}
\end{table}

For mislabelled data, results, not surprisingly, are much better for UTADIS. On the original scale of categories, LS-UTADIS is lower than UTADIS by \magn{\flipsGainLsuUtaMin} to \magn{\flipsGainLsuUtaMax} points, more variable than in other parts of this work, and only attains \flipsCleanLsuLarge{} ($m=\simMlarge$), compared with \flipsCleanUtaLarge. Considerable improvement follows recalibration (although best is only \flipsCleanLsuJLarge, without recalibration), and one reason may be that mislabelled observations induce incorrect marginals. UTADIS-MM deteriorates by similar magnitudes, best being \flipsCleanUtaMMLarge{} ($m=\simMlarge$) and up to \magn{\flipsGainLsuUtaMMMin} points different from LS-UTADIS, significantly once. It has no margin for initial estimation, which may well be important. Regularised UTADIS is much less adversely affected: best \flipsCleanUtaRegLarge, differing by up to \magn{\flipsGainUtaRegUtaMin} points from UTADIS, significantly once, never in its favour. Proportional odds lies intermediate, best \flipsCleanOlogALarge, always exceeding LS-UTADIS but \flipsLossOlogAUtaMin{} to \flipsLossOlogAUtaMax{} points less than UTADIS. Thus all best methods include an absolute penalty and some tuned parameter (margin or penalties), but not an absolute penalty alone. Presumably in practice there would be no indication of this degree of misclassification, but it argues for including a tuned margin if such errors are possible.

An additive rule is best, for one or other form, over all $m$, under either form of error in specifying marginals, smooth in each criterion (Figure~\ref{fig:sim}a and b). Thus there is considerable tolerance of errors in the value function. However, with one non-monotonic marginal utility, a support vector machine does rather better, \simNonmonSvmLarge{} ($m=\simMlarge$), compared with \simNonmonLsuLarge, although all forms of additive rule are better for $m=\simMsmall$. Clearly many extensions could be considered if the DM's utilities were non-monotonic \citep[for example,][]{kadzinski2020partial,liu2019regularization}.

\subsection{Large degree of randomness in category: published data}
\label{sec:openml}

Results for the various published data sets are given in Table~\ref{tab:openml}; none exceeds \openmlMaxHalfWidth{} as half-width of the 95 per cent confidence interval. No method achieves better than \openmlAccMax. If each observation were simply given the most frequent category, accuracies would be \majEraAcc, \majEslAcc, \majLevAcc{} and \majSwdAcc, with average error \majEraMae, \majEslMae, \majLevMae{} and \majSwdMae{} categories, for ERA, ESL, LEV and SWD. Thus results are only marginally better for ERA. Proportional odds with linear index exceeds LS-UTADIS, significantly, for \openmlGainLsuOlogLosses{} out of \openmlCells{} data sets and $m$, by up to \magn{\openmlGainLsuOlogMin} points, never more than \openmlGainLsuOlogMax{} point; proportional odds does so for \openmlGainLsuOlogALosses{} out of \openmlCells, by up to \magn{\openmlGainLsuOlogAMin} points, and exceeds UTADIS on \openmlGainOlogAUtaWins{} occasions, never being less. Thus something in addition to the linear index accounts for its better results. There is little indication that squaring is beneficial: LS-UTADIS exceeds UTADIS significantly once only, but is significantly worse on \openmlGainLsuUtaLosses{} occasions. Against UTADIS with ridge it is never significantly better, and significantly worse for \openmlGainLsuUtaRidgeLosses{} out of \openmlCells{} (differences from \openmlGainLsuUtaRidgeMax{} to \openmlGainLsuUtaRidgeMin{} points), and differs significantly from regularised UTADIS on \openmlGainLsuUtaRegWins{} occasions in either direction, from \openmlGainLsuUtaRegMax{} to \openmlGainLsuUtaRegMin{} points. \citet{liu2021interacting} found a regularised quadratic to be superior to UTADIS for some sets, presumably as a result of the regularization. However there is no significant superiority of LS-UTADIS over regularisation of the absolute penalty. Differences of regularised UTADIS from UTADIS range from \openmlGainUtaRegUtaMin{} to \openmlGainUtaRegUtaMax{} points, significantly in its favour for \openmlGainUtaRegUtaWins{} set, in favour of UTADIS \openmlGainUtaRegUtaLosses{} times.

\begin{table}[t]
\centering
\caption{Public ordinal data sets: test accuracy and mean absolute category error (means over 100 reference sets of size $m$). Best mean per column in bold. Abbreviations: LS-UTADIS, least-squares UTADIS; UTADIS-MM, UTADIS with a max-margin second stage; centred LS, centred least squares; SVM (RBF), support vector machine with a Gaussian kernel.}
\label{tab:openml}
\footnotesize
\setlength{\tabcolsep}{2.4pt}
\resizebox{\textwidth}{!}{%
\begin{tabular}{@{}lcccccccccccc@{}}
\toprule
 & \multicolumn{3}{c}{ERA} & \multicolumn{3}{c}{ESL} & \multicolumn{3}{c}{LEV} & \multicolumn{3}{c}{SWD} \\
\cmidrule(lr){2-4}\cmidrule(lr){5-7}\cmidrule(lr){8-10}\cmidrule(lr){11-13}
Method & 50 & 100 & 200 & 50 & 100 & 200 & 50 & 100 & 200 & 50 & 100 & 200 \\
\midrule
\multicolumn{13}{@{}l}{\emph{Accuracy}} \\
LS-UTADIS & 0.223 & 0.240 & 0.253 & 0.631 & 0.661 & 0.681 & 0.551 & 0.575 & 0.588 & 0.515 & 0.535 & 0.552 \\
LS-UTADIS, no recalibration & 0.215 & 0.218 & 0.218 & 0.630 & 0.667 & 0.681 & 0.539 & 0.567 & 0.575 & 0.502 & 0.531 & 0.543 \\
Relative margin & 0.221 & 0.240 & 0.253 & 0.621 & 0.651 & 0.672 & 0.545 & 0.571 & 0.587 & 0.513 & 0.533 & 0.551 \\
Centred LS & 0.178 & 0.175 & 0.176 & 0.585 & 0.586 & 0.556 & 0.418 & 0.413 & 0.415 & 0.351 & 0.373 & 0.369 \\
Huber shortfalls & 0.226 & 0.240 & 0.250 & 0.638 & 0.663 & 0.688 & 0.559 & 0.583 & 0.595 & 0.512 & 0.538 & 0.558 \\
LS-UTADIS, all thresholds & 0.232 & 0.244 & 0.252 & 0.631 & 0.662 & 0.682 & 0.549 & 0.575 & 0.586 & 0.518 & 0.534 & 0.554 \\
UTADIS & 0.227 & 0.244 & 0.248 & 0.625 & 0.663 & 0.685 & 0.557 & 0.582 & 0.594 & 0.512 & 0.540 & 0.557 \\
UTADIS with ridge & 0.229 & 0.242 & 0.250 & 0.637 & 0.666 & 0.690 & 0.559 & 0.584 & 0.595 & 0.511 & 0.539 & 0.557 \\
UTADIS-MM & 0.223 & 0.241 & 0.250 & 0.621 & 0.657 & 0.676 & 0.554 & 0.581 & 0.593 & 0.508 & 0.534 & 0.550 \\
Regularised UTADIS & 0.229 & 0.247 & 0.256 & 0.630 & 0.659 & 0.687 & 0.542 & 0.575 & 0.588 & 0.508 & 0.536 & 0.558 \\
UTADIS, all thresholds & 0.236 & \textbf{0.256} & \textbf{0.261} & 0.626 & 0.661 & 0.690 & 0.562 & 0.584 & 0.594 & 0.519 & 0.544 & 0.560 \\
Proportional odds, additive value & 0.239 & 0.249 & 0.251 & 0.649 & 0.679 & \textbf{0.698} & 0.570 & 0.592 & 0.602 & \textbf{0.529} & \textbf{0.551} & \textbf{0.567} \\
Proportional odds, linear index & \textbf{0.243} & 0.251 & 0.254 & \textbf{0.651} & \textbf{0.681} & 0.696 & \textbf{0.581} & \textbf{0.597} & \textbf{0.606} & 0.514 & 0.547 & 0.559 \\
SVM (RBF) & 0.193 & 0.209 & 0.229 & 0.582 & 0.626 & 0.665 & 0.501 & 0.540 & 0.576 & 0.479 & 0.519 & 0.539 \\
Random forest & 0.196 & 0.208 & 0.221 & 0.564 & 0.600 & 0.636 & 0.493 & 0.524 & 0.550 & 0.491 & 0.510 & 0.517 \\
\addlinespace
\multicolumn{13}{@{}l}{\emph{Mean absolute category error}} \\
LS-UTADIS & 1.53 & 1.41 & 1.33 & 0.39 & 0.36 & 0.34 & 0.50 & 0.47 & 0.45 & 0.53 & 0.50 & 0.48 \\
LS-UTADIS, no recalibration & 1.70 & 1.67 & 1.66 & 0.39 & 0.35 & 0.34 & 0.52 & 0.48 & 0.47 & 0.56 & 0.51 & 0.50 \\
Relative margin & 1.54 & 1.40 & 1.31 & 0.41 & 0.37 & 0.34 & 0.50 & 0.47 & 0.45 & 0.54 & 0.50 & 0.48 \\
Centred LS & 2.03 & 2.04 & 2.00 & 0.46 & 0.46 & 0.48 & 0.72 & 0.73 & 0.72 & 0.83 & 0.78 & 0.78 \\
Huber shortfalls & 1.52 & 1.43 & 1.36 & 0.39 & 0.36 & 0.33 & 0.49 & 0.46 & 0.44 & 0.54 & 0.50 & 0.47 \\
LS-UTADIS, all thresholds & 1.48 & 1.39 & 1.34 & 0.39 & 0.36 & 0.33 & 0.50 & 0.47 & 0.45 & 0.53 & 0.50 & 0.47 \\
UTADIS & 1.51 & 1.40 & 1.36 & 0.40 & 0.36 & 0.33 & 0.49 & 0.46 & 0.44 & 0.54 & 0.50 & 0.47 \\
UTADIS with ridge & 1.51 & 1.42 & 1.35 & 0.39 & 0.35 & 0.33 & 0.49 & 0.46 & 0.44 & 0.54 & 0.50 & 0.47 \\
UTADIS-MM & 1.52 & 1.40 & 1.36 & 0.41 & 0.36 & 0.34 & 0.49 & 0.46 & 0.45 & 0.55 & 0.50 & 0.48 \\
Regularised UTADIS & 1.41 & 1.31 & 1.26 & 0.40 & 0.36 & 0.33 & 0.51 & 0.47 & 0.45 & 0.54 & 0.50 & 0.47 \\
UTADIS, all thresholds & 1.43 & 1.34 & 1.31 & 0.40 & 0.36 & 0.33 & 0.48 & 0.46 & 0.44 & 0.53 & 0.49 & 0.47 \\
Proportional odds, additive value & 1.33 & 1.28 & 1.26 & \textbf{0.37} & \textbf{0.34} & \textbf{0.32} & 0.47 & 0.45 & \textbf{0.43} & \textbf{0.51} & \textbf{0.47} & \textbf{0.45} \\
Proportional odds, linear index & \textbf{1.30} & \textbf{1.26} & \textbf{1.24} & \textbf{0.37} & \textbf{0.34} & \textbf{0.32} & \textbf{0.46} & \textbf{0.44} & \textbf{0.43} & 0.53 & 0.48 & 0.46 \\
SVM (RBF) & 1.66 & 1.54 & 1.45 & 0.48 & 0.41 & 0.36 & 0.57 & 0.52 & 0.47 & 0.58 & 0.52 & 0.50 \\
Random forest & 1.68 & 1.62 & 1.55 & 0.49 & 0.44 & 0.39 & 0.59 & 0.55 & 0.51 & 0.57 & 0.55 & 0.54 \\
\bottomrule
\end{tabular}}
\end{table}

Of interest for these solutions: (a) limits would be changed in \openmlRecalShare{} per cent of cases, and the loss of accuracy from prohibiting recalibration is as much as \openmlGainLsuLsuJMax{} points. There must therefore be substantial changes to the limits, plausibly because many distant observations would each receive a large, squared penalty. Second, $\delta$ is often large relative to widths; 1 or more widths (whether with observations inside them or not) is less than twice $\delta$ for \openmlCrossedMin{} to \openmlCrossedMax{} per cent of solutions. A more pertinent quantity might be the number of observations falling between crossed zone ends. For extreme values of $m$, these are given in Section~\ref{sec:time} by data set; for these, LS-UTADIS behaves similarly to centring, while for all other observations it behaves similarly to using a margin.

\subsection{Huber and all-threshold shortfalls}
\label{sec:v08}

Tables~\ref{tab:qs}, \ref{tab:sim} and~\ref{tab:openml} include results for Huber and each of the 2 versions of all-threshold shortfalls, for all reference sets used for any other method. There were no failures in the \vFits{} cases with tuning, or \vfixFits{} without, in this and the following Section.

\begin{figure}[t]
\centering
\includegraphics{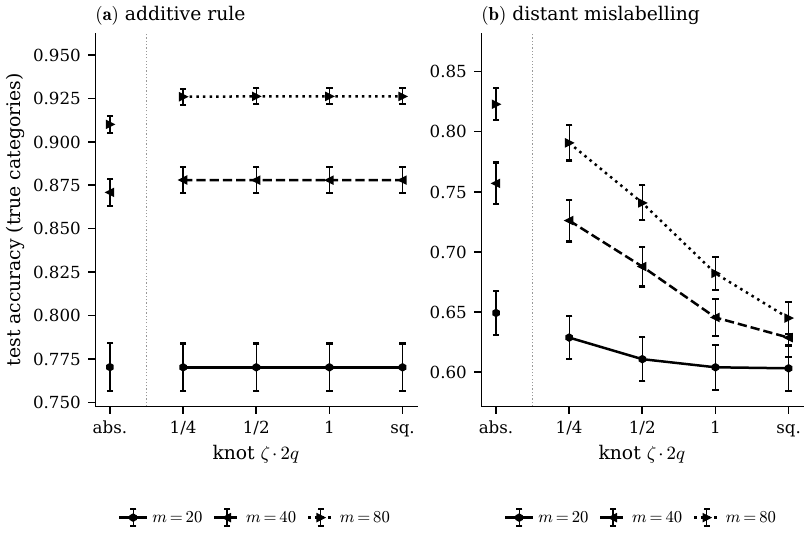}
\caption{Accuracy for Huber shortfalls with no tuning, $d=0.25$, $\rho=\eta=10^{-4}$ and 3 sub-intervals, over 100 reference sets, with 95 per cent confidence intervals, for knots of $\tfrac14$, $\tfrac12$ and 1 times $1/(2q)$, continued to correspond to squared shortfalls; LS-UTADIS, denoted sq., coincides for sufficiently large knots; also absolute shortfalls plus ridge terms, denoted abs., which are not limiting values. (a) Additive rule to generate categories. (b) Mislabeling; accuracy relative to actual categories.}
\label{fig:huber}
\end{figure}

For all generating rules within the class, Huber shortfalls are a little worse than LS-UTADIS. On QS, for $m=\qsMsmall$ they are lower by \magn{\vqsGainHubLsuSmall} points on average (\vqsGainHubLsuSmallSE), by amounts ranging from \magn{\vqsGainHubLsuMax} to \magn{\vqsGainHubLsuMin} over the 5 $m$, significantly so for \vqsGainHubLsuLosses; additively, for $m=\simMsmall$ lower on average by \magn{\vaddGainHubLsuSmall} points (\vaddGainHubLsuSmallSE), ranging from \magn{\vaddGainHubLsuMax} to \magn{\vaddGainHubLsuMin} over the 4 $m$ and significantly so for all. As expected they are considerably better than UTADIS, by \vqsGainHubUtaSmall{} points (\vqsGainHubUtaSmallSE) on QS, \vaddGainHubUtaSmall{} points (\vaddGainHubUtaSmallSE) additively, for $m=\qsMsmall$. For the remaining scenarios without mislabelling, differences from LS-UTADIS range from \vmisGainHubLsuMin{} to +\vmisGainHubLsuMax{} points over the \vmisCells{} combinations, significantly negative for \vmisGainHubLsuLosses{} and positive for \vmisGainHubLsuWins. With no tuning there was no effect, to 1 place of decimal accuracy, of including the parameter to locate a knot, either on QS with only 1 sub-interval, not plotted, or additively (Figure~\ref{fig:huber}a). Hence some of the deterioration with tuning could be due to selecting from 72 rather than 24 possibilities when there are few divisions. A similar caution was expressed in relation to tuning the likelihood and centres (Section~\ref{sec:fixed}).

For badly mislabelled observations, Huber shortfalls are substantially better than LS-UTADIS. Relative to the actual categories, they are on par with UTADIS, ranging from \vflipsGainHubUtaMin{} to \vflipsGainHubUtaMax{} points and never significantly different, and better than LS-UTADIS by \vflipsGainHubLsuMin{} to \vflipsGainHubLsuMax{} points, always significantly. They also exceed proportional odds, evaluated additively, by \vflipsGainHubOlogAMin{} to \vflipsGainHubOlogAMax{} points. The least knot was preferred for \hubKnotSmallestShareFlips{} per cent of the reference sets, compared with \hubKnotSmallestShare{} per cent over all the experiments, with ties counted for the least. For the published data, there are significant differences in favour of Huber shortfalls relative to LS-UTADIS for \vopenmlGainHubLsuWins{} out of \openmlCells{} combinations (\vopenmlGainHubLsuWinsBH{} after the adjustment described above), and none in favour of LS-UTADIS. They range from \vopenmlGainHubUtaMin{} to \vopenmlGainHubUtaMax{} points relative to UTADIS.

Thus squaring only the smaller of the shortfalls retains much of the benefit, when fitting is possible, and all of the resistance to very erroneous observations (apart from up to \hubFlipsLossMax{} points), of not squaring. This lends some support to, but does not verify, the suggestion of Section~\ref{sec:fixed} that squaring is beneficial in determining which of several equally good models is preferred, and undesirable in the greater emphasis placed on the largest shortfalls.

Figure~\ref{fig:huber} plots corresponding quantities without tuning, for each of 3 locations of the knot, also plotting absolute shortfalls plus ridge penalty and LS-UTADIS, with $d=0.25$ and $\rho$ and $\eta$ both $10^{-4}$ and 3 sub-intervals, for reference sets of Section~\ref{sec:fixed} generated additively, and similar numbers from the scenario with mislabeling. As before, there is no difference between knots to 1 decimal place, either additively or on QS with 1 sub-interval, not plotted. With substantial mislabeling, all 3 exceed LS-UTADIS, by \vfixFlipsHubQuarterLsuMin{} to \vfixFlipsHubQuarterLsuMax{} points for the smallest and \vfixFlipsHubWholeLsuMin{} to \vfixFlipsHubWholeLsuMax{} points for the greatest knot. Absolute shortfalls plus ridge terms exceed it by \vfixFlipsUtaRidgeLsuMin{} to \vfixFlipsUtaRidgeLsuMax{} points. Clearly there can be no difference with correctly labelled observations provided none of the shortfalls of LS-UTADIS is greater than the knot, since then it also minimises the Huber criterion (Proposition~\ref{prop:huber}(ii)); otherwise, under mislabelling, accuracy decreases towards that of LS-UTADIS with increasing knot. Note that the plotted absolute shortfalls plus ridge terms are not limiting values corresponding to a vanishing knot, since the weights on the ridge terms would not tend to 0; we simply wished to include them. It thus seems prudent to use small knots if grossly erroneous observations are anticipated.

All-threshold shortfalls show little change from the simpler form, when generated according to the model: means differ by at most \atConsistentAbsMax{} points on QS and additively, for either form and any $m$. Both forms are deteriorated by mislabeling: relative to actual categories they are lower by \magn{\vflipsGainLsuAtLsuMax} to \magn{\vflipsGainLsuAtLsuMin} points if squared, \magn{\vflipsGainUtaAtUtaMax} to \magn{\vflipsGainUtaAtUtaMin} if not, and mean error, relative to the given categories, is greater by \vflipsMaeGainLsuAtLsuMin{} to \vflipsMaeGainLsuAtLsuMax{} and \vflipsMaeGainUtaAtUtaMin{} to \vflipsMaeGainUtaAtUtaMax{} per 100 categories, respectively. Only reference alternatives more than 1 category away from their assigned category are affected, by the addition of 1 or more terms (Proposition~\ref{prop:huber}(iii)), tending on average to be more heavily mislabeled observations.

For the available data, using the non-squared penalty improves accuracy. Significantly better than UTADIS for \vopenmlGainUtaAtUtaWins{} of the \openmlCells{} sets of cells, \vopenmlGainUtaAtUtaWinsBH{} after adjustment, not significantly worse for any; they vary from \vopenmlGainUtaAtUtaMin{} to \vopenmlGainUtaAtUtaMax{} points. It gives best accuracy overall on ERA for 100 and 200 observations. The mean category error is also significantly reduced for \vopenmlMaeGainUtaAtUtaLosses{} sets of cells, increased for none. The corresponding comparisons for the squared version with LS-UTADIS give \vopenmlGainLsuAtLsuWins{} cell of significant improvement, none of deterioration; reduction of mean category error for \vopenmlMaeGainLsuAtLsuLosses{} cells, none increased. The comparisons over the 3 data sets of Section~\ref{sec:further} give \vuciGainUtaAtUtaWins{} cell of improvement (\vuciGainUtaAtUtaWinsBH{} after adjustment for the non-squared version, \vuciGainLsuAtLsuWinsBH{} for the squared) and none of deterioration, for each version compared with its immediate-threshold counterpart. Hence all thresholds are an improvement for the non-squared penalty over these judged data, but a deterioration if labels may be far in error. We have been unable to explain why the direction is different. As noted above it would relate to the proportion of reference alternatives more than 1 category away from their estimated positions, which we did not save; all mean errors are high on ERA, where improvement is most marked (Table~\ref{tab:openml}). It has recently been pointed out \citep{yamasaki2024remarks} that estimation with a piecewise linear function tends to produce many identical fitted values, adversely affecting the ease of assigning to categories; it might therefore be advantageous to spread the non-squared penalty over all the limits. We have not investigated the pattern of fitted values. Note also that only the population minimiser has been shown consistent for all thresholds, for absolute category error \citep{pedregosa2017consistency}, and without restriction on the range of values that may be taken by the scores. Here all scores are restricted to be an additive value, limits lying between 0 and 1, and sample sizes are modest.

\subsection{Other published data sets, varying criteria and categories}
\label{sec:further}

\begin{table}[t]
\centering
\caption{Three further monotone data sets: test accuracy and mean absolute category error (means over 100 reference sets of size $m$). CEV, car evaluation (labels from a known hierarchical rule); CPU, computer hardware, and MPG, cars by fuel economy (quartile categories). Best mean per column in bold. Abbreviations: LS-UTADIS, least-squares UTADIS; UTADIS-MM, UTADIS with a max-margin second stage; centred LS, centred least squares; SVM (RBF), support vector machine with a Gaussian kernel.}
\label{tab:uci}
\footnotesize
\setlength{\tabcolsep}{2.4pt}
\resizebox{\textwidth}{!}{%
\begin{tabular}{@{}lccccccccc@{}}
\toprule
 & \multicolumn{3}{c}{CEV} & \multicolumn{3}{c}{CPU} & \multicolumn{3}{c}{MPG} \\
\cmidrule(lr){2-4}\cmidrule(lr){5-7}\cmidrule(lr){8-10}
Method & 20 & 50 & 100 & 20 & 50 & 100 & 20 & 50 & 100 \\
\midrule
\multicolumn{10}{@{}l}{\emph{Accuracy}} \\
LS-UTADIS & 0.724 & \textbf{0.830} & 0.882 & 0.590 & \textbf{0.652} & 0.669 & 0.708 & 0.748 & 0.763 \\
Huber shortfalls & 0.726 & 0.827 & 0.880 & 0.583 & 0.640 & 0.667 & 0.707 & 0.751 & 0.767 \\
LS-UTADIS, all thresholds & 0.724 & 0.827 & \textbf{0.883} & 0.589 & 0.649 & \textbf{0.675} & 0.708 & 0.749 & 0.761 \\
Relative margin & 0.726 & 0.802 & 0.870 & 0.586 & 0.643 & 0.673 & 0.684 & 0.736 & 0.755 \\
Centred LS & 0.709 & 0.725 & 0.732 & 0.520 & 0.519 & 0.515 & 0.627 & 0.633 & 0.641 \\
UTADIS & 0.719 & 0.817 & 0.869 & 0.576 & 0.636 & 0.659 & 0.700 & 0.750 & 0.768 \\
UTADIS with ridge & 0.726 & 0.812 & 0.872 & 0.577 & 0.636 & 0.657 & 0.708 & 0.750 & 0.767 \\
UTADIS-MM & 0.731 & 0.808 & 0.867 & 0.599 & 0.645 & 0.666 & 0.691 & 0.743 & 0.756 \\
Regularised UTADIS & 0.715 & 0.809 & 0.870 & 0.567 & 0.635 & 0.656 & 0.678 & 0.751 & 0.769 \\
UTADIS, all thresholds & 0.719 & 0.820 & 0.874 & 0.580 & 0.637 & 0.665 & 0.699 & 0.748 & 0.768 \\
Proportional odds, additive value & \textbf{0.755} & 0.829 & 0.880 & \textbf{0.611} & 0.649 & 0.668 & \textbf{0.718} & \textbf{0.761} & \textbf{0.774} \\
Proportional odds, linear index & 0.702 & 0.764 & 0.779 & 0.569 & 0.633 & 0.657 & 0.692 & 0.751 & 0.768 \\
SVM (RBF) & 0.735 & 0.816 & 0.867 & 0.496 & 0.600 & 0.647 & 0.643 & 0.723 & 0.756 \\
Random forest & 0.738 & 0.802 & 0.848 & 0.562 & 0.634 & 0.669 & 0.672 & 0.736 & 0.768 \\
\addlinespace
\multicolumn{10}{@{}l}{\emph{Mean absolute category error}} \\
LS-UTADIS & 0.32 & \textbf{0.19} & \textbf{0.12} & 0.45 & \textbf{0.37} & 0.35 & 0.30 & 0.26 & 0.24 \\
Huber shortfalls & 0.32 & \textbf{0.19} & 0.13 & 0.46 & 0.38 & 0.35 & 0.31 & 0.26 & 0.24 \\
LS-UTADIS, all thresholds & 0.32 & \textbf{0.19} & \textbf{0.12} & 0.45 & \textbf{0.37} & \textbf{0.34} & 0.30 & 0.26 & 0.24 \\
Relative margin & 0.33 & 0.22 & 0.14 & 0.46 & 0.38 & 0.35 & 0.33 & 0.27 & 0.25 \\
Centred LS & 0.38 & 0.34 & 0.33 & 0.61 & 0.60 & 0.60 & 0.42 & 0.40 & 0.39 \\
UTADIS & 0.34 & 0.20 & 0.14 & 0.47 & 0.39 & 0.36 & 0.31 & 0.26 & 0.24 \\
UTADIS with ridge & 0.32 & 0.21 & 0.14 & 0.46 & 0.39 & 0.36 & 0.30 & 0.26 & 0.24 \\
UTADIS-MM & 0.32 & 0.22 & 0.14 & 0.44 & 0.38 & 0.35 & 0.33 & 0.26 & 0.25 \\
Regularised UTADIS & 0.35 & 0.21 & 0.14 & 0.48 & 0.39 & 0.36 & 0.34 & 0.26 & 0.24 \\
UTADIS, all thresholds & 0.33 & 0.20 & 0.13 & 0.46 & 0.39 & 0.36 & 0.32 & 0.26 & 0.24 \\
Proportional odds, additive value & \textbf{0.28} & \textbf{0.19} & 0.13 & \textbf{0.41} & \textbf{0.37} & 0.35 & \textbf{0.29} & \textbf{0.24} & \textbf{0.23} \\
Proportional odds, linear index & 0.36 & 0.26 & 0.24 & 0.50 & 0.39 & 0.36 & 0.33 & 0.26 & 0.24 \\
SVM (RBF) & 0.34 & 0.23 & 0.16 & 0.63 & 0.46 & 0.39 & 0.39 & 0.29 & 0.25 \\
Random forest & 0.33 & 0.24 & 0.18 & 0.50 & 0.40 & 0.36 & 0.35 & 0.28 & 0.24 \\
\bottomrule
\end{tabular}}
\end{table}

Table~\ref{tab:uci} reports on 3 data sets, \uciReps{} reference sets generated for each of \uciMSmall, \uciMSecond{} and \uciMLarge{} observations; all half-widths of 95 per cent confidence intervals at most \uciMaxHalfWidth. LS-UTADIS is better than UTADIS significantly for \vuciGainLsuUtaWins{} cells (\vuciGainLsuUtaWinsBH{} after adjustment) out of \vuciCells, by up to \vuciGainLsuUtaMax{} points, and worse significantly for \vuciGainLsuUtaLosses. For \uciMSmall{} observations the differences (SEs) are \vuciGainLsuUtaCevSmall{} (\vuciGainLsuUtaCevSmallSE), \vuciGainLsuUtaCpuSmall{} (\vuciGainLsuUtaCpuSmallSE), \vuciGainLsuUtaMpgSmall{} (\vuciGainLsuUtaMpgSmallSE) points for CEV, CPU and MPG respectively. Thus squaring is advantageous for 2 rules not known to be additive and disadvantaged for 1. In particular, no additive function with all category widths at least 0.01 would reproduce the assignments on CEV. Considering all \nCev{} cars and allowing 3 sub-intervals, more than sufficient, with minimum width 0.01, minimum total shortfall was \cevShortfall, with margin $10^{-6}$. If all cars were assigned correctly to their own category, the total shortfall would be no greater than \cevBound. The minimum shortfall varies approximately directly with the minimum allowed width (\cevShortfallMilli{} when this is $10^{-3}$), so a close approximation to the actual rule requires much narrower categories. It is noteworthy that on CEV the differences are significant only for 2 larger sample sizes, contrasting with QS and additive generation, for which the smallest sample gives greatest difference. We did not record whether there were compatible models, so cannot comment further on the cause. Proportional odds, additively evaluated, gives best accuracy for \uciMSmall{} observations for each data set, \vuciCevOlogASmall, \vuciCpuOlogASmall, \vuciMpgOlogASmall, compared with \vuciCevLsuSmall, \vuciCpuLsuSmall, \vuciMpgLsuSmall{} for LS-UTADIS; the difference (LS-UTADIS minus proportional odds) ranges from \vuciGainLsuOlogAMin{} to \vuciGainLsuOlogAMax{} points, never significantly positive. Differences for Huber and all-threshold from their counterparts range from \vuciGainHubLsuMin{} to \vuciGainHubLsuMax{} points, and \vuciGainLsuAtLsuMin{} to \vuciGainLsuAtLsuMax{} points if squared and \vuciGainUtaAtUtaMin{} to \vuciGainUtaAtUtaMax{} if not. LS-UTADIS exceeds UTADIS-MM significantly for \vuciGainLsuUtaMMWins{} cells, is below it for \vuciGainLsuUtaMMLosses.

\begin{table}[t]
\centering
\caption{Other numbers of criteria and categories: test accuracy (means over 100 populations) in the additive scenario and under distant mislabelling (accuracy against the true categories), with three criteria and three categories and with six criteria and five. The Chebyshev centre is defined only in the additive scenario. Half-widths of the 95\% confidence intervals are at most \shapeMaxHalfWidth. Best mean per column in bold. Abbreviations: LS-UTADIS, least-squares UTADIS; UTADIS-MM, UTADIS with a max-margin second stage; centred LS, centred least squares; SVM (RBF), support vector machine with a Gaussian kernel.}
\label{tab:shape}
\footnotesize
\setlength{\tabcolsep}{2.4pt}
\resizebox{\textwidth}{!}{%
\begin{tabular}{@{}lcccccccccccc@{}}
\toprule
 & \multicolumn{6}{c}{3 criteria, 3 categories} & \multicolumn{6}{c}{6 criteria, 5 categories} \\
\cmidrule(lr){2-7}\cmidrule(l){8-13}
 & \multicolumn{3}{c}{Additive} & \multicolumn{3}{c}{Mislabelling} & \multicolumn{3}{c}{Additive} & \multicolumn{3}{c}{Mislabelling} \\
\cmidrule(lr){2-4}\cmidrule(lr){5-7}\cmidrule(lr){8-10}\cmidrule(lr){11-13}
Method & 20 & 40 & 80 & 20 & 40 & 80 & 20 & 40 & 80 & 20 & 40 & 80 \\
\midrule
LS-UTADIS & 0.836 & 0.887 & 0.937 & 0.698 & 0.767 & 0.810 & 0.642 & 0.767 & 0.878 & 0.501 & 0.587 & 0.684 \\
Huber shortfalls & 0.833 & 0.886 & 0.933 & 0.765 & \textbf{0.834} & \textbf{0.879} & 0.639 & 0.757 & 0.871 & 0.528 & 0.678 & \textbf{0.808} \\
LS-UTADIS, all thresholds & 0.836 & 0.887 & 0.937 & 0.695 & 0.752 & 0.788 & 0.643 & 0.767 & 0.879 & 0.493 & 0.568 & 0.642 \\
Relative margin & 0.820 & 0.878 & 0.933 & 0.714 & 0.772 & 0.811 & 0.620 & 0.741 & 0.853 & 0.482 & 0.569 & 0.659 \\
Centred LS & 0.671 & 0.674 & 0.667 & 0.627 & 0.645 & 0.658 & 0.497 & 0.557 & 0.585 & 0.429 & 0.445 & 0.454 \\
UTADIS & 0.820 & 0.877 & 0.928 & \textbf{0.769} & 0.832 & 0.878 & 0.614 & 0.738 & 0.854 & 0.548 & 0.678 & 0.802 \\
UTADIS with ridge & 0.827 & 0.879 & 0.930 & 0.766 & 0.831 & 0.877 & 0.633 & 0.748 & 0.861 & 0.545 & 0.681 & 0.802 \\
UTADIS-MM & 0.821 & 0.883 & 0.925 & 0.711 & 0.772 & 0.817 & 0.638 & 0.766 & 0.872 & 0.502 & 0.596 & 0.699 \\
Regularised UTADIS & 0.812 & 0.873 & 0.929 & 0.753 & 0.826 & 0.866 & 0.622 & 0.752 & 0.863 & \textbf{0.549} & \textbf{0.695} & \textbf{0.808} \\
UTADIS, all thresholds & 0.813 & 0.878 & 0.926 & 0.746 & 0.809 & 0.850 & 0.616 & 0.745 & 0.856 & 0.531 & 0.620 & 0.728 \\
Chebyshev centre & 0.834 & \textbf{0.895} & 0.935 & -- & -- & -- & 0.667 & 0.790 & 0.882 & -- & -- & -- \\
Proportional odds, additive value & \textbf{0.837} & \textbf{0.895} & \textbf{0.941} & 0.746 & 0.787 & 0.844 & \textbf{0.676} & \textbf{0.792} & \textbf{0.889} & 0.544 & 0.648 & 0.753 \\
Proportional odds, linear index & 0.824 & 0.857 & 0.869 & 0.728 & 0.779 & 0.826 & 0.610 & 0.687 & 0.720 & 0.528 & 0.629 & 0.694 \\
SVM (RBF) & 0.739 & 0.833 & 0.892 & 0.663 & 0.765 & 0.843 & 0.408 & 0.529 & 0.634 & 0.367 & 0.448 & 0.530 \\
Random forest & 0.703 & 0.787 & 0.847 & 0.644 & 0.729 & 0.803 & 0.410 & 0.481 & 0.552 & 0.381 & 0.450 & 0.518 \\
\bottomrule
\end{tabular}}
\end{table}

Table~\ref{tab:shape} considers generation additively and with mislabeling, with 3 criteria, 3 categories, and 6 criteria, 5 categories. For \shapeReps{} populations there are thus \shapeCells{} sets of cells. With additive labels, LS-UTADIS exceeds UTADIS significantly for all \shapeCells{} sets, by \vshapeAddGainLsuUtaMin{} to \vshapeAddGainLsuUtaMax{} points. With mislabelled observations, it is significantly exceeded in all \shapeCells{} cases, by up to \magn{\vshapeFlipsGainLsuUtaMin} points, this being for 6 criteria and 5 categories. Results are qualitatively similar for Huber shortfalls; with true labels they range from \magn{\vshapeAddGainHubLsuMax} to \magn{\vshapeAddGainHubLsuMin} points less than LS-UTADIS (significant for \vshapeAddGainHubLsuLosses{} sets of cells), \vshapeAddGainHubUtaMin{} to \vshapeAddGainHubUtaMax{} points more than UTADIS; with mislabeling \vshapeFlipsGainHubLsuMin{} to \vshapeFlipsGainHubLsuMax{} points greater than LS-UTADIS, \vshapeFlipsGainHubUtaMin{} to \vshapeFlipsGainHubUtaMax{} points compared with UTADIS, less for \vshapeFlipsGainHubUtaLosses{} set of cells. All-threshold versions show little difference with true labels, up to \shapeAddAtAbsMax{} points either way, but decreased accuracy with mislabeling, significantly so for \vshapeFlipsGainLsuAtLsuLosses{} sets of cells if squared and \vshapeFlipsGainUtaAtUtaLosses{} if not. LS-UTADIS exceeds UTADIS-MM significantly for \vshapeAddGainLsuUtaMMWins{} sets of cells and is below it for none, with true labels; for 6 criteria and 5 categories and $m=\shapeMSmall$, the amount is \vshapeAddGainLsuUtaMMSixSmall{} points (\vshapeAddGainLsuUtaMMSixSmallSE), and max-margin is exceeded by other methods based on values. The 6 choices of consistent function run in every setting (UTADIS, UTADIS-MM, UTADIS with ridge, Huber shortfalls, LS-UTADIS and the Chebyshev centre) were computed in each case. On QS and the simulated additive scenario, UTADIS and UTADIS-MM give rather lower average accuracy with $m=\shapeMSmall$; also UTADIS, and UTADIS with ridge, with 6 attributes and 5 categories. For this last, the difference between UTADIS with ridge and UTADIS-MM is only \vshapeSixGainMMRidgeSmall{} (\vshapeSixGainMMRidgeSmallSE) points, not significant. There seems to be more scope for variation between compatible models, with 6 criteria and 5 categories; for \vshapeSixAddGainLsuChebLosses{} of the 3 sizes the Chebyshev centre is better than LS-UTADIS significantly, and for \vshapeThreeAddGainLsuChebLosses{} with 3 criteria and 3 categories, differences up to \magn{\vshapeAddGainLsuChebMin} points. Proportional odds additively evaluated is never significantly less accurate; it is better significantly for \vshapeSixAddGainLsuOlogALosses{} and \vshapeThreeAddGainLsuOlogALosses{} of the sizes with true labels, and for \vshapeSixFlipsGainLsuOlogALosses{} and \vshapeThreeFlipsGainLsuOlogALosses{} with false labels, differences up to \magn{\vshapeAddGainLsuOlogAMin} and \magn{\vshapeFlipsGainLsuOlogAMin} points respectively.

\subsection{Sensitivity to examples}
\label{sec:stability}

An advantageous property of finding a single solution is lack of sensitivity both to which of possibly many is found, and to leaving out, or including, individual observations. A number of stability criteria have been considered by \citet{doumpos2004experimental} for typical ranges of parameters. We wish to know the effect of omitting 1 observation from determining the number of test observations misclassified. A number of other measures have been suggested \citep[for example][]{kadzinski2017expressiveness,vetschera2010robustness} concerning properties of all possible solutions; we are concerned only with the solution actually found and hence, presumably, applied. We have varied this aspect only, without other modifications such as recalibration or cross-validation. Each time 1 of 10 randomly selected observations was omitted and the rest fitted, using the parameters already found, the proportion of test observations assigned to a different category was noted (Table~\ref{tab:stability}). For $m=\stabMsmall$ these range over \stabQsLsuSmall, \stabQsUtaRidgeSmall, \stabQsUtaMMSmall{} and \stabQsUtaSmall{} per cent on QS for LS-UTADIS, UTADIS with ridge, UTADIS-MM and LP respectively, and over \stabAddLsuSmall, \stabAddUtaRidgeSmall, \stabAddUtaMMSmall{} and \stabAddUtaSmall{} per cent additively. The effects of the various amounts of ridge term are apparent at low $m$; also the square tends to lead to less fluctuation than the regularised absolute for larger sets, for example, \stabQsLsuLarge{} versus \stabQsUtaRidgeLarge{} per cent ($m=\stabMlarge$, QS). Differences of LS-UTADIS from UTADIS with ridge range over \stabGainLsuUtaRidgeMax{} to \stabGainLsuUtaRidgeMin{} points, all \stabCellsAll{} data sets and values of $m$, significantly negative on \stabGainLsuUtaRidgeSig{} occasions, \stabGainLsuUtaRidgeAddSmall{} points (\stabGainLsuUtaRidgeAddSmallSE) and \stabGainLsuUtaRidgeQsSmall{} points (\stabGainLsuUtaRidgeQsSmallSE) for $m=\stabMsmall$ additively and on QS. They range from \stabGainLsuUtaMMMax{} to \stabGainLsuUtaMMMin{} points and from \stabGainLsuUtaMax{} to \stabGainLsuUtaMin{} points compared with UTADIS-MM and LP respectively, significantly negative all \stabGainLsuUtaSig{} times.

Similar stability is shown by proportional odds with the same numbers of sub-intervals, \stabQsOlogASmall{} per cent on QS, \stabAddOlogASmall{} per cent additively ($m=\stabMsmall$). Differences of these proportions for LS-UTADIS range from \stabGainLsuOlogAMin{} to \stabGainLsuOlogAMax{} points, significantly so \stabGainLsuOlogASig{} times. Only solutions actually returned have been compared, so no exploration has been made of other possible solutions of the LP.

\begin{table}[t]
\centering
\caption{Sensitivity to the reference set: percentage of test alternatives whose assignment changes when one reference alternative is removed (mean over 100 reference sets and ten removals each, with 95\% half-widths). Fixed hyperparameters: margin $d=0.25$, no recalibration, one sub-interval per criterion on QS and three in the additive scenario; proportional odds on the additive value has no margin and is fitted with the same number of sub-intervals. Smallest value per column in bold. Abbreviations: LS-UTADIS, least-squares UTADIS; UTADIS-MM, UTADIS with a max-margin second stage.}
\label{tab:stability}
\small
\resizebox{\textwidth}{!}{%
\begin{tabular}{@{}lcccccc@{}}
\toprule
 & \multicolumn{3}{c}{QS rankings, by $m$} & \multicolumn{3}{c}{Additive scenario, by $m$} \\
\cmidrule(lr){2-4}\cmidrule(l){5-7}
Method & 20 & 40 & 80 & 20 & 40 & 80 \\
\midrule
LS-UTADIS & \textbf{2.5} {\scriptsize$\pm$0.3} & \textbf{0.8} {\scriptsize$\pm$0.1} & \textbf{0.2} {\scriptsize$\pm$0.0} & \textbf{3.2} {\scriptsize$\pm$0.4} & 1.9 {\scriptsize$\pm$0.2} & \textbf{0.5} {\scriptsize$\pm$0.1} \\
UTADIS & 5.6 {\scriptsize$\pm$1.0} & 3.1 {\scriptsize$\pm$0.5} & 0.6 {\scriptsize$\pm$0.1} & 13.5 {\scriptsize$\pm$1.5} & 3.2 {\scriptsize$\pm$0.5} & 0.9 {\scriptsize$\pm$0.1} \\
UTADIS with ridge & 2.7 {\scriptsize$\pm$0.3} & 1.1 {\scriptsize$\pm$0.1} & 0.5 {\scriptsize$\pm$0.1} & \textbf{3.2} {\scriptsize$\pm$0.4} & 2.2 {\scriptsize$\pm$0.3} & 0.9 {\scriptsize$\pm$0.1} \\
UTADIS-MM & 3.7 {\scriptsize$\pm$0.4} & 3.5 {\scriptsize$\pm$0.6} & 1.3 {\scriptsize$\pm$0.3} & 6.0 {\scriptsize$\pm$0.5} & 2.1 {\scriptsize$\pm$0.3} & 0.6 {\scriptsize$\pm$0.1} \\
Proportional odds, additive value & \textbf{2.5} {\scriptsize$\pm$0.2} & \textbf{0.8} {\scriptsize$\pm$0.1} & \textbf{0.2} {\scriptsize$\pm$0.0} & 3.8 {\scriptsize$\pm$0.3} & \textbf{1.5} {\scriptsize$\pm$0.1} & \textbf{0.5} {\scriptsize$\pm$0.1} \\
\bottomrule
\end{tabular}}
\end{table}

\subsection{Extent of tuning, iterations, and time taken}
\label{sec:time}

Table~\ref{tab:tuning} gives some indication of extent of tuning required. For \fitsCount{} solutions, final, requiring an accuracy measure, mean number of iterations in Algorithm~\ref{alg:rls} is \iterMean, ranging up to \iterMax. All but a few converge because the solution based on the then current subdivision coincides exactly with that of previous iteration, or satisfy the alternative criterion, the remainder failing to show any significant reduction in value of the function to be minimised; none continue to the maximum allowed 500.

\begin{table}[t]
\centering
\caption{Least-squares UTADIS with unknown limits, by data set and for its smallest and largest $m$: mean selected margin $d$ (as a fraction of the mean half-width of a category) and number of sub-intervals $\gamma$; share of reference sets in which the limits are recalibrated; mean percentage of reference alternatives that violate both ends of their zone at the solution (crossed ends); mean number of inner folds; mean number of outer iterations of Algorithm~\ref{alg:rls} in the final fit.}
\label{tab:tuning}
\small
\resizebox{\textwidth}{!}{%
\begin{tabular}{@{}lrcccccc@{}}
\toprule
Data & $m$ & $d$ & $\gamma$ & Recalibrated & Both ends (\%) & Folds & Iterations \\
\midrule
QS rankings & 20 & 0.31 & 1.58 & 0.24 & 1.2 & 3.7 & 7.4 \\
QS rankings & 320 & 0.12 & 1.11 & 0.40 & 0.0 & 5.0 & 5.8 \\
ERA & 50 & 0.38 & 1.78 & 0.56 & 8.6 & 3.9 & 3.6 \\
ERA & 200 & 0.40 & 1.71 & 0.87 & 6.5 & 3.6 & 3.6 \\
ESL & 50 & 0.34 & 1.81 & 0.33 & 6.6 & 4.5 & 5.0 \\
ESL & 200 & 0.51 & 1.86 & 0.44 & 6.3 & 4.0 & 4.4 \\
LEV & 50 & 0.44 & 1.70 & 0.58 & 2.1 & 4.0 & 3.6 \\
LEV & 200 & 0.51 & 1.56 & 0.81 & 0.8 & 4.4 & 3.6 \\
SWD & 50 & 0.43 & 1.76 & 0.68 & 2.1 & 4.0 & 4.2 \\
SWD & 200 & 0.50 & 1.89 & 0.78 & 1.7 & 4.8 & 3.7 \\
Additive & 20 & 0.43 & 1.90 & 0.22 & 8.1 & 3.2 & 6.5 \\
Additive & 160 & 0.19 & 2.93 & 0.27 & 0.3 & 5.0 & 7.5 \\
Smooth marginals & 20 & 0.43 & 1.68 & 0.31 & 7.9 & 3.2 & 6.5 \\
Smooth marginals & 160 & 0.22 & 2.52 & 0.27 & 0.8 & 5.0 & 6.4 \\
Interacting criteria & 20 & 0.40 & 1.86 & 0.30 & 7.2 & 3.2 & 6.6 \\
Interacting criteria & 160 & 0.35 & 2.72 & 0.25 & 4.6 & 5.0 & 5.6 \\
Non-monotone criterion & 20 & 0.42 & 2.24 & 0.39 & 9.8 & 3.3 & 5.4 \\
Non-monotone criterion & 160 & 0.43 & 2.93 & 0.61 & 7.1 & 5.0 & 4.9 \\
Noisy assignments & 20 & 0.36 & 1.84 & 0.36 & 7.2 & 3.0 & 6.1 \\
Noisy assignments & 160 & 0.45 & 2.67 & 0.44 & 7.2 & 5.0 & 4.6 \\
Distant mislabelling & 20 & 0.42 & 1.93 & 0.46 & 14.8 & 3.3 & 5.2 \\
Distant mislabelling & 160 & 0.69 & 2.14 & 0.81 & 28.4 & 5.0 & 4.0 \\
\bottomrule
\end{tabular}}
\end{table}

A reasonable quantity to monitor convergence is the largest component of $\tht-P_\Theta(\tht-\nabla\Psi(\tht))$, zero if $\tht$ is exactly a solution. We have aimed for $10^{-10}$ and disallowed any solution greater than $10^{-7}$. None of \kktFits{} solutions, final, with all amounts of tuning, of variants of LS-UTADIS, relative margin, and centring, with and without specified limits, is greater than this, \kktAboveTarget{} failing to reach $10^{-10}$ and maximum \kktMaxAll. Re-solving \auditConicSolved{} of them with Clarabel to $10^{-12}$ leads to differences of parameters no more than \auditDevMax{} and no change of allocated categories. Over solutions during cross-validation, the largest for a procedure is greater than $10^{-10}$ for \kktInnerAboveTarget{} procedures, maximum \kktMaxInner, although we have not kept sufficient details to examine these individually. Maximum for profiles and for stability tests is \kktMaxProfile{} and \kktMaxStability. We did not note this quantity for final solutions of Section~\ref{sec:fixed}, although they are all tested similarly, any failures to be assumed incorrectly classified: none occur. For UTADIS with ridge, all \conicRidgeFits{} reach $10^{-12}$, none ceasing at $10^{-8}$. The least that any observation lies inside its current interval, over all these solutions, is \mindistMin, although clearly no inference can be made regarding the true solution and Assumption~\ref{ass:nondeg}. Solutions arising in the course of cross-validation have, as before, been taken as incorrect classifications if no solution is found or rejected by the above criterion, which does not occur for any problem involving shortfalls (but c.f.\ the 2 centres above). For \conicSolvedShare{} per cent of those for which time is noted, Clarabel finds a solution to $10^{-12}$, agreeing with ours to \conicGapMax{} (value of function) and \conicDevMax{} (parameters).

Median times taken for 1 application of various procedures are given in Table~\ref{tab:timing}. Thus at most \timeLsuMax{} milliseconds, as a median, to fit a final problem for LS-UTADIS, and \timeUtaMax{} milliseconds for the LP. There would therefore be little reason to prefer one to the other, both being extremely quick. Proportional odds takes from \timeOlogAMin{} to \timeOlogAMax{} milliseconds (median, for the stability experiments, others being run at the same time), using a general purpose minimiser and numerical first derivatives, trying 3 initial values. This is probably slow, as much more information is available to Algorithm~\ref{alg:rls}; we do not wish to imply any comparison of speed of solution. However as it is possible there may be several local minima, we have restarted all \ologAuditFits{} final solutions from 12 initial points. Only \ologAuditLower{} lead to any decrease exceeding $10^{-6}$ proportionately (maximum \ologAuditGapMax), corresponding to no more than \ologAuditChangedMax{} per cent difference of observations allocated, for any solution. The largest component of gradient is \ologAuditGradMax, so there is no strong presumption these are the global minima. All solutions for regularised UTADIS converge, \regTrivial{} having all increments zero.

\begin{table}[t]
\centering
\caption{Median time of a single fit in milliseconds on the QS data (20 reference sets per cell), unknown limits, margin $d=0.25$, for one and three sub-intervals per criterion. The last row solves the problem of the first with the conic solver Clarabel at tolerance $10^{-12}$. Abbreviations: LS-UTADIS, least-squares UTADIS; centred LS, centred least squares; HiGHS, the dual simplex method of the HiGHS solver.}
\label{tab:timing}
\footnotesize
\setlength{\tabcolsep}{3.5pt}
\resizebox{\textwidth}{!}{%
\begin{tabular}{@{}lrrrrrrrrrrrr@{}}
\toprule
 & \multicolumn{6}{c}{$\gamma=1$, by $m$} & \multicolumn{6}{c}{$\gamma=3$, by $m$} \\
\cmidrule(lr){2-7}\cmidrule(l){8-13}
Method & 20 & 40 & 80 & 160 & 320 & 640 & 20 & 40 & 80 & 160 & 320 & 640 \\
\midrule
LS-UTADIS (Algorithm~\ref{alg:rls}) & 3.8 & 2.5 & 2.0 & 1.8 & 1.8 & 2.2 & 5.3 & 6.8 & 5.0 & 3.4 & 2.8 & 2.6 \\
Centred LS (one step) & 0.3 & 0.3 & 0.3 & 0.4 & 0.4 & 0.5 & 0.6 & 0.5 & 0.4 & 0.4 & 0.5 & 0.6 \\
UTADIS, linear programme (HiGHS) & 1.3 & 1.5 & 1.9 & 3.3 & 8.8 & 27.8 & 1.4 & 1.7 & 2.5 & 4.1 & 10.5 & 33.0 \\
UTADIS with ridge, conic solver & 4.8 & 5.1 & 5.6 & 7.2 & 10.1 & 16.0 & 5.2 & 5.4 & 6.3 & 8.5 & 13.2 & 22.9 \\
LS-UTADIS, conic solver & 5.0 & 5.2 & 6.2 & 8.4 & 10.1 & 16.9 & 5.0 & 5.7 & 7.0 & 9.8 & 16.1 & 29.3 \\
\bottomrule
\end{tabular}}
\end{table}

\subsection{Cautions regarding tuning}
\label{sec:fixed}

Several features of these and other experiments with tuning merit comment. Only 1 pair of values of $\rho$ and $\eta$ has been used, in order to attribute any improvement to these and to squaring; and since 24, 72, and \cenCand{} sets of parameters would be tuned, respectively, for most shortfall criteria other than Huber, for Huber, and for both centres and proportional odds, some improvements may be attributable to that. Most likely to be affected are the likelihood and the Chebyshev centre.

It is therefore useful to examine the case where no tuning is done, but with true generating rule admissible. We use QS, with 1 sub-interval for each criterion, and additive marginals with 3 sub-intervals, giving all methods this number of sub-intervals. No recalibration is allowed and margin $d$ is set at 0.25, as in experiments on stability and times. Hence there are a number of differences from experiments with tuning. In particular all observations fall strictly inside their true intervals, so both centres exist. Each sample of relevant size is used as a reference set in turn. For each, we find: UTADIS and LS-UTADIS, each for \fixMarginCount{} values of $d$, only results for $d=0.25$ reported here; each, with \fixRhoCount{} values of $\rho=\eta$ between $10^{-6}$ and $10^{-2}$, again $d=0.25$; UTADIS-MM; the Chebyshev centre (of largest circle contained in the feasible set) by LP; the analytic centre \citep{bous2010acuta}; proportional odds, with 3 values of ridge term. A solution was obtained in all \fixFits{} cases; results are shown in Table~\ref{tab:fixed}.

\begin{table}[t]
\centering
\caption{Fixed hyperparameters when the rule that generated the labels belongs to the model class: test accuracy (means over 100 reference sets of size $m$) on the QS data with one sub-interval per criterion and in the additive scenario with three. No method is tuned: margin $d=0.25$ for UTADIS, UTADIS with ridge and LS-UTADIS (UTADIS-MM has no fixed margin), $\rho=\eta=10^{-4}$ for the ridge terms, ridge $10^{-3}$ for the likelihood; the two centres are those of the polyhedron of parameters compatible with the reference set. Best mean per column in bold (all values that print equal to it). Abbreviations: LS-UTADIS, least-squares UTADIS; UTADIS-MM, UTADIS with a max-margin second stage.}
\label{tab:fixed}
\small
\setlength{\tabcolsep}{4pt}
\resizebox{\textwidth}{!}{%
\begin{tabular}{@{}lcccccc@{}}
\toprule
 & \multicolumn{3}{c}{QS, by $m$} & \multicolumn{3}{c}{Additive scenario, by $m$} \\
\cmidrule(lr){2-4}\cmidrule(l){5-7}
Method & 20 & 40 & 80 & 20 & 40 & 80 \\
\midrule
UTADIS, vertex solution & 0.813 & 0.910 & 0.939 & 0.718 & 0.867 & 0.910 \\
UTADIS with ridge & 0.845 & 0.911 & 0.938 & 0.770 & 0.871 & 0.910 \\
UTADIS-MM & 0.812 & 0.905 & 0.948 & 0.753 & 0.870 & 0.914 \\
Chebyshev centre & 0.849 & 0.918 & \textbf{0.953} & 0.775 & 0.878 & 0.918 \\
Analytic centre & 0.827 & 0.911 & 0.949 & 0.760 & 0.867 & 0.919 \\
LS-UTADIS & \textbf{0.853} & 0.920 & 0.951 & 0.770 & 0.878 & \textbf{0.926} \\
Proportional odds, additive value & 0.843 & \textbf{0.921} & \textbf{0.953} & \textbf{0.785} & \textbf{0.884} & \textbf{0.926} \\
\bottomrule
\end{tabular}}
\end{table}

It can be seen that much of the improvement at low $m$ is due to the ridge terms: \fixGainUtaRidgeUtaQSSmall{} points (\fixGainUtaRidgeUtaQSSmallSE) on QS and \fixGainUtaRidgeUtaAddSmall{} points (\fixGainUtaRidgeUtaAddSmallSE) additively are gained by their inclusion with an absolute penalty, relative to no ridge term, that is the vertex. Change to a square adds only \fixGainLsuUtaRidgeQSSmall{} points (\fixGainLsuUtaRidgeQSSmallSE) on QS, \fixGainLsuUtaRidgeAddSmall{} points (\fixGainLsuUtaRidgeAddSmallSE) additively. For all \fixCells{} values of $m$ and type of data, LS-UTADIS exceeds UTADIS with ridge, significantly for \fixGainLsuUtaRidgeWins, by at most \fixGainLsuUtaRidgeMax{} points. $m=\fixGainLsuUtaRidgeMaxM$ gives the greatest difference.

Little change is apparent with amount of ridge term, as expected from the arguments in Section~\ref{sec:model}: the maximum variation of mean accuracies, over \fixRhoCount{} values of $\rho=\eta$, of UTADIS with ridge is \fixRhoUtaRidgeRange{} points. We did not determine, for any reference set, the value of the parameters for the ridge terms below which they would merely have an effect on choice between minimisers (Proposition~\ref{prop:vanishing}(i)); the small differences suggest most would be below this value. On the other hand variation is fairly marked for LS-UTADIS, up to \fixRhoLsuRange{} points, so that differences of these methods could range over \fixSquareByRhoQSSmallMin{} to \fixSquareByRhoQSSmallMax{} points (QS, $m=\fixMsmall$), and \fixSquareByRhoMin{} to \fixSquareByRhoMax{} points generally. Thus the amount due to squaring is similar in magnitude to that arising over this range of ridge terms, so we would hesitate to attribute it to this feature. To the extent that all observations are well inside their intervals for some parameters, there would of course be no difference as $\rho\to0$, both choosing this solution (Section~\ref{sec:model}); to the extent otherwise, differences should be due to differences in the functions minimised. We do not have an easy means of ascertaining for how many reference sets the former applies, with $d=0.25$. Furthermore, as discussed in Section~\ref{sec:model}, other solutions to the problem without penalties increase the value minimised, to first order in the case of the absolute penalty, but only second order if squared, so there will be interaction with the amounts of the penalties. For this reason, we would consider the square together with the penalties as one entity.

It can be seen from Table~\ref{tab:fixed} that the Chebyshev centre performs similarly. Differences of paired observations range from \fixGainLsuChebMin{} to \fixGainLsuChebMax, and are significant in favour of LS-UTADIS in one cell only (additive, $m=\fixChebBehindM$) and in favour of the Chebyshev centre in none. For $m=\fixMsmall$ the analytic centre is \fixGainLsuAcenQSSmall{} points (\fixGainLsuAcenQSSmallSE) lower than LS-UTADIS on QS, the max-margin second stage \fixGainLsuUtaMMQSSmall{} points (\fixGainLsuUtaMMQSSmallSE) lower. So, if the true rule is within the class of models considered, to use squared rather than absolute shortfalls gains, relative to solving the linear programme, about what choosing some centre of the set of compatible models would gain.

Likelihood and LS-UTADIS differ by at most \fixLossLsuOlogAMax{} points without tuning; differences of paired observations range from \fixGainLsuOlogAMin{} to \fixGainLsuOlogAMax, equal to \fixGainLsuOlogAQSSmall{} points (\fixGainLsuOlogAQSSmallSE) for $m=\fixMsmall$ on QS and to \fixGainLsuOlogAAddSmall{} points (\fixGainLsuOlogAAddSmallSE) additively. They are significant in \fixGainLsuOlogALosses{} cells, all in favour of the likelihood, none in favour of LS-UTADIS. It is less clear a priori whether the superiority of the tuned likelihood will persist without tuning, and it will vary with data set and $m$. It does not persist on QS: the tuned likelihood is significantly superior for $m=\qsOlogALeadSizes$ only, while for $m=\fixMsmall$ the difference is reversed and for $m=\qsMsecond$ nearly vanishes. All methods would be expected to be somewhat advantaged without tuning, since all would use the number of sub-intervals of the true rule, LS-UTADIS would use nearly the most favourable margin and the likelihood one of its ridges. For $m=\fixMlarge$ on QS the likelihood is superior without tuning, LS-UTADIS being lower by \magn{\fixGainLsuOlogAQSLarge} points (\fixGainLsuOlogAQSLargeSE), a small but significant difference. Additively, the likelihood is superior without tuning for $m=\fixMsmall$ and \fixMsecond, but not \fixMlarge; it is more superior without than with tuning for $m=\fixMsmall$, less so for $m=\fixMsecond$. These are differences of means, not of paired observations, since different reference sets are involved.

From the margin profiles in Section~\ref{sec:margin}, using the most favourable margins, there is little difference between the two procedures on QS, LS-UTADIS being superior only for the smallest $m$. But additively, the likelihood is superior at all margins shown, both for $m=\simMsmall$ and \plAddLargeM, including $m=\plAddLargeM$, where there is no apparent difference between them without tuning. Thus selection over the ridge is important for the likelihood, which varies by as much as \fixRidgeOlogARange{} points according to the value of the ridge, again on reference sets all separable.

Table~\ref{tab:fixed} reports results for a single margin, $d=0.25$. The mean over all observations at the other values of $d$ in the grid used never exceeds that for $d=0.25$ by more than \fixMarginLsuGainMax{} points, and $d=0.25$ is the best value for \fixMarginLsuBestAtFixed{} out of the \fixCells{} cells. Using the best $d$ in each cell, LS-UTADIS differs from the Chebyshev centre between \fixMarginLsuBestChebMin{} and \fixMarginLsuBestChebMax{} points, and from the likelihood between \fixMarginLsuBestOlogAMin{} and \fixMarginLsuBestOlogAMax{} points. As the vertex solution can improve by as much as \fixMarginUtaGainMax{} points by changing $d$, some of its inferiority relative to others in Table~\ref{tab:fixed} will be attributable to $d$. Only 2 sources of observations have been used and there is no mislabelling. This is no substitute for the experiments under tuning, and of course various centres cannot be calculated if there are no compatible models.

\subsection{Summary}
\label{sec:summary}

\begin{figure}[t]
\centering
\includegraphics{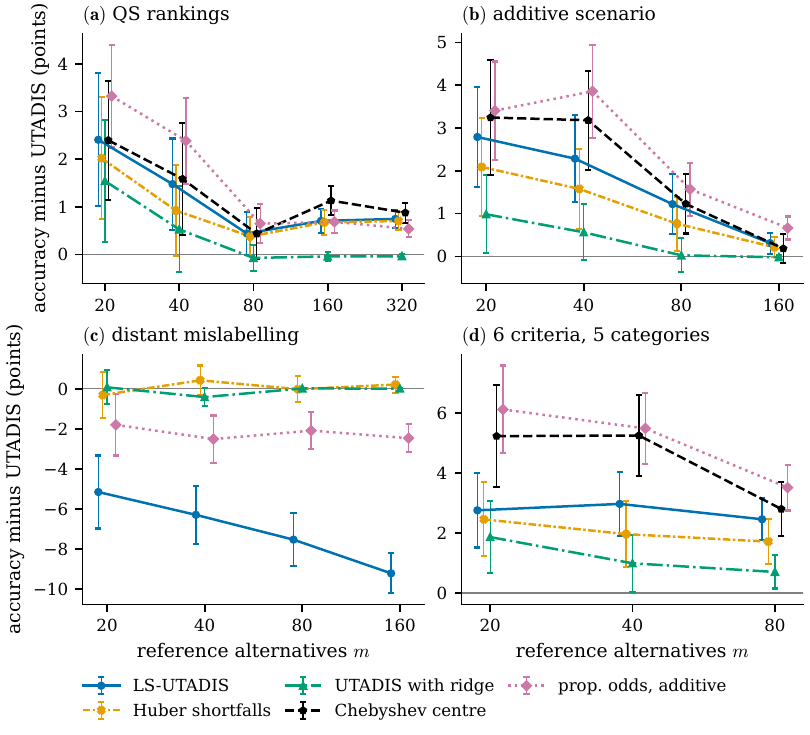}
\caption{Difference in accuracy from UTADIS, on identical reference sets, in points: average, over 100 sets or populations, of differences for each set, with 95 per cent confidence intervals, of LS-UTADIS, Huber shortfalls, UTADIS plus ridge terms, Chebyshev centre (only if the generating mechanism is within the model), proportional odds additively evaluated. (a) QS. (b) Additive rule. (c) Mislabeling, accuracy relative to true categories. (d) 6 criteria, 5 categories, additive rule. Confidence intervals offset horizontally to avoid overlap.}
\label{fig:gains}
\end{figure}

Figure~\ref{fig:gains} summarises many results of this section.

Apart from associated changes of convention, there is little difference in performance due to squaring the shortfalls. For reference sets of \qsMsmall{} or \qsMsecond{} (with 4 or 5 criteria and 4 categories), on QS and additively, LS-UTADIS exceeds the solution to the linear programme by \sumGainLsuUtaMin{} to \sumGainLsuUtaMax{} points, tending to be less for larger $m$. A number of experiments serve to illustrate causes of improvements, although not always agreeing for larger values of $m$. Some improvement may be achieved without squaring: LS-UTADIS is not significantly better than the Chebyshev centre of all compatible models, under tuning, for any $m$, although there are only \cenCand{} such candidates compared with 24 for LS-UTADIS; and they are close together for fixed values (Section~\ref{sec:fixed}). \citet{wojcik2023representative} compare various representatives of the set of compatible models and it seems plausible that one near the centre would perform well. They find the two centres considered here in the opposite order to us. Adding ridge terms to the un-squared shortfalls recovers much of the benefit of squaring, for fixed values at the smallest size and for QS under tuning, but not additively under tuning, LS-UTADIS exceeding them significantly for most fixed cells (Section~\ref{sec:fixed}) and \qsGainLsuUtaRidgeWins{} out of the 5 sizes on QS under tuning. It therefore seems that, at least for these values of $\rho$, and assuming improvements are due to choice between equally good models (unresolved in Section~\ref{sec:fixed}), squared terms effect a better choice than do ridge terms, for $m$ above its minimum. It is not possible to infer from this what happens as $\rho$ approaches zero (Section~\ref{sec:fixed}). The max-margin second stage is inferior in all cells of the fixed experiments, by \qsGainLsuUtaMMSmall{} points (QS) and \addGainLsuUtaMMSmall{} points (additive) for $m=\qsMsmall$ under tuning, and by no more than \mmFromSecondMax{} points from $m=\qsMsecond$ on (Section~\ref{sec:sim}). Regularised UTADIS, tuning its single constant, ranges from \regVsUtaMin{} points to \regVsUtaMax{} points relative to UTADIS, over all \regVsUtaCells{} cells of the experiments measuring accuracy, and is significantly inferior in \regVsUtaLosses{} cells, superior in \regVsUtaWins. With errors in observation squaring is of no benefit, and substantially inferior with distant mislabeling.

Neither loss function is optimal, under tuning. For fixed values they are fairly close (Section~\ref{sec:fixed}). Over the range of the margin profiles, again choosing most favourable values ex post, LS-UTADIS exceeds the proportional-odds likelihood on QS for $m=\qsMsmall$ and is inferior for all margins additively (Section~\ref{sec:margin}). Fitting this to the same additive generating value, it is inferior significantly only on QS, $m=\qsLsuOlogAWinSizes$, by \qsGainLsuOlogALarge{} points, and superior significantly over most of the simulation cells and for the published data; its mean is greater than for the linear programme for all cells except distant mislabelling. It varies by less than \simLossLsuOlogAMax{} points from LS-UTADIS for all \simScenariosNoFlips{} scenarios, without mislabelled observations, and up to \magn{\vshapeAddGainLsuOlogAMin} points if 6 criteria and 5 categories.

An advantage of least squares over the linear programme is that individual estimates vary less on deletion of one of the reference observations for fixed values, and overall accuracy is rather less variable over the values of $d$ tried. Deletion of one reference observation has rather little effect on the likelihood. The squared criterion also gives rise to a least-squares problem (Section~\ref{sec:solutions}), identifying conditions under which it reduces to least-squares fitting of the centres. Solution requires very little computation (Algorithm~\ref{alg:rls}). It will terminate in a finite number of iterations if Assumption~\ref{ass:nondeg} holds; this is unestablished, and otherwise a termination criterion is applied (Section~\ref{sec:time}). Of course all these comments are only relative to other procedures considered (Section~\ref{sec:design}). We have aimed primarily at various features of 2 objectives for a particular family of solutions, and comparisons with some alternatives; not at an extensive comparison of methods for deriving an ordering. Adding a centring constraint, and allowing the margin to vary relatively over categories, are not beneficial. For a linear rule the proportional-odds likelihood with linear indices would be expected to be still more accurate, as would an SVM if any index was non-monotonic and there were sufficient observations. Neither would give all marginal values and trade-offs on a comparable scale, unlike the proportional-odds likelihood with an additive generating value.

Huber loss on shortfall retains much both of the advantage of squaring, never more than \hubConsistentLossMax{} points less accurate than LS-UTADIS if true labels are generated from a model, and of resilience to mislabelling, never more than \hubMainFlipsLossMax{} points less accurate than UTADIS if labels are very different (\hubFlipsLossMax{} points with other numbers of attributes and categories). Using all cutoff points makes little difference if labels are generated from a model, but reduces accuracy substantially with very mislabelled data, as explained in Proposition~\ref{prop:huber}(iii). They perform better, with absolute values, on the original sets of judgements. Broadly similar differences between LS-UTADIS and UTADIS, and proportional-odds logistic regression to represent the additive function, are observed with 3 and 6 attributes, and 3 and 5 categories, on the 3 additional published sets, except for 1 significant inferiority of LS-UTADIS to UTADIS, on MPG. For 6 and 5 both Chebyshev centre and maximum likelihood are further superior.

\section{Conclusion}
\label{sec:conclusion}

These results suggest a different issue of prime importance in applying UTADIS, namely how to select from among compatible models, if it is reasonable to suppose that the generating rule lies within the class (Section~\ref{sec:summary}). That a centre of this set would go a long way to achieving the benefit of squaring is shown by the result for the Chebyshev centre. It is an open issue whether it would retain superiority over adding ridge terms, as the size of these approaches 0, since this would turn on which reference sets admit parameters keeping all observations inside their intervals, which was not recorded (Section~\ref{sec:fixed}).

It seems that, given a limited number of reference observations, plausibly generated according to some rule in the model class, that is, some additive function of piecewise-linear marginals, it would be better to choose towards the centre of all rules consistent with the data. In the main experiments, at the smallest size of reference set used, the two most natural alternatives to LS-UTADIS, namely stopping at a vertex, and continuing to the solution maximizing the margin, both gave worst accuracy of the 6 choices of consistent function of Section~\ref{sec:further}, regardless of whether the parameters were varied. With 6 attributes and 5 categories, max margin was not worst, but UTADIS with ridge gave lower accuracy, non-significantly. That all observations can be accommodated by such a rule, given only a few observations, is of course far from establishing that they were so generated. Obviously this argument does not apply if gross errors may have occurred in labelling. If these are a possibility then, subject to the qualifications of Section~\ref{sec:sim}, tuning the margin in UTADIS appears preferable. But they can only be suspected, not known to have occurred.

Thus of alternatives to least squares, Huber shortfalls retain much both of the advantage of squaring, never more than \hubConsistentLossMax{} point less accurate with appropriate labels, and of insensitivity to extreme observations, not more than \hubFlipsLossMax{} points less accurate (\hubMainFlipsLossMax{} in the main simulation) with badly mislabelled ones. Shortfalls measured at all limits tends to exaggerate the effect of very wrong labels, so seems unattractive in that situation; that it performs better for absolute error on available data sets is surprising and warrants investigation. Squared terms also proved advantageous with 2 of 3 other published data sets, 1 generated by a monotonic rule not representable by any value model fitted here with category widths of at least 0.01, and for other combinations of numbers of criteria and categories. We did not investigate whether this was due to the effect on choice of model, as compatible models were not identified; but proportional odds evaluated additively was never significantly less accurate, and Chebyshev centre better, with 6 criteria, 5 categories.

More investigation is warranted of the likelihood. Proportional odds evaluated additively was, with tuning, better than LS-UTADIS significantly for \ologMainWinCells{} out of \ologMainCells{} sets of cells, worse for \ologMainLossCells, over the principal range of experiments, and never significantly worse for others. However, its advantage on QS disappeared without tuning, at least at the size for which it was significant. This may well reflect an adverse effect of tuning. The program to evaluate it was written for ease of implementation; although it was run from several initial positions, optimality was not certified. Further investigation into its numerical properties and the nature of its set of optima, analogous to those established here for the squared criterion, seems worthwhile. Regret bounds have recently been established for various point likelihoods applicable to choices and rankings \citep{grillo2025online}.

Many phenomena observed might be predicted from the results of Section~\ref{sec:solutions}. A fairly common consequence of cross-validation would be a margin exceeding half 1 or more widths, causing all observations between the relevant pair of crossing points to be fitted as for centred least squares; such observations were a minority in all cases (Table~\ref{tab:tuning}). On the other hand, regression of the category centres, which theory identifies as the wrong target, failed in all cases.

There are various other qualifications on these experiments, in addition to the limitations of the data. Without tuning, only 1 margin was used for most comparisons, and no data containing labelling error except with Huber shortfalls. The value of the fixed ridge terms of the likelihood determines which minimum is returned, given separability. Also, its rescaled upper and lower bounds were outside the interval 0 to 1 in some instances (Section~\ref{sec:design}), so did not necessarily define admissible category limits. The non-simulated data consist of one ranking and a number of standard test problems. There are too few observations in any set of elicited preferences to permit repetition of samples of varying sizes and testing on remaining observations. Hence nothing can be concluded about behaviour on such data.

A number of other procedures have not been applied, for the reasons given in Section~\ref{sec:design}. We have applied regularised UTADIS as we interpret the published description, with 1 tuning constant applicable equally to all categories. There may be a more favourable rule. None of these procedures yields any indication of reliability of classification of an individual object. Selection of further observations to elicit and aggregation over several decision makers are also outside its scope.

\section*{Data availability statement}
Data for the 4 main problems may be downloaded from OpenML, data set numbers 1030, 1027, 1029, 1028. Those for the other 3 problems are available from the UCI machine-learning repository \citep{bohanec1988car,feldmesser1987cpu,quinlan1993mpg}. Lens scores underlying the 2026 QS rankings of universities are published by QS Quacquarelli Symonds. All programs used to obtain results reported here, relevant output, description of hardware and software, and output and comparison of a rerun of all calculations, may be obtained from the corresponding author.

\section*{Author contributions}
Yuqun Zhou: Conceptualization, Methodology, Software, Formal analysis, Investigation, Data curation, Visualization, Writing -- original draft, Writing -- review and editing. Yuxi Piao: Investigation, Writing -- review and editing. Fangyuan Shi: Investigation, Writing -- review and editing. Jiufan Wang: Investigation, Writing -- review and editing.

\section*{Conflict of interest}
None declared.

\section*{Use of generative artificial intelligence}
Various generative AI programs (Claude, Anthropic; Codex, OpenAI) were used to write and run programs, to check results, to write and edit parts of the text, and to comment on drafts. Current versions run between 12 and 30 November 2026 are Claude Fable 5.2 and Opus 5.5, run under Claude Code, and Codex running GPT-6-astra. All content is the responsibility of the authors.

\appendix
\section{Some proofs}
\label{app:proofs}

\begin{proof}[Lemma~\ref{lem:identity}]
On expanding the 2 squares, the remaining cross-product terms are equal and opposite.
\end{proof}

\begin{proof}[Proposition~\ref{prop:convex}]
Each term of \eqref{eq:loss} is a convex, non-decreasing and continuously differentiable function applied to a linear function of $(\vv,\cw)$, and so is convex and continuously differentiable. There must therefore be at least one minimiser over compact convex $\Theta$. The additional ridge terms increase strong convexity by $2\min(\rho,\eta)$, ensuring a unique minimiser if both are strictly positive, similarly for $(\mathrm K)$ if $\rho>0$. If there are no ridge terms and only centring, the function to be minimised in $(\mathrm K^{\mathrm c})$ is $\|\Ph \vv-\cc\|^2$, strictly convex provided $\Ph\T\Ph$ is non-singular.
\end{proof}

\begin{proof}[Proposition~\ref{prop:vanishing}]
Omitting the last 2 terms, denote the function to be minimised by $\Psi_0$, and set $R(\tht)=\|\vv\|^2+\|\cw-q^{-1}\one\|^2$. Since $\one\T\vv=1$, $R$ measures distance from $(N^{-1}\one,q^{-1}\one)$ apart from an additive constant. If $\tht_\rho$ minimises $\Psi_0(\tht)+\rho R(\tht)$ over $\Theta$ and $\bar\tht$ is any point of $\mathcal M_0$, then clearly $R(\tht_\rho)\le R(\bar\tht)$ and $\Psi_0(\tht_\rho)\le\Psi_0(\bar\tht)+\rho\,R(\bar\tht)$. There will be a limit point $\tht^\ast$ of $\tht_\rho$ as $\rho$ approaches 0, and by continuity $\Psi_0(\tht^\ast)\le\Psi_0(\bar\tht)$, so $\tht^\ast\in\mathcal M_0$, and $R(\tht^\ast)\le R(\bar\tht)$ whatever point $\bar\tht$ of $\mathcal M_0$. Since $R$ is strictly convex, there is just 1 point of $\mathcal M_0$ minimising $R$, which all limit points of $\tht_\rho$ must equal, so $\tht_\rho$ converges to it. Clearly it suffices that $\Psi_0$ be merely convex and continuous. In particular the proof goes through if it is the sum of the un-squared shortfalls.

(i) Replacing all positive parts with variables greater than or equal to them, minimising $\Psi_0$ is then a linear programme in $\tht$ augmented by these additional variables. Minimisers lie in $\mathcal M_0$, and $R$ depends convexly only on $\tht$. Adding a convex function to a linear programme to regularise it is exact \citep{friedlander2007exact}, that is, for some $\bar\rho$, all smaller positive $\rho$ give minimisers also solving the original programme and with least $R$ amongst them; in particular, with $\tht$ equal to $\tht_0$.

(ii) If for positive $\rho$, $\tht_\rho$ is in $\mathcal M_0$, then $R$ is minimised over $\mathcal M_0$ at $\tht_\rho$, which is therefore equal to $\tht_0$. Clearly if shortfalls are all squared, then for $\tht_0$ the function to be minimised, over the whole space, is differentiable and non-negative with minimum 0, so has zero gradient. The necessary and sufficient condition that this should minimise the augmented function over $\Theta$ is that $\tht_0$ minimise $R$, which it will do if and only if $(N^{-1}\one,q^{-1}\one)$ lies in $\mathcal M_0$. If so, clearly it minimises each function separately.
\end{proof}

\begin{proof}[Proposition~\ref{prop:family}]
(i) If both $\tau$ and $\delta$ are 0 the 2 ends coincide at $c_i$, and the expression is simply $(\upsilon-c_i)^2$ for all $\upsilon$. (ii) For $\tau=1$ they are $b_{y_i-1}+\delta$ and $b_{y_i}-\delta$, so the 2 terms in \eqref{eq:loss} are the 2 squared shortfalls, both 0 if and only if $b_{y_i-1}+\delta\le U(a_i)\le b_{y_i}-\delta$. (iii) For $\delta=0$, $\ell_i=0$ if and only if $U(a_i)$ is no further than $\tau h_i$ from $c_i$, that is, if $U(a_i)-b_{y_i-1}$ and $b_{y_i}-U(a_i)$ are both no less than $(1-\tau)h_i$. (iv) Follows from \eqref{eq:zone}; each term of \eqref{eq:loss} and hence their sum share the same monotonicity properties.
\end{proof}

\begin{proof}[Proposition~\ref{prop:huber}]
(i) Straightforward: the relevant function has value and first derivative, for non-positive $x$, between 0 and $\zeta$, and greater than $\zeta$, equal to $0$, $x^2$, $2\zeta x-\zeta^2$, and $0$, $2x$, $2\zeta$ respectively. Hence each term has the stated properties, as do the additional terms if the shortfalls are all squared; adding the ridge terms gives strong convexity. If not squared and without these terms, a linear programme results as for Proposition~\ref{prop:vanishing}. (ii) Denote the criterion including Huber terms by $F_\zeta$ and its minimiser by $\tht_\zeta$. Clearly $F_\zeta$ is continuous jointly in its arguments. Hence if $\zeta_n$ tends to a strictly positive limit, any limit point $\bar\tht$ of corresponding minimisers must itself be a minimiser for that value of $\zeta$, and so all tend to $\tht_\zeta$. The 2 functions to minimise have equal derivatives for arguments up to $\zeta$. Hence if the largest shortfall at the minimiser for least squares does not exceed $\zeta$, that point satisfies the conditions for being $\tht_\zeta$. But all arguments are between 0 and $1+\delta$, and so all functions are identical if $\zeta$ is at least this bound. With fixed ridge weights, each Huber term lies between 0 and $2\zeta(1+\delta)$, tending uniformly over the domain to 0 as $\zeta$ tends to 0, so $\tht_\zeta$ tends to the minimiser of the ridge terms alone, $(N^{-1}\one,q^{-1}\one)$. To find the limit of the rescaled criteria, note that the difference of identity and Huber functions divided by $2\zeta$ is never negative and never exceeds $\zeta/2$. Hence dividing the whole criterion by $2\zeta$ gives a function differing from that with absolute values, call it $f$, by never more than $m\zeta$. So its minimiser satisfies $f$ no greater than at its own minimiser $\tht^\ast$ by $2m\zeta$; strong convexity then gives the result. (iii) Immediate from the form of the additional terms and the monotonicity of the limits.
\end{proof}

\begin{proof}[Theorem~\ref{thm:known}]
This finds the minimum of $\tfrac12\vv\T G\vv-\vv\T\Ph\T \cc$, strictly convex, over $\Delta_N$. There is therefore a unique solution $\vv^\star$ satisfying the KKT conditions \citep{boyd2004}: for some $\mu$ and some $\boldsymbol\lambda\in\R^N_{+}$,
\[
G\vv-\Ph\T \cc-\mu\one-\boldsymbol\lambda=\zero,\qquad \one\T \vv=1,\qquad \vv\ge\zero,\qquad \lambda_jv_j=0\ \ \text{all } j .
\]
For $j\in F=\operatorname{supp}(\vv^\star)$ this reduces to $G_F\vv_F=\mu\one+\Ph_F\T \cc$. Solving for $\vv_F$ and applying $\one\T$ to both sides, since $\one\T\vv_F=1$, shows that $\mu=\mu_F$ and yields \eqref{eq:vstar}. The other requirement of \eqref{eq:cert} follows at once, taking $\lambda_j=(G\vv-\Ph\T \cc)_j-\mu_F$. If \eqref{eq:cert} is satisfied, $\vv$, $\mu_F$ and $\boldsymbol\lambda$ with $\boldsymbol\lambda_F=\zero$ clearly meet the KKT conditions and so define the required minimum. Thus $\vv^\star$ varies affinely with $\cc$ over each region corresponding to $F$. Since there are a finite number of such regions and $\vv^\star$ may alternatively be regarded as projecting $G^{-1}\Ph\T\cc$ onto $\Delta_N$ with respect to an inner product defined by $G$ \citep{bemporad2002explicit}, it varies continuously with $\cc$.
\end{proof}

\begin{proof}[Corollary~\ref{cor:free}]
Taking $F$ to be all coordinates, there is no second requirement of \eqref{eq:cert}; hence $\vv^\circ$ will be the solution if non-negative. Otherwise it would be the unique minimum over all $\vv$ satisfying $\one\T\vv=1$, of which $\Delta_N$ is a subset, and so there would be strictly lower values over this set. Nor could $\vv^\star$ lie strictly inside $\Delta_N$, since then it would also be the minimum over all $\vv$ satisfying $\one\T\vv=1$.
\end{proof}

\begin{proof}[Theorem~\ref{thm:unknown}]
There is a unique minimum, $\tht^\star$, by Proposition~\ref{prop:convex}. The inequalities not binding at this minimum can be ignored, so the problem is equivalent to minimising $\Psi^{\mathrm c}$ subject only to $\vv_{F^c}=\zero$, $\one\T \vv=1$, $\cw_{E^c}=\varepsilon\one$, $\one\T\cw=1$. By Theorem~\ref{thm:known}, with $\cw$ fixed, $\vv_F$ should be as shown, with no restriction on positivity, that is, $\vv_F(\cw)=W_F\Ph_F\T \bar Z\cw+\ww_F$. Differentiating with respect to $\cw$ yields
\[
2\bar Z\T\bigl(\bar Z\cw-\Ph_F\vv_F(\cw)\bigr)+2\eta(\cw-q^{-1}\one)=2(S\cw-\rr).
\]
So the remaining problem is to minimise $\tfrac12\cw\T S\cw-\rr\T\cw$ over the appropriate set. Restricting to $E$ therefore requires $S_E\cw_E+\varepsilon S_{E,E^c}\one=\rr_E+\nu\one$, with $\nu$ chosen so that $\one\T\cw_E=1-\varepsilon|E^c|$. Equation~\eqref{eq:kstar} follows. It follows from $W_F\preceq G_F^{-1}$, $\Ph_FG_F^{-1}\Ph_F\T\preceq I$ and $W_F\succeq0$ that $\Pi_F\preceq I$ and $\Pi_F\succeq0$, respectively; hence $S\succeq\eta I$ and all its principal submatrices, including $S_E$, are non-singular.
\end{proof}

\begin{proof}[Corollary~\ref{cor:ordered}]
Clearly $\cw\ge\varepsilon\one$ is a constraint on all solutions $(\mathrm J)$. It is the only condition which might restrict elements of $S^{-1}(\rr+\nu\one)$ being negative.
\end{proof}

\begin{proof}[Theorem~\ref{thm:fixed}]
Clearly $Q(\tht;\tht)=\Psi(\tht)$. Also each term of $Q(\cdot\,;\tht)$ will either coincide locally with a term of $\Psi$, if the relevant observation strictly exceeds one end of its zone, or be 0 with 0 gradient at $\tht$, if the observation falls inside its zone or coincides with its lower end. Hence $Q(\cdot\,;\tht)$ and $\Psi$ have common gradients at $\tht$. Both are convex and continuously differentiable functions. Hence $\tht^\star$ will minimise $\Psi$ over convex $\Theta$ if and only if $\nabla \Psi(\tht^\star)\T(\vartheta-\tht^\star)\ge0$ whatever $\vartheta\in\Theta$, which is equivalent to $\nabla Q(\tht^\star;\tht^\star)\T(\vartheta-\tht^\star)\ge0$ whatever $\vartheta\in\Theta$, that is, to $\tht^\star$ minimising $Q(\cdot\,;\tht^\star)$ over $\Theta$.
\end{proof}

\begin{proof}[Corollary~\ref{cor:general}]
By Theorem~\ref{thm:fixed} we require minimisation of \eqref{eq:Q} over $\vv$ and $\cw$, which reduces, by an argument parallel to that for Theorem~\ref{thm:unknown}, to the appropriate linear subspace. Thus with $\cw$ fixed, $\vv_F(\cw)=W_F\Ph_{V,F}\T(T_V\cw+\boldsymbol a_V)+\ww_F$. Differentiating through w.r.t.\ $\cw$ gives the required expressions for $S$ and $\rr$, Corollary~\ref{cor:general}. The remainder of the proof is analogous. There is no requirement that observations differ, so 2 could correspond to 1 observation violating both ends of an empty interval.
\end{proof}

\begin{proof}[Theorem~\ref{thm:conv}]
Set $\varsigma=2\min(\rho,\eta)$, $\boldsymbol p^t=\hat\tht^t-\tht^t$ and $Q_t=Q(\cdot\,;\tht^t)$. All $Q_t$ are $\varsigma$-strongly convex. There are only a finite number of possible Hessians $H_t$ of $Q_t$, corresponding to partitions of the observations, with maximum modulus $L$ say. Hence also the maximum modulus of the pieces of the piecewise-linear gradient of $\Psi$, so $L$ may also be regarded as its modulus of continuity.

(i) Since $\hat\tht^t$ minimises $Q_t$,
\[
\tfrac{\varsigma}{2}\|\boldsymbol p^t\|^2\le Q_t(\tht^t)-Q_t(\hat\tht^t)\le-\nabla Q_t(\tht^t)\T \boldsymbol p^t=-\nabla \Psi(\tht^t)\T \boldsymbol p^t,
\]
the second inequality following from convexity. Hence $\boldsymbol p^t$ is a direction of descent unless 0, in which case $\tht^t=\tht^\star$.

(ii) For any $\chi$ between 0 and 1,
\[
\Psi(\tht^t+\chi \boldsymbol p^t)\le \Psi(\tht^t)+\chi\nabla \Psi(\tht^t)\T \boldsymbol p^t+\tfrac{L}{2}\chi^2\|\boldsymbol p^t\|^2\le \Psi(\tht^t)-\chi\tfrac{\varsigma}{2}\|\boldsymbol p^t\|^2+\tfrac{L}{2}\chi^2\|\boldsymbol p^t\|^2 .
\]
Taking $\chi=\min(1,\varsigma/(2L))$ and noting that the line search is exact,
\[
\Psi(\tht^t)-\Psi(\tht^{t+1})\ge \tfrac{\varsigma}{4}\min\bigl(1,\tfrac{\varsigma}{2L}\bigr)\|\boldsymbol p^t\|^2 .
\]
Since $\Psi\ge0$, $\boldsymbol p^t$ must approach 0. Since $\hat\tht^t$ minimises $Q_t$ over $\Theta$,
\[
(\nabla \Psi(\tht^t)+H_t\boldsymbol p^t)\T(\vartheta-\hat\tht^t)\ge0\quad\text{whatever } \vartheta\in\Theta .
\]
Let $\tht^{\infty}$ be a limit point of the sequence $\tht^t$. Then, along an appropriate subsequence, $\hat\tht^t$ will also tend to $\tht^{\infty}$, and $H_t\boldsymbol p^t$ to 0, so, by continuity of the gradient, $\nabla \Psi(\tht^{\infty})\T(\vartheta-\tht^{\infty})\ge0$ whatever $\vartheta\in\Theta$, that is, $\tht^{\infty}=\tht^\star$. Since any limit point must equal $\tht^\star$, the whole sequence converges to $\tht^\star$.

(iii) If Assumption~\ref{ass:nondeg} holds, all relevant inequalities are strict at $\tht^\star$, hence over some neighbourhood $\mathcal V$ of $\tht^\star$. From (ii), some iterate will lie in $\mathcal V$. For this and all subsequent $t$, $Q_t=Q(\cdot\,;\tht^\star)$ and so $\hat\tht^t=\tht^\star$; hence $\chi_t=1$ since $\tht^\star$ uniquely minimises $\Psi$ and $\tht^{t+1}=\tht^\star$. For centred least squares, $\underline t_i=\overline t_i=c_i$ and so all observations fall into $\mathcal B(\tht^0)$ or $\mathcal A(\tht^0)$ and $Q(\cdot\,;\tht^0)=\Psi^{\mathrm c}$. Hence $\hat\tht^0=\tht^\star$ and $\chi_0=1$.
\end{proof}

\begin{proof}[Proposition~\ref{prop:compatible}]
(i) follows from Proposition~\ref{prop:family}(ii), since all observations must be correctly classified to attain the minimum possible value 0 of the loss function. (ii) from Proposition~\ref{prop:family}(i).
\end{proof}

\begin{proof}[Proposition~\ref{prop:population}]
Trivially, with $\cw=(b,1-b)$, the centres are $b/2$ and $(1+b)/2$; and $g_1(a)=x$ would be assigned to category 1 iff $x$ is less than $b^\dagger$. Hence average loss would be $f(b)=\mathrm E\bigl[(x-b/2)^2\mathbf 1\{x<b^\dagger\}+(x-(1+b)/2)^2\mathbf 1\{x\ge b^\dagger\}\bigr]$. Differentiating,
\[
f'(b)=-\mathrm E[x]+\tfrac b2\Pr(x<b^\dagger)+\tfrac{1+b}2\Pr(x\ge b^\dagger)=\tfrac12\bigl(b-b^\circ\bigr),\qquad f''(b)=\tfrac12 .
\]
So average loss is convex quadratic in $b$, minimised without constraint at $b^\circ$; constrained to an interval it will be minimised at the nearest point to $b^\circ$. Constraints reduce to restricting $b$ to lie between $\varepsilon$ and $1-\varepsilon$.
\end{proof}

\begin{proof}[Proposition~\ref{prop:recal}]
Which upper or lower limit determines the category of any value depends only on the number of other limits at or below it, so the total loss for any $\boldsymbol b$ depends only on the sequence of numbers $0\le x_1\le\dots\le x_{q-1}\le n_z$ of values less than $b_1,\dots,b_{q-1}$, with ties all allocated to one category. Setting $x_0=0$, $x_q=n_z$, the total loss is $\sum_{k=1}^q e_k(x_{k-1},x_k)$. Define $\Lambda_k(x)$ as the minimum total loss $\sum_{s\le k}e_s(x_{s-1},x_s)$ subject to $0=x_0\le\dots\le x_k=x$. Since only the final term varies with $x_{k-1}$ this is easily evaluated recursively, and the required minimum loss is given by $\Lambda_q(n_z)$. Calculating all $e_k(x',x)$ requires $O(n_z^2)$ operations, after calculating suitable cumulative totals, so evaluation of all $\Lambda_k(x)$ requires $O(qn_z^2)$ operations. It remains only to note that all such sequences may arise from suitable choices of the limits, counting the category of any value $z$ as 1 plus the number of $b_k\le z$. If $1\le x_k\le n_z-1$ this will be true of any $b_k$ between $z_{(x_k)}$ and $z_{(x_k+1)}$; if $x_k=0$, of any $b_k\le z_{(1)}$, including 0 if any value is estimated as 0; if $x_k=n_z$, of any $b_k>z_{(n_z)}$, necessarily exceeding 1 if $z_{(n_z)}=1$, explaining why recalibrated bounds need not lie between 0 and 1. Obvious modifications apply if successive $x_k$ are equal, corresponding to equal limits and hence an empty category, and the limits may evidently be taken monotone.
\end{proof}

\bibliographystyle{apacite}
\bibliography{refs_apa}

\end{document}